\pdfoutput=1
\documentclass[11pt]{article}

\usepackage{acl}
\usepackage{times}
\usepackage{latexsym}
\usepackage{graphicx}
\usepackage{subcaption}
\usepackage{comment}
\usepackage{amsmath}
\usepackage{float}
\usepackage{multirow}
\usepackage{amsmath, amssymb}

\usepackage[T1]{fontenc}
\usepackage[utf8]{inputenc}

\usepackage{microtype}

\usepackage{inconsolata}

\title{Model Editing at Scale leads to Gradual and Catastrophic Forgetting}

\author{Akshat Gupta, Anurag Rao, Gopala Anumanchipalli\\
 UC Berkeley\\
 \texttt{akshat.gupta@berkeley.edu}}

\begin{document}
\maketitle
\begin{abstract}
Editing knowledge in large language models is an attractive capability that allows us to correct incorrectly learned facts during pre-training, as well as update the model with an ever-growing list of new facts. While existing model editing techniques have shown promise, they are usually evaluated using metrics for reliability, specificity and generalization over one or few edits. We argue that for model editing to have practical utility, we must be able to make multiple edits to the same model. With this in mind, we evaluate current model editing methods at scale, focusing on two state of the art methods - ROME and MEMIT. With the lens of scalability, we evaluate model editing methods for three crucial properties - editing proficiency, fact forgetting and downstream performance. We find that as a model is edited sequentially with multiple facts, it continually becomes less editable, forgets previously edited facts and loses the ability to perform downstream tasks. For ROME and MEMIT, this "\textit{forgetting}" happens in two phases - an initial gradual but progressive forgetting phase followed by an abrupt or catastrophic forgetting. Both gradual and catastrophic forgetting limit the usefulness of model editing methods at scale - the former makes model editing less effective as multiple edits are made to the model while the latter caps the scalability of such model editing methods. Our analysis also highlights other key limitations of ROME and MEMIT at scale. With our work, we push for better evaluation of model editing and development of model editing methods keeping scalability in mind. More information can be found at the paper webpage - \url{https://scalable-model-editing.github.io/catastrophic}
\end{abstract}

\section{Introduction}
Editing knowledge in large language models (LLM) has recently emerged as a sought after capability for natural language processing (NLP) practitioners. It is a known fact that LLMs memorize some part of their training data \citep{gpt-2, quantifying_memorization}. Model editing, sometimes also called knowledge editing\footnote{In this paper, we use knowledge editing and model editing interchangeably}, is the task of modifying existing knowledge or injecting new facts into the model \cite{ripple-effects}, without updating the entire model. While knowledge in LLMs can be updated by continually pre-training models on new facts, this is not always viable due to the huge costs of training LLMs and an ever-growing list of facts. Even then, LLMs do not perfectly memorize training data \cite{quantifying_memorization} and will inevitably memorize facts incorrectly, thus requiring the capabilities of model editing. This is what makes model editing a very useful tool in the arsenal when working with LLMs.

Various methods have been proposed over the years to do so. Some of these methods \citep{metamodel, MEND} require training a hypernetwork \cite{hypernetwork} that generates new weights for the model being edited. Other methods \citep{ROME, MEMIT} directly update specific parts of the model after locating stored facts inside it \citep{key-value-memories}. The success of knowledge editing is usually evaluated along three dimensions \citep{editing-survey, ROME, MEMIT} - (i) reliability, which measures if the post-edit model accurately recalls the newly edited fact, (ii) generalization, which measures if the post-edit model accurately recalls the newly edited facts for different phrasings of the edited fact, and (iii) locality (also known as specificity \citep{ROME, MEMIT}), which measures if the edited fact changes unrelated or neighboring facts.

%While many of these methods have shown promise \cite{editing-survey}, recent work analyzing the after-effects of these editing methods have highlighted the shortcomings of these methods. Specifically, while some of these editing methods rank high on reliability, generalization and locality metrics \citep{editing-survey, MEND, SERAC, ROME, MEMIT}, the edited knowledge is not used consistently by the model. \citet{ripple-effects} propose a new evaluation system where the "ripple effects" or implications of an edited fact are evaluated. An example of such ripple effects would be - if an edited fact updates the president of a country to the new president, then prompting for the birthplace of the president should output the birthplace of the new president. \citet{pitfalls} extend this by introduce the concept of "knowledge conflict" and additional edit types like reverse-edits and round-edits, thus evaluating the logical consistency of model editing in more complex scenarios.

%One very important aspect that is not given enough attention during evaluation of model editing is making multiple edits on the same model. 

Most prior works \citep{metamodel, MEND, ROME, ripple-effects, pitfalls} focus on evaluating model editing performance by making one edit at a time. While some recent works \citep{MEND, SERAC, MEMIT} have begun evaluating model editing with multiple edits made to the same model, we find the evaluation in this setting to not be exhaustive. In an ideal realization of model editing, these methods should be used to update thousands if not hundreds of thousands of facts. When multiple edits were made to the same model sequentially, \citet{MEMIT} see a significant decrease in performance for all methods, starting as low as 10 edits. This brings to attention a very important question - do these methods scale? Which is why in this paper, \textbf{we keep scalability at the center when evaluating model editing}. 

\begin{table}
\centering

\scalebox{0.9}{
\begin{tabular}{c|p{4cm}}
%\hline
\textbf{Dataset Name} & \textbf{Query} \\ \hline
zsRE & Who was the architect of Villa Kampen? \\
CounterFact & Porto is a twin city of \\
%\hline
\end{tabular}
}

\caption{Examples of input prompts from zsRE and CounterFact.}
\label{table:dataset}
\end{table}

%Most of prior works \citep{metamodel, MEND, ROME, ripple-effects, pitfalls} focus on evaluating model editing performance when only a single edit is made to the model. Other works find that making multiple edits, either in a sequential or batched manner, leads to significant degradation of performance \citep{MEND, SERAC, MEMIT}. In an ideal realization of editing knowledge in LLMs, these methods should be used to updates thousands if not hundreds of thousands of facts. When multiple edits were made to the same model sequentially, \citet{MEMIT} see a significant decrease in performance for all methods, starting as low as 10 edits. This brings to attention a very important question - do these methods scale? With the popularity of in-context learning \cite{icl} and retrieval augmented generation (RAG) methods \cite{rag}, model editing only starts to become useful at scale \citep{ripple-effects, mquake}. Which is why in this paper, \textbf{we keep scalability at the center of evaluating model editing in LLMs}. 

In this paper, we present a framework to study the dynamics of model editing at scale. We make multiple edits to the same model sequentially and analyze the effects of these edits on the model as a function of the number of edits. Specifically, we compare model editing at scale for three of the most popular model editing methods - MEND \cite{MEND}, ROME \cite{ROME} and MEMIT \cite{MEMIT} against a fine-tuning baseline, and make multiple edits on the same model. Along with the usual metrics for evaluating model editing, we propose evaluating post-edit models for three important properties - editing proficiency, fact forgetting, and downstream task performance. 

We find that ROME and MEMIT outperform MEND and fine-tuning baselines at scale, but edits made by ROME and MEMIT are not as localized as previously believed to be. We show that new edits consistently bleed into other facts stored in the model. We also find the model editing methods are prone to gradual and catastrophic forgetting. To the best of our knowledge, our work is the first to associate model editing methods with catastrophic forgetting. We define gradual forgetting as a progressive loss of ability of the model to perform its regular functions as the model is continuously modified through knowledge edits. This includes forgetting the ability to recall previously edited facts as well doing downstream tasks. For ROME and MEMIT, gradual forgetting is followed by sudden or catastrophic forgetting \citep{catastrophic, catastophic2}, where the model gets crippled due to a single update made to the model. We name these edits - \textit{\textbf{disabling edits}}. A disabling edit decapitates the model - new knowledge edits are no longer successful on the model, the model forgets all previously edited facts and is unable to perform any downstream tasks. We also study different properties of these disabling edits. With this paper we highlight some serious limitations of current model editing techniques, especially their lack of robustness when scaled and call for further research in developing scalable model editing methods. Our code can be found \href{https://github.com/scalable-model-editing/gradual-catastrophic-forgetting}{here}.

\section{Methods, Models and Datasets}
In this paper, we focus on two prominent model editing methods in literature - Rank-One Memory Editing (ROME) \cite{ROME} and Mass-Editing Memory in a Transformer (MEMIT) \cite{MEMIT}, whereas Model Editor Networks using Gradient Decomposition (MEND) \cite{MEND} and fine-tuning are used as baselines. MEND is a hypernetwork \cite{hypernetwork} based model editing method that generates the weight updates for the model being edited. ROME and MEMIT first localize knowledge within a model using causal tracing \cite{causal} and then update the weights of the selected layers to inject knowledge. The major difference between ROME and MEMIT is that while ROME works under the assumption that knowledge in LLMs can be updated by updating a single layer, MEMIT updates the weights of multiple layers. We evaluate these model editing methods over two models - GPT2-XL (1.5B) \cite{gpt-2} and GPT-J (6B) \cite{gpt-j}. Note that all prior works \citep{MEND, ROME, MEMIT} edit base language models and not chat models. We refer the reader to appendix \ref{sec:related_work} for a detailed survey of model editing techniques. This section will be added to the main paper upon acceptance.

Knowledge editing is usually evaluated on two datasets - the zsRE (zero-shot relation extraction) dataset \cite{zsre} and the CounterFact dataset \cite{ROME}. The zsRE dataset contains facts in the form of question-answer (QA) pairs created from Wikipedia. A key distinction between zsRE and CounterFact datasets is that zsRE contains true facts, which are easier for the model to learn, whereas CounterFact contains counterfactual examples where the new target has lower probability when compared to the original answer \cite{ROME}. Examples from these datasets can be seen in Table \ref{table:dataset}, which highlights another key difference between the two datasets. The queries in the zsRE dataset are present in the form of questions whereas for the CounterFact dataset, the queries are in the form of a prompt and is followed by an answer, which is a more natural formulation for base language models as they are trained to complete a sentence. 

To check for the applicability of the zsRE dataset for editing base models, we first edit a fact in the QA format (as shown in Table \ref{table:dataset}), and then evaluate the success of the edit by prompting the question in a text completion format ("The architect of Villa Kampen is" ). We find that the model is unable to produce the correct answer in approximately 70\% scenarios when prompted in a text completion format after successful edits in the QA format. This shows that facts edited using the zsRE dataset do not become a part of the text generation process (details in Appendix \ref{sec:appendix:zsre}). We thus choose the CounterFact dataset for our experiments. 

The CounterFact dataset contains 21,919 counterfactual statements. Each datapoint in the dataset is a triplet of the type \texttt{(subject, relation, object)}. The dataset is created such that the target object is less likely than the original object in the relation triplet. CounterFact is thus also a more difficult dataset \cite{ROME} as we are going against the knowledge of the model, which is exactly what we do when we edit or inject facts in a model.

\begin{figure*}
    \centering
    \begin{subfigure}{.24\textwidth}
        \centering
        \includegraphics[width=\linewidth]{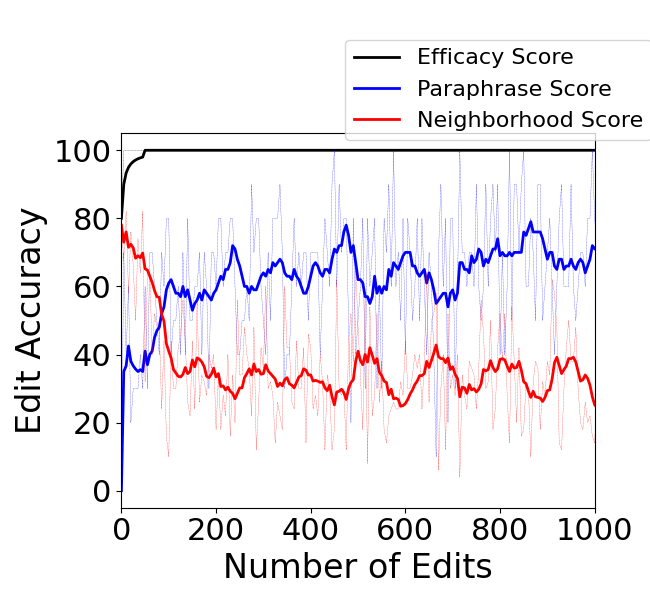}
        \caption{FT-C}
        \label{fig:editing_proficiency:FT-C}
    \end{subfigure}%
    \begin{subfigure}{.24\textwidth}
        \centering
        \includegraphics[width=\linewidth]{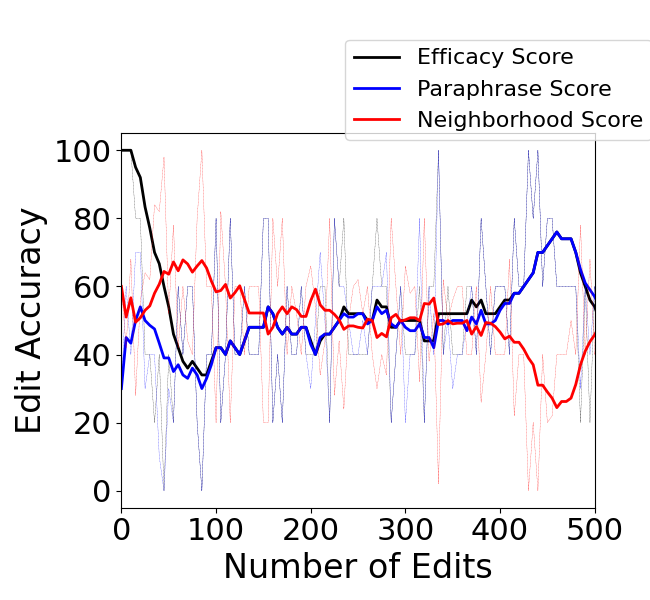}
        \caption{MEND}
        \label{fig:editing_proficiency:MEND}
    \end{subfigure}%
    \begin{subfigure}{.24\textwidth}
        \centering
        \includegraphics[width=\linewidth]{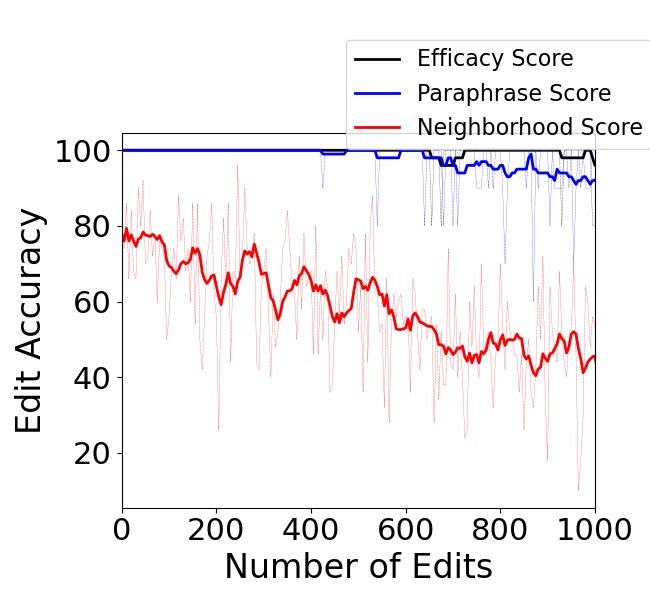}
        \caption{ROME}
        \label{fig:editing_proficiency:ROME}
    \end{subfigure}
    \begin{subfigure}{.24\textwidth}
        \centering
        \includegraphics[width=\linewidth]{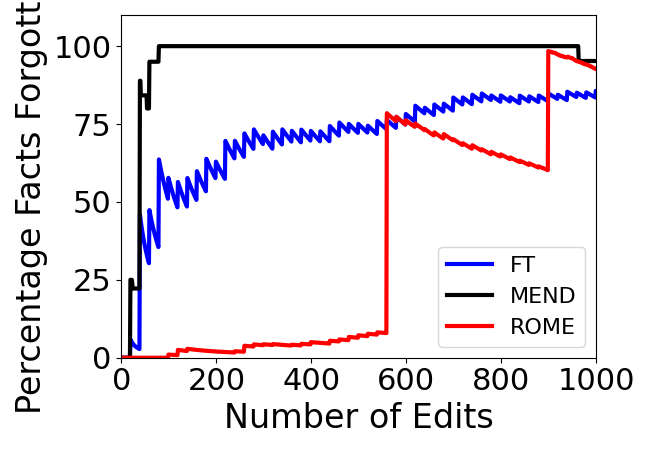}
        \caption{\% Facts Forgotten}
        \label{fig:forgetting}
    \end{subfigure}
    
    \caption{This figure shows the editing proficiency of FT-C, MEND, and ROME on GPT-J (6B). The dotted line represents the metric averaged over a past window size of 5, whereas the solid lines represent the metric averaged over a past window size of 50. Figure \ref{fig:forgetting} show the percentage of previously edited facts forgotten as a function of number of edits.}
    \label{fig:editing_proficiency}
\end{figure*}

\section{Scaling ROME}\label{sec:rome}
In this section, we evaluate the performance of Rank-One Memory Editing (ROME) method \cite{ROME} when multiple sequential edits are made to the same model. We compare the performance of ROME with MEND and fine-tuning (FT-C) baseline. We use the standard implementation of ROME and MEND. For the fine-tuning baselines, we fine tune the same layer being edited by ROME and constrain the norm of change in weights. Our experiments show that if norm is not constrained during fine-tuning, its leads to immediate model degradation. We provide more elaborate implementation details in appendix \ref{sec:appendix:implementation}. 

Edits to the models are made using the CounterFact dataset. To perform these experiments, we create four random subsets of 1000 examples from the CounterFact dataset and sequentially edit the model on the selected datapoints. We do so to find patterns that are independent of the effect of the order in which facts are edited. There is no knowledge conflict \cite{pitfalls} in any of these samples as the same subject is not edited twice.

\subsection{Editing Proficiency at Scale}\label{sec:rome:editing_proficiency}
We first begin by evaluating editing proficiency of ROME as multiple edits are made to the model sequentially. This is done by analyzing the success of a new edit as a function of number edits made. To do so, we measure three metrics, following the convention of \citet{ROME} - efficacy score, paraphrase score, and neighborhood score. 

\textbf{Efficacy score (ES)} measures success when editing a fact, and is measured as true if $P(new fact) > P(old fact)$\footnote{$P(.)$ measures the probability of an event}. 

\textbf{Paraphrase score (PS)} measures if the model is able to recall the edited fact with larger probability when prompted with a paraphrase of the sentence that was used to edit the fact. It is measured as true if $P(new fact) > P(old fact)$ for a paraphrased prompt. Paraphrase score represents the generalization ability of the model editing method. 

\textbf{Neighborhood score (NS)} measures the effect of editing the model on related facts with a different subject, and is measured true if $P(neighborhood fact) > P(new fact)$. Neighborhood score represents specificity of the editing method, and is measured on a set of distinct but semantically related subjects. In this scenario, we want the neighborhood facts to not be affected by model editing. %The paraphrase score and neighborhood score is measured by evaluating performance of editing on 2 paraphrases and 10 neighborhood facts respectively, as given in the CounterFact dataset \citet{ROME, MEMIT}.

Figure \ref{fig:editing_proficiency} shows the different metrics used for measuring editing proficiency of FT-C, MEND, and ROME on GPT-J (6B) as a function of the number of edits made to the model for one of the four samples. Additional experiments for other samples as well as GPT2-XL are shown in appendix \ref{sec:appendix:rome:editability}. The dotted lines represent average metric over a past window of size 5, which is the average of the given metric over 5 previous edits. The solid lines represents the average metric over a window size of 50.

We find that both FT-C and ROME are able to successfully edit facts sequentially, whereas MEND is unable to make multiple sequential edits to the same model, as shown by the efficacy score. This reiterates previous observations \citep{MEND, SERAC, MEMIT} that MEND cannot be used to reliably edit knowledge at scale. For ROME, we find that the efficacy score, which measures the success of an edit, is almost 100\% until a point where it begins to decline. This point of decline can come as early as 100 edits, or as late as 1000 edits made to the model as can be seen in other samples (appendix \ref{sec:appendix:rome:editability}). Prior to this inflection point, while ROME is successful at making edits to the model, its neighborhood accuracy consistently declines as more edits are made to the model, \textbf{indicating that the edits made start to bleed into other fact stored in the model}. We will provide more evidence for this in later sections. These trends are consistent across multiple samples and multiple models. %The same is true for fine tuning, where it exhibits almost perfect efficacy although with declining neighborhood score. % In our analysis, we also find that paraphrase accuracy and neighborhood accuracy are not the most effective measures of generalization and specificity as presented by \citep{ROME, MEMIT}. However, they can be seen as approximate indicators of the above properties. We provide a detailed explanation of this in appendix \ref{sec:appendix:metrics}.

%To summarize, we find that ROME is extremely proficient at successfully editing knowledge but only up to a certain point, after which, suddenly the method breaks down. Although even before that point is reached, the edited knowledge does begins to impact the unrelated knowledge stored in the model, as shown by the consistently declining neighborhood score.

%Write a discussion in appendix of why neighborhood accuracy is a weak predictor of specificity.

\begin{comment}

\begin{figure*}
    \centering
    \begin{subfigure}{.32\textwidth}
        \centering
        \includegraphics[width=\linewidth]{Plots/forgetting/ROME_forget_cummulative_correct_1.png}
        \caption{Sample 1}
        \label{fig:forgetting:sample_1}
    \end{subfigure}%
    \begin{subfigure}{.32\textwidth}
        \centering
        \includegraphics[width=\linewidth]{Plots/forgetting/ROME_forget_cummulative_correct_2.png}
        \caption{Sample 2}
        \label{fig:sub2}
    \end{subfigure}%
    \begin{subfigure}{.32\textwidth}
        \centering
        \includegraphics[width=\linewidth]{Plots/forgetting/ROME_forget_cummulative_correct_3.png}
        \caption{Sample 3}
        \label{fig:sub3}
    \end{subfigure}
    \caption{This figure shows the percentage of facts forgotten as a function of number of edits made to GPT-J (6B) using ROME on the same three samples as used in Figure \ref{fig:editing_proficiency}. The percentage is calculated over the number of facts that were successfully edited up to given number of edits.}
    \label{fig:forgetting}
\end{figure*}

\end{comment}

\begin{figure*}
    \centering
    \begin{subfigure}{.24\textwidth}
        \centering
        \includegraphics[width=\linewidth]{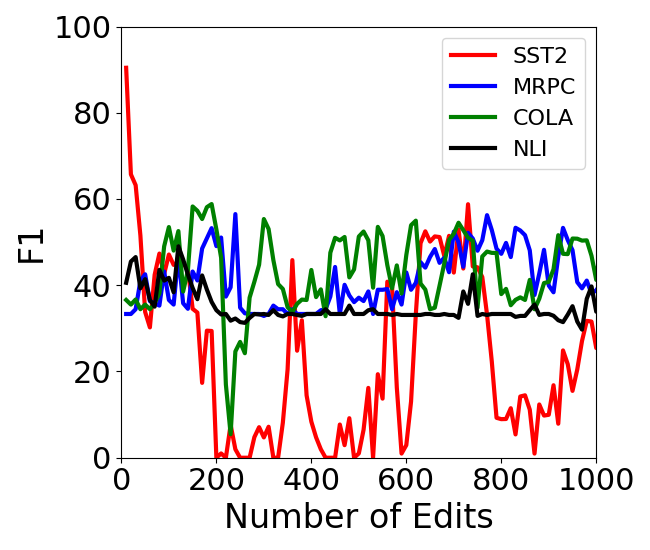}
        \caption{FT-C}
        \label{fig:downstream:FT-C}
    \end{subfigure}%
    \begin{subfigure}{.24\textwidth}
        \centering
        \includegraphics[width=\linewidth]{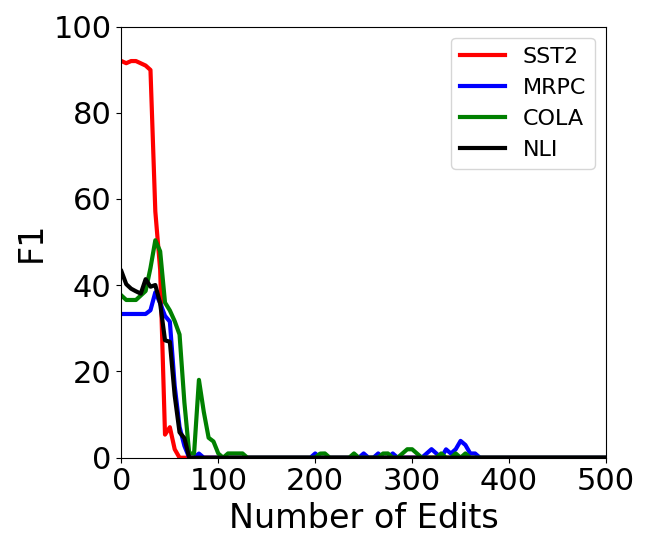}
        \caption{MEND}
        \label{fig:downstream:MEND}
    \end{subfigure}%
    \begin{subfigure}{.24\textwidth}
        \centering
        \includegraphics[width=\linewidth]{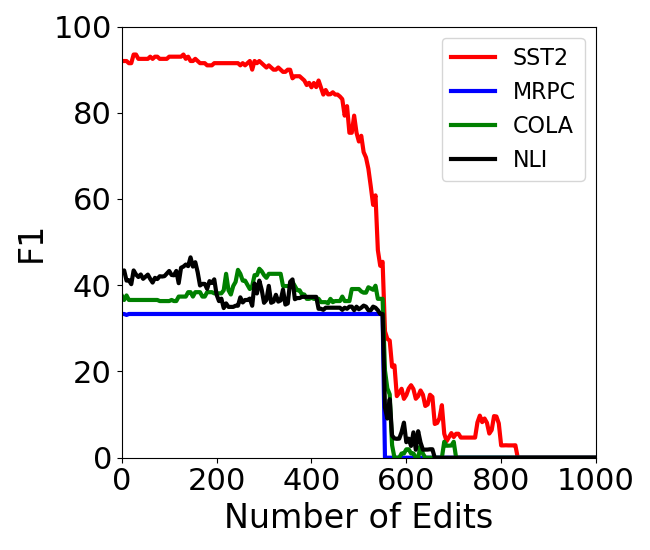}
        \caption{ROME}
        \label{fig:downstream:ROME}
    \end{subfigure}
    \begin{subfigure}{.24\textwidth}
        \centering
        \includegraphics[width=\linewidth]{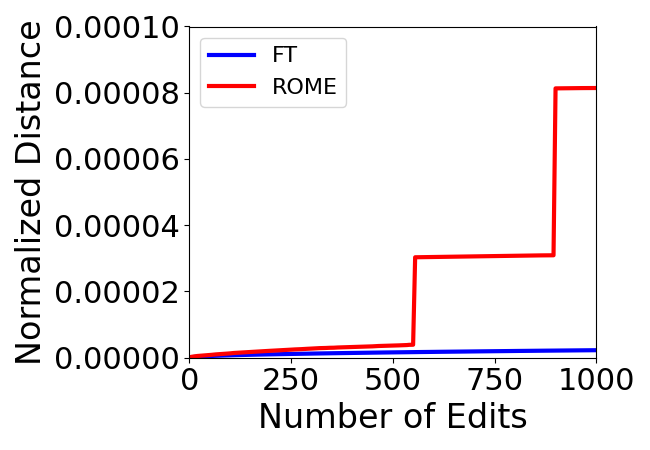}
        \caption{Distance}
        \label{fig:distance}
    \end{subfigure}%
    \caption{This figure shows the downstream performance of editing GPT2-J on four GLUE tasks for different model editing methods. Figure \ref{fig:distance} shows the the normalized distance between the edited layer and its original weights.}
    \label{fig:downstream}
\end{figure*}

\subsection{Gradual and Catastrophic Forgetting}\label{sec:rome:forgetting}
As new facts get added successfully to the model, is the model able to remember previously edited facts? This is the question we try to answer in this section. Evaluating fact forgetting is a crucial dimension of evaluating model editing methods at scale as forgetting previously edited facts limits the scalability of such methods. Additionally, forgetting is a direct indication of locality. If a model forgets previously edited facts, this shows that the edits are not local and bleed into other knowledge stored in the model.

Figure \ref{fig:forgetting} shows the number of previous correctly edited facts that get forgotten as a function of new edits made to GPT-J. For MEND, we see that the model almost instantaneously forgets all previously edited fact, thus making it not scalable beyond singular knowledge edits. For FT-C, we find that the model forgets previously edited facts rapidly as a function of newer edits made to the model, and at a time only retains a handful of prior edits. This also means that edits made using FT-C are highly non-local. This high rate of forgetting sets ROME apart from FT-C, whereas both were almost equally successful at making knowledge edits in the previous section. 

 %This indicates that the inflection point observed in Figure \ref{fig:editing_proficiency} has additional consequences other than just the loss of editability of the model.. %We will discuss this inflection point in more detail in the coming sections. 

%For ROME, we see a clear spike in forgetting previously edited facts at approximately the the same locations where the efficacy score in Figure \ref{fig:editing_proficiency:ROME} starts to decline. Prior to this point,

For ROME, Figure \ref{fig:forgetting} initially shows a slowly increasing relationship between the number of forgotten facts and the number of edits made to the model\footnote{The graphs in figure \ref{fig:forgetting} are evaluated after every 10 edits for computational reasons.} at a rate which is much smaller than the forgetting rate of FT-C. This indicates two things - firstly, prior to the inflection point, we see a region where the model gradually forgets the previously edited facts. Since all edited facts correspond to different subjects, this indicates that editing a single fact with ROME results in implicitly changing of unrelated facts, supporting what was shown in section \ref{sec:rome:editing_proficiency}. Thus, edits made using ROME are not as local as previously believed to be. Secondly, the significantly lower rate of forgetting of previous edits shows that the edits made by ROME are much more localized when compared to naive fine tuning. The same trends are true across different samples and models (appendix \ref{sec:appendix:rome:forgetting}). 

After this region of gradual forgetting of facts for ROME, we reach an inflection point where we find that a catastrophically large number of facts are forgotten by the model. This is the same point where any further knowledge editing also starts to slowly become unsuccessful using ROME (Figure \ref{fig:editing_proficiency:ROME}). This phenomenon of sudden forgetting of a huge number of facts is a realization of catastrophic forgetting in machine learning literature \citep{catastrophic, catastophic2}. Catastrophic forgetting is defined as the sudden loss of ability of a model to perform a prior task when it is further trained to perform a new task. The phenomenon observed in the above example is a perfect realization of how "catastrophic" or abrupt catastrophic forgetting can be, where it literally "forgets" an exploding number of facts with one gradient update. To the best of our knowledge, our work is the first to show that model editing methods are also prone to catastrophic forgetting. 

But is catastrophic forgetting just limited to abruptly forgetting previously edited facts? In the next section, we show that it goes beyond that.

%We see that the effect of this inflection point on editing new facts is slightly delayed, which we believe is a combination of the effect of the moving average window used in Figure \ref{fig:editing_proficiency} and a second reason which we discuss in section \ref{sec:rome:functional_limit}. 

 %We call this point the \textit{dysfunction threshold}.

\subsection{Downstream Evaluation of Edited Models}\label{sec:rome:downstream}
One implicit feature expected out of all model editing methods is that as a fact is edited or inserted into model memory, it does not affect the model's ability to perform its regular functions. This means that knowledge editing should not affect the model's ability to perform common NLP tasks which the model is used for. We call this an implicit assumption because to the best of our knowledge, none of prior works try to directly measure the effect of model editing on downstream tasks\footnote{A concurrent work also proposes evaluating model editing on downstream tasks \cite{hurt}, which is completely coincidental. While their work solely focus on downstream evaluation, our work goes beyond that.}. 

%\citet{ROME, MEMIT} try to capture this effect in some form by calculating the generation entropy, which measures weighted average of the bi-gram and tri-gram entropy of text generated by the post-edit model. The idea is to measure model degradation when generating text using a post-edit model. Ideally, we want the generation entropy of cases with lots of repetitions (a common failure case of model editing \citep{ROME, MEMIT}) to be lower than coherently generated text. We find that this metric is unable to directly quantify the degradation of model ability as multiple edits are made to the model. We explain this in more detail and give examples of failures of generation fluency metric in appendix \ref{sec:appendix:generate_cluency}.

We quantify model degradation by measuring the performance of the post-edit model on common downstream NLP tasks. We choose four tasks from the popular GLUE benchmark \cite{glue} - sentiment analysis (SST2) \cite{sst2}, paraphrase detection (MRPC) \cite{mrpc}, natural language inference (NLI) \citep{nli1, nli2, nli3, nli4} and linguistic acceptability classification \cite{cola} for doing downstream evaluation. All tasks are binary classification tasks and we use a balanced subset of 200 test examples and evaluate the models using the F1-metric every few edits. While a more comprehensive selection of downstream tasks can be created, in this paper, our aim is to show the importance of such an evaluation at scale. We leave a more exhaustive analysis of model editing methods on downstream tasks for future work. More details about implementation of downstream evaluation can be found in appendix \ref{sec:appendix:downstream}. %We evaluate the downstream model performance on these tasks as a function of number of edits made to the model in a few-shot setting. The in-context examples and prompt template used for evaluation are presented in Tables \ref{table:few_shot_sst} and \ref{table:few_shot_mrpc}.

Figure \ref{fig:downstream} depicts the effect of model editing on downstream tasks as a function of the number of edits made to GPT-J. We see consistent model degradation as edits are made to the model across different methods of editing. There is a gradual but continuous degradation of model performance using FT-C, while the effect is sudden for MEND. For ROME, we again see two regions of model degradation. Initially, there is a gradual decline in downstream performance of the model with an increasing number of edits made to the model. We then see an inflection point with an abrupt loss of ability of the model to perform any downstream task, which coincides with the point of sudden decrease in editability of the model (Figure \ref{fig:editing_proficiency:ROME}) and a catastrophic increase in forgetting previously edited facts (Figure \ref{fig:forgetting}). %Additionally, we want to point the user to Sample 1 in Figure \ref{fig:downstream} which contains the maximum number of edits made to the model before the inflection point. We see that the model is no longer able to perform paraphrase detection long before the inflection point, showing that \textit{as the model is edited multiple times, it continuously loses its ability to perform downstream functions and may be unusable for certain downstream tasks even before the inflection point}. ####GOOOD ONE INCLUDE SOMEHOW#####

\begin{comment}

\begin{figure*}
    \centering
    \begin{subfigure}{.32\textwidth}
        \centering
        \includegraphics[width=\linewidth]{Plots/distance/ROME_distance_1.png}
        \caption{Sample 1}
        \label{fig:sub1}
    \end{subfigure}%
    \begin{subfigure}{.32\textwidth}
        \centering
        \includegraphics[width=\linewidth]{Plots/distance/ROME_distance_2.png}
        \caption{Sample 2}
        \label{fig:sub2}
    \end{subfigure}%
    \begin{subfigure}{.32\textwidth}
        \centering
        \includegraphics[width=\linewidth]{Plots/distance/ROME_distance_3.png}
        \caption{Sample 3}
        \label{fig:sub3}
    \end{subfigure}
    \caption{This figure shows the normalized L2 distance between the edited layer weights and its original weights as a function of the number of edits made to the model. The editing algorithm is ROME and the model being edited is GPT2-XL.}
    \label{fig:distance}
\end{figure*}

\end{comment}

While the inflection point is a big concern for model editing methods, there is also a gradual decrease of general ability of the model even prior to that point, which is only visible if the model is evaluated on downstream tasks. This shows the usefulness of evaluation of model editing methods on downstream tasks, which we urge the research community to adopt along with other knowledge editing metrics. We define the first region where the model progressively loses its ability to do prior tasks (like recalling previously edited facts or performing downstream tasks) as \textbf{\textit{gradual forgetting}}, juxtaposing it with catastrophic forgetting. Note that forgetting here does not just refer to forgetting previously edited facts but a general loss of ability to perform a certain function. \textbf{We find that sequential editing of a model leads to these two phases of forgetting in ROME - gradual forgetting and catastrophic forgetting}. We associate the region beyond the point of catastrophic forgetting with catastrophic forgetting as, after this point, model editing becomes ineffective and the model is almost unusable. For FT-C, we only observe a gradual forgetting, whereas For ROME, we see both gradual and catastrophic forgetting.

%This means the the effect of editing knowledge in LLMs on its capability of performing common NLP tasks must be known. None of the prior works try to quantify the effects on knowledge editing on the downstream performance of the model. 

%The effect of knowledge editing on downstream tasks becomes prominent as multiple edits are made to the same model.
\begin{table}
\centering

\scalebox{0.9}{
\begin{tabular}{c|p{2.5cm}|p{2cm}}
%\hline
 & \textbf{DISABLING EDITS} & \textbf{NORMAL EDITS}   \\ \hline
DISTANCE & $3.339  \times 10^{-4 }$ & $8.156 \times 10^{-7}$  \\
%\hline
\end{tabular}
}
\caption{Table showing average distance between edited layer weights from its original weights for disabling versus normal edits. }
\label{table:disabling}
\end{table}

\subsection{The Source of Forgetting}\label{sec:rome:functional_limit}
So far, we've seen that sequential editing of multiple facts in LLMs leads to the model gradually forgetting previously edited facts and losing the ability to be useful for downstream tasks. For ROME, this is followed by an abrupt inflection point, which not only leads to forgetting almost all previously edited facts, but also a complete loss of model ability to perform regular NLP tasks, thus rendering the model useless. Generation examples of model at this point can be seen in Table \ref{table:generation_examples}. This inflection point is a fundamental feature of ROME which can be seen across all samples and models (appendix \ref{sec:appendix:downstream}), and is a realization of extreme catastrophic forgetting. But is this point an outcome of continuous editing of the model or the result of a single edit to the model? What is the reason behind these two phases of forgetting? In this section, we answer these question in more detail.

%Prior to reaching this point, we have a region of gradual forgetting of previously edited facts and loss of model's capability to do downstream task. Once this point is reached, the model begins to completely break down, which also reduces the efficacy of ROME drastically. We call this point the \textbf{dysfunction threshold}. 

Model editing methods are designed with the objective of editing or inserting specific facts stored inside the model without changing all the weights of the model \citep{knowledgeneurons, MEND, ROME, editing-survey}. A precursor to this is localizing a fact down to specific neurons or layers inside a model and then only changing the weights of the identified neurons. The ROME method is built on the assumption that a fact can be changed by changing the weights of any one out of a set of knowledge-storing layers of a model while keeping the rest of the model the same, which is showed to work empirically and backed by causal tracing experiments \cite{ROME}. Each time we make such edits, the edited layer becomes slightly different from its original version. The transformer can be thought of as a machine made from very specific parts working together, where each layer combines the information coming from previous layers with the information contained inside the current layer \citep{transformers, key-value-memories}. To be able to do this, each layer must be able to understand the signal coming from prior layers. In simpler words, there is a notion of compatibility between the different layers of the transformer when they are trained together. As we edit one specific layer of the model continuously while keeping the rest of the model constant, we are constantly changing one part of the model while keeping the remaining part the same. Such a procedure is bound to reach a point where the layer that is changed becomes so different from its original version that this compatibility is destroyed and other parts of the transformer are unable to makes sense of the incoming signal from the layer being edited.

\begin{comment}
    
\begin{figure*}
    \centering
    \begin{subfigure}{.24\textwidth}
        \centering
        \includegraphics[width=\linewidth]{Plots/memit_gpt2xl/MEMIT_editing_score.png}
        \caption{Editing Proficiency}
        \label{fig:memit_gpt2xl:edit_score}
    \end{subfigure}%
    \begin{subfigure}{.24\textwidth}
        \centering
        \includegraphics[width=\linewidth]{Plots/memit_gpt2xl/MEMIT_forget_cummulative_correct.png}
        \caption{Forgetting}
        \label{fig:memit_gpt2xl:forgetting}
    \end{subfigure}%
    \begin{subfigure}{.24\textwidth}
        \centering
        \includegraphics[width=\linewidth]{Plots/memit_gpt2xl/MEMIT_glue_f1.png}
        \caption{Downstream Performance}
        \label{fig:memit_gpt2xl:downstream}
    \end{subfigure}
    \begin{subfigure}{.24\textwidth}
        \centering
        \includegraphics[width=\linewidth]{Plots/memit_gpt2xl/MEMIT_distance.png}
        \caption{Downstream Performance}
        \label{fig:memit_gpt2xl:distance}
    \end{subfigure}
    \caption{This figure shows the editing proficiency of MEMIT on GPT-2XL for Sample 1 over 6000 sequential edits made to the model.}
    \label{fig:memit_gpt2xl}
\end{figure*}
\end{comment}

\begin{figure*}
    \centering
    \begin{subfigure}{.24\textwidth}
        \centering
        \includegraphics[width=\linewidth]{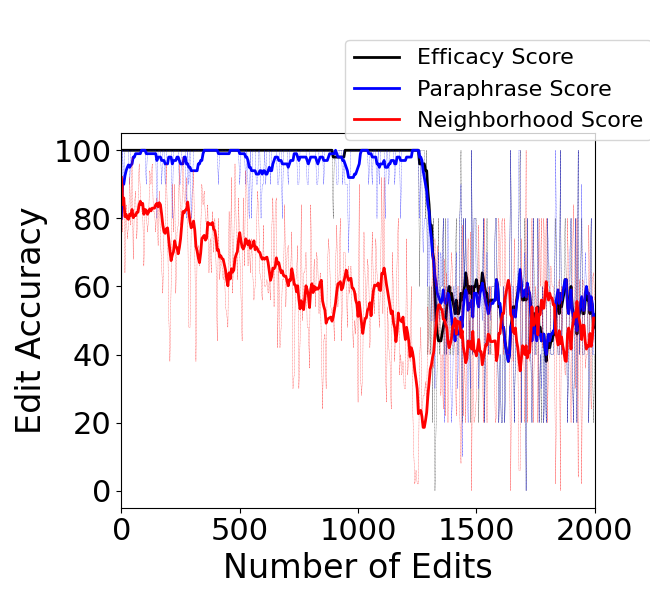}
        \caption{Editing Proficiency}
        \label{fig:memit_gptj:edit_score}
    \end{subfigure}%
    \begin{subfigure}{.24\textwidth}
        \centering
        \includegraphics[width=\linewidth]{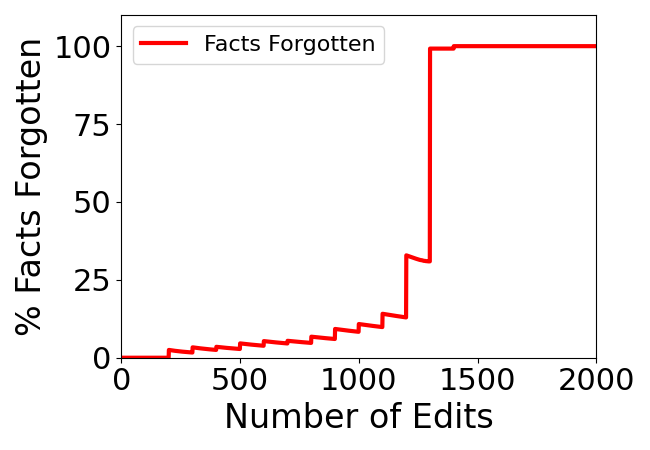}
        \caption{Forgetting}
        \label{fig:memit_gptj:forgetting}
    \end{subfigure}%
    \begin{subfigure}{.24\textwidth}
        \centering
        \includegraphics[width=\linewidth]{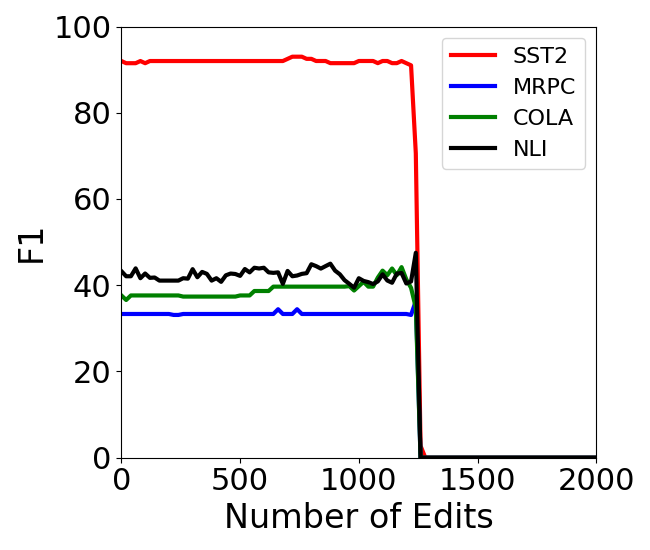}
        \caption{Downstream Performance}
        \label{fig:memit_gptj:downstream}
    \end{subfigure}
    \begin{subfigure}{.24\textwidth}
        \centering
        \includegraphics[width=\linewidth]{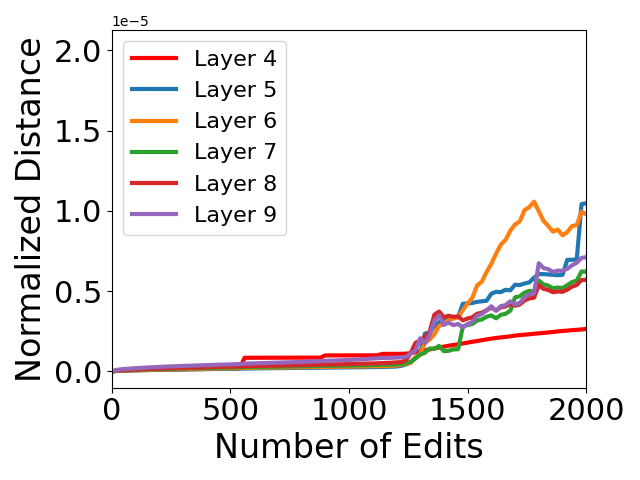}
        \caption{Distance}
        \label{fig:memit_gptj:distance}
    \end{subfigure}
    \caption{This figure shows the editing proficiency of MEMIT on GPT-J for Sample 1 over 2000 sequential edits made to the model.}
    \label{fig:memit_gptj}
\end{figure*}

This is exactly what happens as we continue to make multiple edits to a single layer of the model while keeping the rest of the model the same. This can be seen in Figure \ref{fig:distance}. Figure \ref{fig:distance} shows the normalized\footnote{We first take the L2 norm between original and post-edit weights of the layer being edited, and then normalize it by number of neurons in the layer.} L2 distance between the weights of the edited layer and the original weights of the layer as a function the number of edits made to the model. We find that as more edits are made to the model, the distance between the original and edited layer continuously increases until it suddenly explodes. This is the point where the edited layer becomes incompatible with the rest of the model. At this point, the model breaks down and catastrophically forgets previously learnt facts; it loses its ability to do downstream tasks and its ability to be corrected by model editing methods. The gradual increase in distance between the original weight and new weights leads to the gradual region of forgetting, whereas the spike in the distance with a single updates leads to catastrophic forgetting.  %This could be a result of either continuous sequential editing that accumulates over time or one specific fact that it especially hard for a model to learn.

\subsubsection{Disabling Edits in ROME}

Finally, we take a deeper look at the specific edits that cause the inflection point in ROME. We call these edits \textbf{\textit{disabling edits}}, as they disable the model and make it unusable for downstream tasks. Note that these are facts that ROME is successfully able to edit in the model. Are these disabling edits a result of continuous sequential editing that accumulates over time or of one specific fact that is especially hard for a model to learn? 

We find that when we edit the facts corresponding to disabling edits as the first edit made to the model, the model is still left completely disabled, and the normalized distance of the layer weights from the original weights is comparable to the distances seen around the spikes. This can be seen in Table \ref{table:disabling}, where we present the average normalized L2 norm between the edited model layer and its original weights when only one edit is made to the model. We find that disabling edits have three orders of magnitude larger distance than non-disabling edits. This shows that the disabling edits in ROME are not a result of continuous sequential editing of the model, but a fundamental limitation of ROME. We can describe disabling edits as facts that ROME is unable to successfully edit without crippling the model. Such disabling edits can also be a source of potential adversarial attacks.

%We also want to point out that while existence of such disabling edits are a fundamental limitation of ROME, it can also be used as a feature. If we can test the editability of a fact prior to editing by checking if it is a disabling fact, we can possibly prevent catastrophic forgetting when using ROME. Such explorations are beyond the scope of this paper and should be part of future research. 

\begin{table*}
\centering

\scalebox{0.9}{
\begin{tabular}{c|c|c|c}
%\hline
\textbf{Property} & \textbf{FT-C} & \textbf{ROME} & \textbf{MEMIT}   \\ \hline
EDITING EFFICACY & 100\% & 100\% until CF & < 100\% until CF  \\
EDIT LOCALITY & Very Low & High & Very High  \\
AVERAGE DURATION BEFORE CF & CF not observed & Short & Long\\
DOWNSTREAM PERFORMANCE LOSS & High & High & Low \\
FACT FORGETTING PERCENTAGE & Very High & High & Low \\
SINGLE DISABLING EDIT & False & True & False\\
%\hline
\end{tabular}
}
\caption{Comparison between ROME and MEMIT at scale. CF refers to the point of catastrophic forgetting.}
\label{table:properties}
\end{table*}

\section{Scaling MEMIT}
In this section, we evaluate the performance of the Mass-Editing Memory in a Transformer (MEMIT) method \cite{MEMIT} when multiple sequential edits are made to the same model. We will follow the same procedure as followed in section \ref{sec:rome}, first evaluating the editing proficiency, fact forgetting and loss of performance on downstream tasks. We perform sequential editing on GPT-J (6B) using a random subset of 2000 examples from the CounterFact dataset when using MEMIT. This subset is a continuation of sample 1 in section \ref{sec:rome}. %The behavior of MEMIT when sequentially editing GPT2-J is shown in Figure \ref{fig:memit_gptj}. %For the first time we see different editing behaviors on different models, showing that the algorithms 

Figure \ref{fig:memit_gptj:edit_score} shows the editing proficiency of MEMIT as a function of the number of edits made to the model. Note that here we edit one fact at a time for MEMIT. While MEMIT is able to make batched edits, we leave that analysis for future work. The dotted lines show a window size of 5 previous edits, whereas solid lines show a window size of 50 previous edits, same as in Figure \ref{fig:editing_proficiency:ROME}. We see that the efficacy score for MEMIT is not as high as ROME. \textbf{This means that knowledge edits made via MEMIT are not always successful, while in ROME we're always able to edit facts successfully}. We also see a continuous decline of neighborhood score for MEMIT as seen for ROME, showing that editing facts also start affecting other facts stored in the model. 

\begin{figure}
    \centering
    \includegraphics[width=0.75\linewidth]{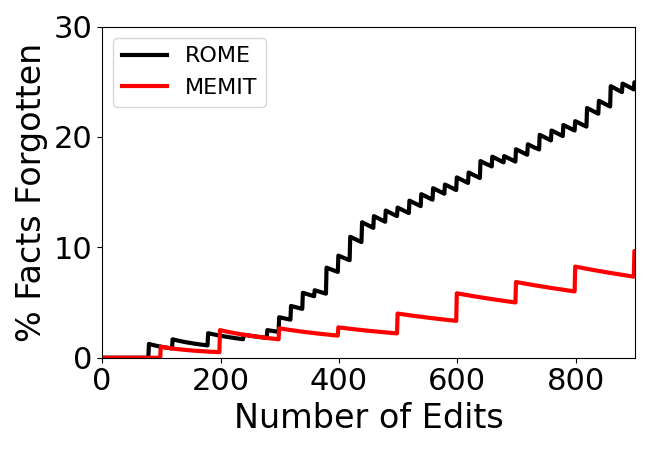}
    \caption{Compares the forgetting rate between ROME and MEMIT.}
    \label{fig:memit-rome_forgetting}
\end{figure}

Figure \ref{fig:memit_gptj:forgetting} shows the percentage of successfully edited facts that get forgotten as new facts are edited using MEMIT. We again begin to see two phases of forgetting - gradual and catastrophic. The catastrophic forgetting phase begins after approximately 1400 edits made to the model. When compared to ROME, we find that edits made using MEMIT have a much longer gradual forgetting phase across multiple samples. Additionally, we also find that \textbf{MEMIT forgets fewer previously edited facts when compared to ROME}, as seen in Figure \ref{fig:memit-rome_forgetting}, where MEMIT forgets almost three times fewer facts when compared to ROME. This can also be seen for other samples in appendix \ref{sec:appendix:rome:forgetting}.

%Forgetting is evaluated in Figure \ref{fig:memit-forget} after every 500 edits beyond 2000 edits for computational reasons, which is why the plot looks spiky afterwards. For this reason, we want to clearly point out the region of gradual forgetting to the reader, which goes on until approximately 4000 edits, which is followed by the region of catastrophic forgetting after which the model forgets all previously learnt facts. When compared with the gradual forgetting region of Figure \ref{fig:forgetting}, we find that \textbf{MEMIT forgets fewer previously edited facts when compared to ROME}. This is seen across multiple samples of edits made using MEMIT, although we show only one due to space constraints. As a comparison, model edited with ROME forgets approximately 30\% facts after 600 edits in sample 1 (Figure \ref{fig:forgetting:sample_1}), whereas when editing the same sample with examples in the exact same order, MEMIT forgets approximately 10\% previously edited facts. Additionally, we don't encounter an inflection point for a larger number of edits when compared to ROME. This is true for multiple samples edited using MEMIT. This shows that although MEMIT is prone to gradual forgetting, it has a longer gradual forgetting phase, thus making it more stable when making a larger number of sequential edits to the model. 

%which means as more edits are done on the model it begins to slowly lose its general abilities, it is more robust to single edits crippling the whole model. We also see this behavior across multiple samples in MEMIT.

Finally, the effect of the number of edits on the downstream performance of the model can be seen in Figure \ref{fig:memit_gptj:downstream}. We see that the model maintains its ability to downstream tasks as more number of edits are made to the model. While this is true for GPT-J, we observe a gradual loss of performance with multiple edits for GPT2-XL (appendix \ref{sec:appendix:downstream}). In fact, we find that for GPT2-XL, the model loses the ability to do paraphrase detection long before the point of catastrophic forgetting, thus showcasing that the model can lose its ability of performing certain downstream tasks even before a disabling edit cripples the model.

Figure \ref{fig:memit_gptj:distance} shows the distance of the edited layers from the respective original layers. We find that the distance from the respective original layer increases gradually until approximately 1400 edits. After this, we find spikes in the distance between the edited layers and the original layers, which coincides with the points of catastrophic forgetting as seen in previous plots. We also evaluate if disabling edits are fundamental to MEMIT. We find that, if the fact that disables the model after a sequence of edits is edited first in the model, it does not lead to catastrophic forgetting. Thus, MEMIT is more robust to a single destabilizing edits when compared to ROME. We conjecture this is because a fact is stored within multiple layers of a model \citep{ROME, MEMIT}, and editing the weights of a single layer to edit facts can lead to larger instabilities in the model. We summarize the properties of ROME and MEMIT at scale in Table \ref{table:properties}, clearly showing MEMIT as a superior method across different parameters except editing efficacy.  

%This supports the hypothesis behind MEMIT that facts are possibly stored across multiple layers and to be able to successfully edit facts, we need to edit the weights across each of these fact storing layers. 

%\section{Limitations of ROME and MEMIT at Scale}
%We find that ROME and MEMIT have significant functional differences. ROME has almost 100\% efficacy when editing facts although the model ends up forgetting a larger number of previously edited facts as multiple edits are made sequentially. MEMIT on the other hand is unable to successfully edit all facts, but fewer facts get forgotten as multiple edits are made to the model. Both ROME and MEMIT have low specificity and edited facts consistently bleed into other knowledge stored inside the model, especially with larger number of edits made to the model. MEMIT is also more robust to a single de-stabilizing edit when compared to ROME. We conjecture this is because a fact is stored within multiple layers of a model, and editing the weights of a single layer to edit facts can lead to instability in the model. Additionally, we see that both ROME and MEMIT are prone to \textit{forgetting} in a machine learning sense, losing the ability to recall previously edited facts, to perform downstream function and the ability to be successfuly edited by these algorithms. This makes these algorithms hard to scale and not usable for practical model editing purposes.

\section{Related Work}\label{sec:related_work}
In this paper, we focus on model editing methods that modify the base language model's parameters. Some of these methods \citep{metamodel, MEND} require training a hypernetwork \cite{hypernetwork} that generates new weights for the model being edited. Other methods \citep{ROME, MEMIT, PMET} directly update specific parts of the model after locating stored facts inside it. \citet{gupta2024unified} unify these methods under the same framework called the preservation-memorization framework and enable batched editing with ROME, an algorithm they call EMMET. Other memory based model-editing methods \citep{SERAC, mquake} are not evaluated in this paper. 

While many of these methods have shown promise \cite{editing-survey}, recent work analyzing the after-effects of these editing methods have highlighted the shortcomings of these methods. Specifically, while some of these editing methods rank high on reliability, generalization and locality metrics \citep{editing-survey, MEND, SERAC, ROME, MEMIT}, the edited knowledge is not used consistently by the model. \citet{ripple-effects} propose a new evaluation system where the "ripple effects" or implications of an edited fact are evaluated. An example of such ripple effects would be - if an edited fact updates the president of a country to the new president, then prompting for the birthplace of the president should output the birthplace of the new president. \citet{pitfalls} extend this by introducing the concept of "knowledge conflict" and additional edit types like reverse-edits and round-edits, thus evaluating the logical consistency of model editing in more complex scenarios.

%To the best of our knowledge, model editing methods have not been evaluated on non-edit related metrics. While they are able to indicate flaws in editing methods' edit success or their impact on unrelated information, it may be that model degradation occurs prior to this analysis. We explain this phenomenon through the introduction of disabling edits. Furthermore, it is important that model editing methods are evaluated on functional tasks, since in practice we want to utilize edited models. In this paper, we shed light on this issue through the introduction of downstream evaluation tasks. For a detailed description of these tasks, reference \ref{sec:appendix:downstream}. 

\section{Conclusion}
In this paper we analyze popular model editing techniques at scale. We use these methods to make multiple sequential edits to the same model and find that they fail in multiple ways. We find that ROME and MEMIT perform the best when scaled to multiple sequential edits as measured using metrics like fact forgetting and downstream performance. As we edit the models, we discover they undergo forgetting of previous knowledge and skills in two phases. Initially, the model gradually forgets previously edited facts and loses the ability to do downstream tasks, a phase which we call \textit{gradual forgetting}. After that, the model abruptly loses all coherence and function including the ability to recall previously edited facts, perform downstream tasks and the ability to be edited, which is a realization of \textit{catastrophic forgetting}. We also find that the source of these two phases of forgetting is that the layers being edited with these methods slowly drift away from their original weight values, thus becoming incompatible with the rest of the model.

%The above findings indicate serious limitations of current state-of-the-art model editing methods at scale.

%Thus, it becomes essential to evaluate post-edit models for remembering previously edited facts as well as downstream performance. 

Practical use of model editing requires us to be able to make multiple sequential edits to a model. We find that these model editing methods, like other fine tuning techniques, are prone to catastrophic forgetting. To be able to scale such methods, we not only need to have high efficacy, specificity and generalization, but we also need these methods to preserve the model's existing abilities. With this paper, we call for an improved evaluation of model editing techniques at scale, including evaluating model performance on downstream tasks and ability to recall previously edited facts. 

Finally, we want to stress upon the implications of the two phases of forgetting discovered in this paper for ROME and MEMIT. Gradual forgetting makes model editing techniques increasingly less effective as we sequentially edit facts, and hence limits their usefulness at scale. While catastrophic forgetting, which renders the model practically useless, caps the extent to which we can scale these methods. Thus we need to create model editing techniques that can counteract both gradual forgetting and catastrophic forgetting when scaled. 

\section{Limitations}
The aim of our work is to present the efficacy of current model editing techniques at scale and the usefulness of our proposed evaluation framework when studying model editing techniques at scale. To do so, in this paper we study models of size 1.5 billion and 6 billion parameters, which are standard models used in previous works \citep{MEND, ROME, MEMIT}. While we see consistent behavior of all model editing methods for the two sizes, it is possible that as models grow even larger, they respond differently to different model editing techniques. Additionally, some model editing methods like MEND \cite{MEND} and MEMIT \cite{MEMIT} have the ability to perform batched edits, that is, make multiple edits is one gradient update. Effects of model editing techniques on larger model sizes, batch edits with increasing batch sizes, as well combining multiple batches of edits sequentially are not presented in this paper. We find that these aspects of model editing are a natural extension of our work but were out of scope for this paper due to space constraints. Yet these settings can be easily evaluated under the framework we have presented in this paper and have been left for future work.

%The existence of disabling edits means that we cannot scale these methods beyond the point of catastrophic forgetting, irrespective of how good these methods are before this point. 

%Finally, we look at the source of such disabling edits.

%Crippling facts. Disabling edits.

%Is there anything special about the fact that caused the spike? If I edit the model from scratch again will it cause the spike?

\bibliography{custom}

\clearpage

\appendix
\section{Appendix}

\subsection{zsRE Compatibility}\label{sec:appendix:zsre}
Two popular datasets are used to evaluate model editing performance - zsRE \cite{zsre} and CounterFact \cite{ROME}. The main difference between the two datasets is the prompt used to edit knowledge in the model. zsRE contains prompts in a question-answer (QA) format, as shown in Table \ref{table:dataset}, whereas CounterFact contains prompts in a text completion format. Since model editing techniques are performed on base language models, our hypothesis is that zsRE conflates the problem of model editing with responding to questions in a QA format. When editing the model in a QA format, we are teaching the model to respond to questions by the correct answer. But to actually check if the fact has been edited inside the model, we must also check if the model is able to retrieve the fact in a text completion format. Otherwise all we've done is train a QA model and not edited knowledge. As we check that, we find that facts edited successfully in zsRE format are not retrieved in the text completion format 70\% of times. Some failure examples are given below (we only show examples that were successfully edited using ROME in GPT2-XL):

\begin{itemize}
    \setlength\itemsep{-0.5em} 
    \item \textbf{zsRE Question:} The date of birth of Martha Neumark is?
    \item\textbf{Edited Answer:} 1904
    \item\textbf{Completion Prompt:} Martha Neumark was born on
    \item\textbf{Generated Answer:} Martha Neumark was born on April 15, 1869, in New York City.
\end{itemize}
\begin{itemize}
    \setlength\itemsep{-0.5em} 
    \item \textbf{zsRE Question:} The college Herb Pomeroy attended was what?
    \item\textbf{Edited Answer:} Harvard University
    \item\textbf{Completion Prompt:} Herb Pomeroy attended the college of
    \item\textbf{Generated Answer:} Herb Pomeroy attended the college of Oxford University
\end{itemize}

\begin{table*}[h]
\centering
\begin{tabular}{{p{1cm}|p{13cm}}}

Task & Few-Shot Prompt\\ \hline

SST-2 & \texttt{Review : excruciatingly unfunny and pitifully unromantic
\newline
Sentiment : negative
\newline
\newline
Review : rich veins of funny stuff in this movie
\newline
Sentiment : positive
\newline
\newline
Review : by far the worst movie of the year
\newline
Sentiment : negative
\newline
\newline
Review : fashioning an engrossing entertainment out
\newline
Sentiment : positive
\newline
\newline
Review : {INPUT SENTENCE}
\newline
Sentiment :}

\end{tabular}
\caption{Few-shot template used to measure downstream model performance for the SST-2 task.}
\label{table:few_shot_sst}
\end{table*}

\subsection{Model Editing Implementation Details}\label{sec:appendix:implementation}
We use the default implementations of FT-C, ROME, MEND and MEMIT for GPT2-XL and GPT-J as used by the authors of \citet{MEMIT} in \texttt{\url{https://github.com/kmeng01/memit}}. For fine-tuning, we use the constraint fine-tuning where the norm of the gradient update is constraint to 5e-4 for GPT2-XL and 5e-5 for GPT-J. These are the default hyperparameters used by the authors. 

%For training Llama2-7b, we use the default parameters used by the authors of \citet{easyedit} in \texttt{\url{https://github.com/zjunlp/EasyEdit}}. 

\subsection{Downstream Evaluation Details}\label{sec:appendix:downstream}
An important dimension to evaluate model editing, especially at scale, is to evaluate the performance of edited models on downstream performance. In this paper, we evaluate models on four tasks of the glue \cite{glue} benchmark - sentiment analysis (SST2) \cite{sst2}, paraphrase detection (MRPC) \cite{mrpc}, natural language inference (NLI) \citep{nli1, nli2, nli3, nli4} and linguistic acceptability classification \cite{cola}. 

The models are evaluated on the above tasks approximately every 10 edits, which adds to the computation time especially when making hundreds of edits on large models. Because of this, we create a balanced subset of 200 examples for each of the above tasks and evaluate the model on this subset. The model performance is measured using the F1 metric.

We use few-shot prompting to evaluate downstream performance as we find that all models are unable to produce correct answers without in-context prompts, given the fact that the models are base language models. We follow the prompt template used by \citet{fewshot1} for our models. The exact prompts used for the different tasks are shown in Tables \ref{table:few_shot_sst}, \ref{table:few_shot_mrpc}, \ref{table:few_shot_cola}, \ref{table:few_shot_nli}.

\begin{table*}[ht]
\centering
\begin{tabular}{{p{1cm}|p{13cm}}}

Task & Few-Shot Prompt\\ \hline

MRPC & \texttt{Are the sentences paraphrases of each other.
\newline
Sentence 1: Mr McDevitt has been granted control of three crucial aspects of policing in the Solomons.
\newline
Sentence 2: Mr McDevitt has been granted control of three aspects of policing by Commissioner William Morrell.
\newline
Answer: No
\newline
\newline
Are the sentences paraphrases of each other.
\newline
Sentence 1: The notification was first reported Friday by MSNBC.
\newline
Sentence 2: MSNBC.com first reported the CIA request on Friday.
\newline
Answer: Yes
\newline
\newline
Are the sentences paraphrases of each other.
\newline
Sentence 1: In 2002, Linksys overtook Cisco Systems as the leading wireless equipment vendor, accounting for 14.1 percent of revenue.
\newline
Sentence 2: Rolfe said Linksys overtook Cisco Systems last year as the leading supplier of WLAN equipment.
\newline
Answer: No
\newline
\newline
Are the sentences paraphrases of each other.
\newline
Sentence 1: "The anticipated global sales improvement in the month of June did not materialize", said Chief Financial Officer Robert Rivet.
\newline
Sentence 2: "The anticipated global sales improvement in the month of June did not materialize as we had anticipated", the company said.
\newline
Answer: Yes
\newline
\newline
Are the sentences paraphrases of each other.
\newline
Sentence 1: That compared with \$ 35.18 million, or 24 cents per share, in the year-ago period.
\newline
Sentence 2: Earnings were affected by a non-recurring \$8 million tax benefit in the year-ago period.
\newline
Answer: No
\newline
\newline
Are the sentences paraphrases of each other.
\newline
Sentence 1: They had published an advertisement on the Internet on June 10, offering the cargo for sale, he added.
\newline
Sentence 2: On June 10, the ship's owners had published an advertisement on the Internet, offering the explosives for sale.
\newline
Answer: Yes}

\end{tabular}
\caption{Few-shot template used to measure downstream model performance for the MRPC task.}
\label{table:few_shot_mrpc}
\end{table*}

\begin{table*}[ht]
\centering
\begin{tabular}{{p{1cm}|p{13cm}}}

Task & Few-Shot Prompt\\ \hline

COLA & \texttt{Is this sentence linguistically acceptable?
\newline
Sentence : Bill pushed Harry off the sofa for hours.
\newline
Answer : No
\newline
\newline
Is this sentence linguistically acceptable?
\newline
Sentence : Bill floated down the river for hours.
\newline
Answer : Yes
\newline
\newline
Is this sentence linguistically acceptable?
\newline
Sentence : It is important for the more you eat, the more careful you to be.
\newline
Answer : No
\newline
\newline
Is this sentence linguistically acceptable?
\newline
Sentence : It is important for you to be more careful, the more you eat.
\newline
Answer : Yes
\newline
\newline
Is this sentence linguistically acceptable?
\newline
Sentence : Mary will believe Susan, and you will Bob.
\newline
Answer : Yes
\newline
\newline
Is this sentence linguistically acceptable?
\newline
Sentence : You will Bob believe.
\newline
Answer : No
\newline
\newline
Is this sentence linguistically acceptable?
\newline
Sentence : {INPUT SENTENCE}
\newline
Answer :}

\end{tabular}
\caption{Few-shot template used to measure downstream model performance for the COLA task.}
\label{table:few_shot_cola}
\end{table*}

\begin{table*}[ht]
\centering
\begin{tabular}{{p{1cm}|p{13cm}}}

Task & Few-Shot Prompt\\ \hline

NLI & \texttt{Cyrus captured Babylon without a battle, and remedied the evils done by previous Assyrian and Babylonian rulers by sending prisoners in Babylonia back to their original homelands and aiding in the restoration of temples of the gods of various nations.
\newline
question: Babylon surrendered to Cyrus without going to battle. True or False?
\newline
answer: False
\newline
\newline
Successful plaintiffs recovered punitive damages in Texas discrimination cases 53\% of the time.
\newline
question: Legal costs to recover punitive damages are a deductible business expense. True or False?
\newline
answer: True
\newline
\newline
The gastric bypass operation, also known as stomach stapling, has become the most common surgical procedure for treating obesity.
\newline
question: Obesity is medically treated. True or False?
\newline
answer: False
\newline
\newline
\{STATEMENT\}
\newline
\{Question\}
\newline
answer:}

\end{tabular}
\caption{Few-shot template used to measure downstream model performance for the NLI task.}
\label{table:few_shot_nli}
\end{table*}

\clearpage

\subsection{Additional Scaling Experiments}\label{sec:appendix:rome}

\subsubsection{Editing Proficiency}\label{sec:appendix:rome:editability}

In this section, we present plots for editing proficiency for GPT2-XL (1.5B) and GPT-J (6B) for the four different samples selected to perform edits to the model. Note that sample 1 is the sample of edits shown in the main paper. Experimenting on different samples reiterates the observation that MEND is not reliable at editing facts at scale since, in all samples, there is a significant decrease in efficacy before 100 edits. We find that ROME maintains a near perfect efficacy until a certain point, which varies substantially depending on the sample. Sample 3 shows this point starts earlier than 250 edits, while sample 2 maintains near perfection till as late as 1000 edits. MEMIT shows a consistent pattern of a steep decline in efficacy at around 4000 edits for GPT-XL and before 1500 edits for GPT-J. ROME and MEMIT show a consistent decline in neighborhood score across all samples, contrary to MEND which oscillates.

\begin{figure*}
    \centering
    \begin{subfigure}{.24\textwidth}
        \centering
        \includegraphics[width=\linewidth]{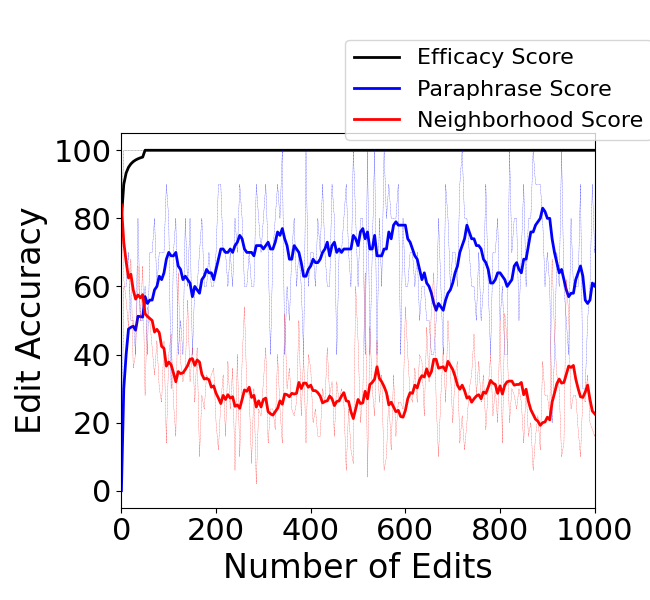}
        \caption{FT-C}
    \end{subfigure}%
    \begin{subfigure}{.24\textwidth}
        \centering
        \includegraphics[width=\linewidth]{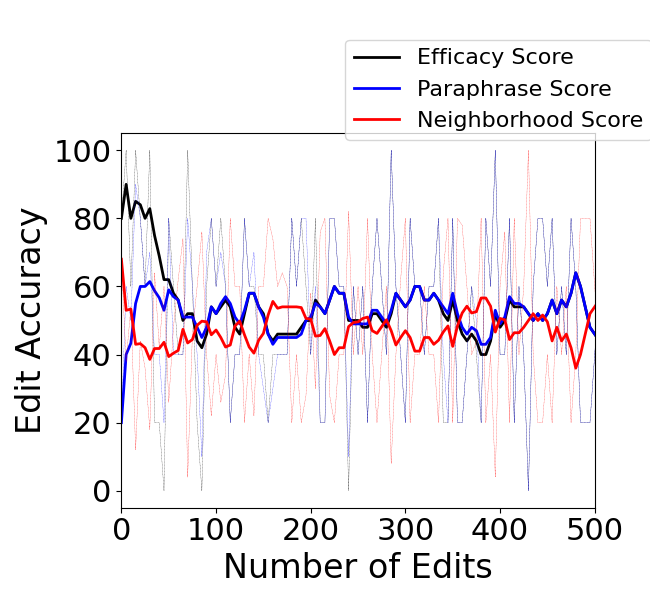}
        \caption{MEND}
    \end{subfigure}%
    \begin{subfigure}{.24\textwidth}
        \centering
        \includegraphics[width=\linewidth]{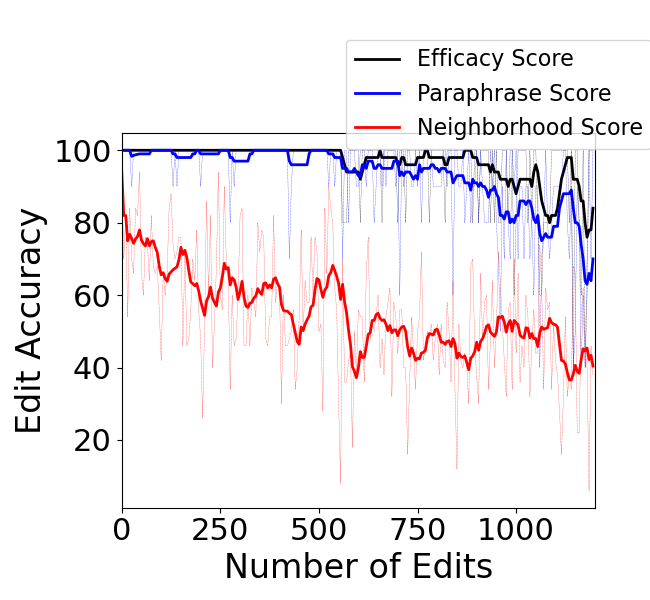}
        \caption{ROME}
    \end{subfigure}
    \begin{subfigure}{.24\textwidth}
        \centering
        \includegraphics[width=\linewidth]{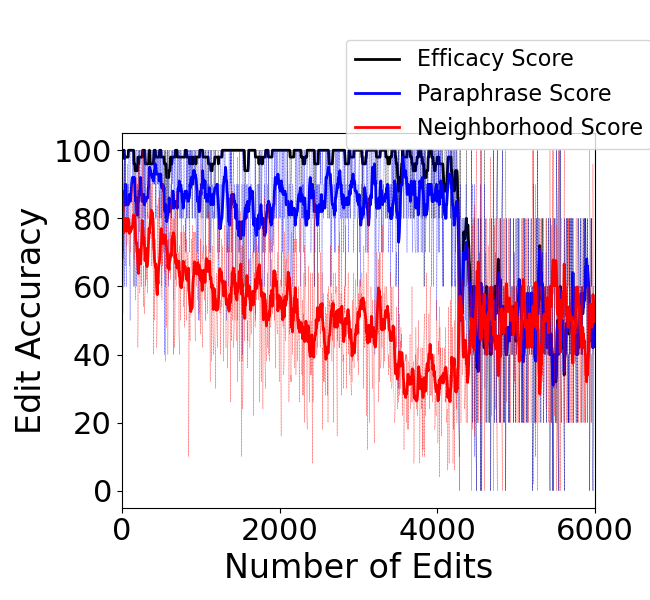}
        \caption{MEMIT}
    \end{subfigure}
    
    \caption{Editing proficiency plots for Sample 1 for GPT-XL (1.5B). }
    \label{fig:app:editing_proficiency_gpt2xl_sample1}
\end{figure*}

\begin{figure*}
    \centering
    \begin{subfigure}{.24\textwidth}
        \centering
        \includegraphics[width=\linewidth]{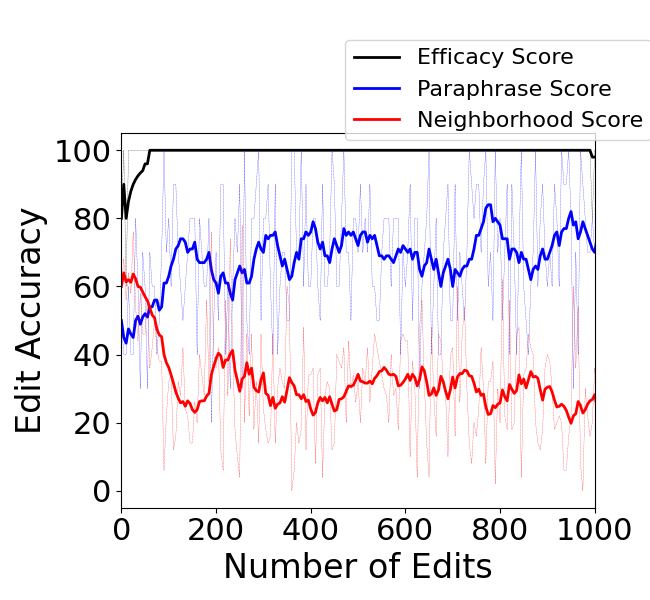}
        \caption{FT-C}
    \end{subfigure}%
    \begin{subfigure}{.24\textwidth}
        \centering
        \includegraphics[width=\linewidth]{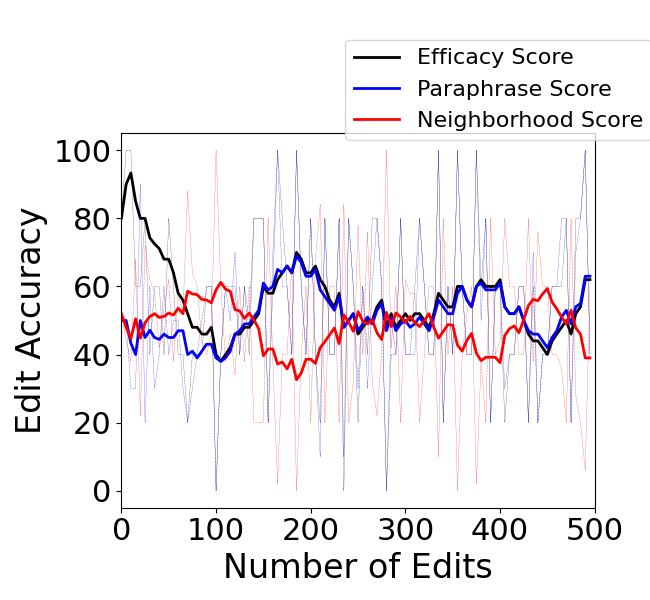}
        \caption{MEND}
    \end{subfigure}%
    \begin{subfigure}{.24\textwidth}
        \centering
        \includegraphics[width=\linewidth]{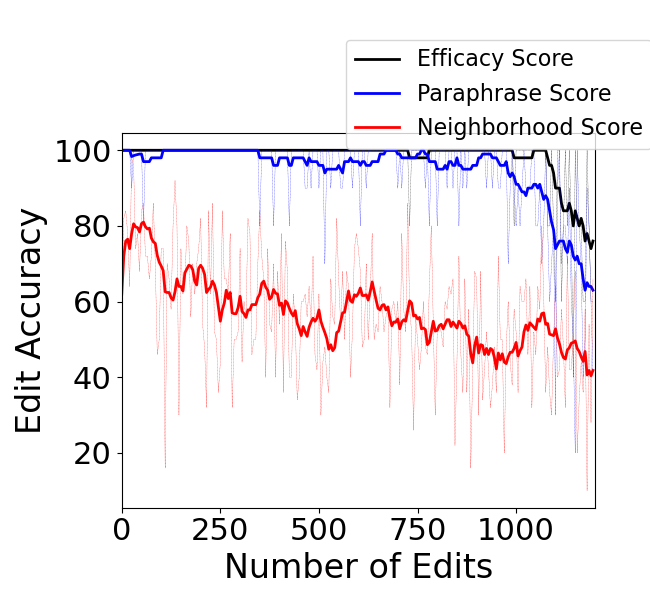}
        \caption{ROME}
    \end{subfigure}
    \begin{subfigure}{.24\textwidth}
        \centering
        \includegraphics[width=\linewidth]{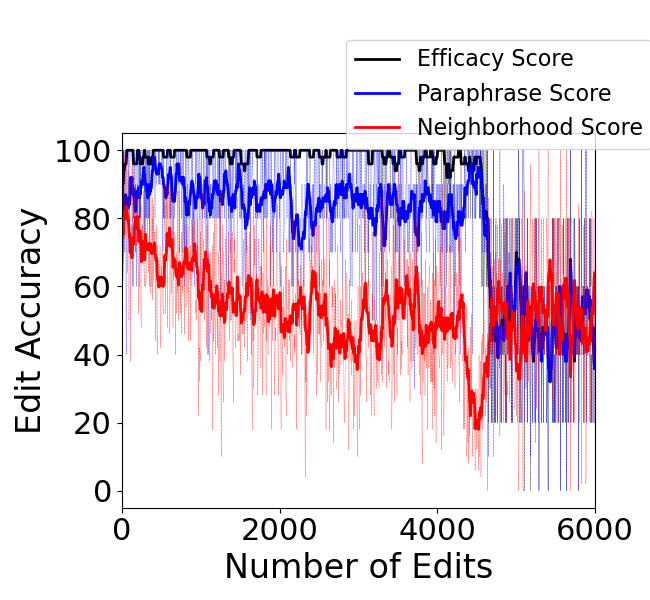}
        \caption{MEMIT}
    \end{subfigure}
    
    \caption{Editing proficiency plots for Sample 2 for GPT-XL (1.5B). }
    \label{fig:app:editing_proficiency_gpt2xl_sample2}
\end{figure*}

\begin{figure*}
    \centering
    \begin{subfigure}{.24\textwidth}
        \centering
        \includegraphics[width=\linewidth]{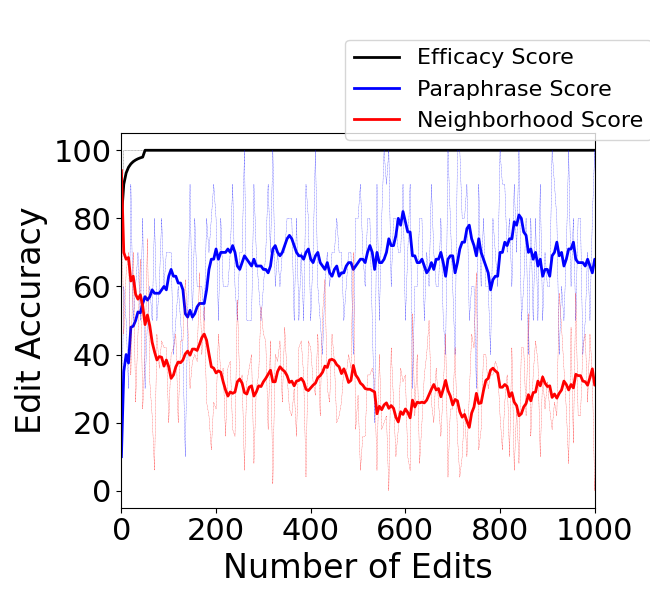}
        \caption{FT-C}
    \end{subfigure}%
    \begin{subfigure}{.24\textwidth}
        \centering
        \includegraphics[width=\linewidth]{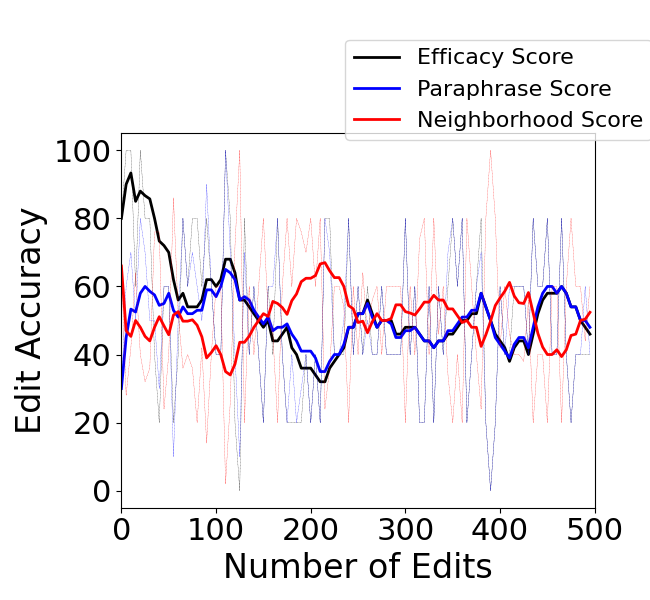}
        \caption{MEND}
    \end{subfigure}%
    \begin{subfigure}{.24\textwidth}
        \centering
        \includegraphics[width=\linewidth]{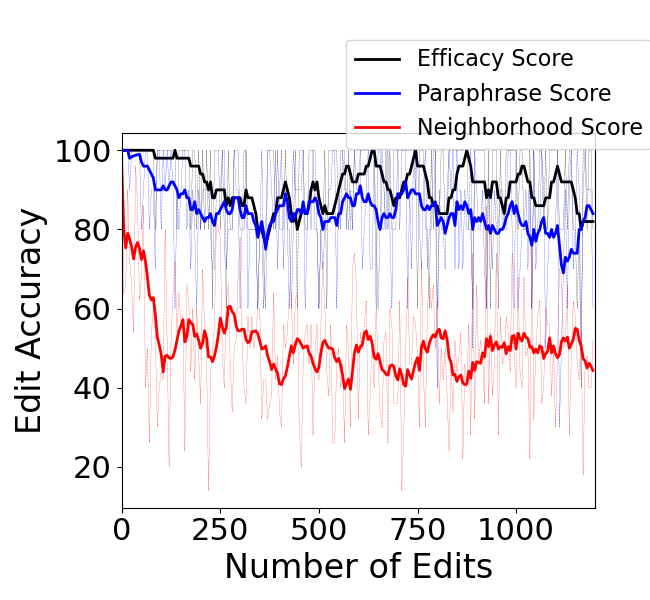}
        \caption{ROME}
    \end{subfigure}
    \begin{subfigure}{.24\textwidth}
        \centering
        \includegraphics[width=\linewidth]{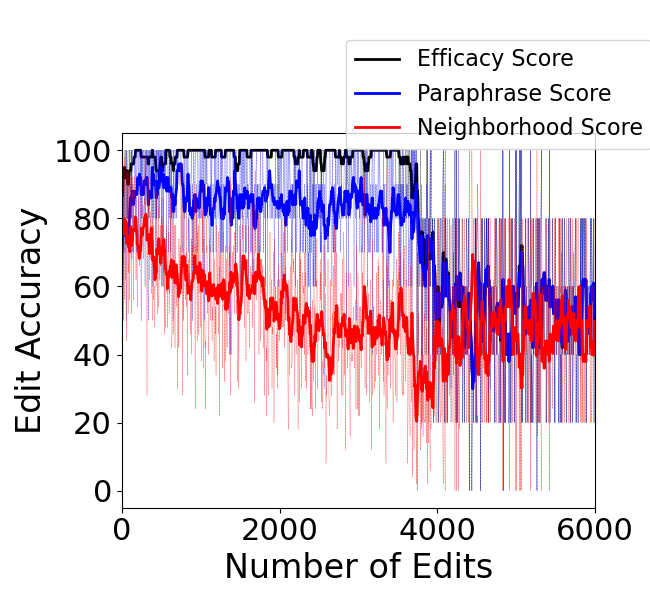}
        \caption{MEMIT}
    \end{subfigure}
    
    \caption{Editing proficiency plots for Sample 3 for GPT-XL (1.5B). }
    \label{fig:app:editing_proficiency_gpt2xl_sample3}
\end{figure*}

\begin{figure*}
    \centering
    \begin{subfigure}{.24\textwidth}
        \centering
        \includegraphics[width=\linewidth]{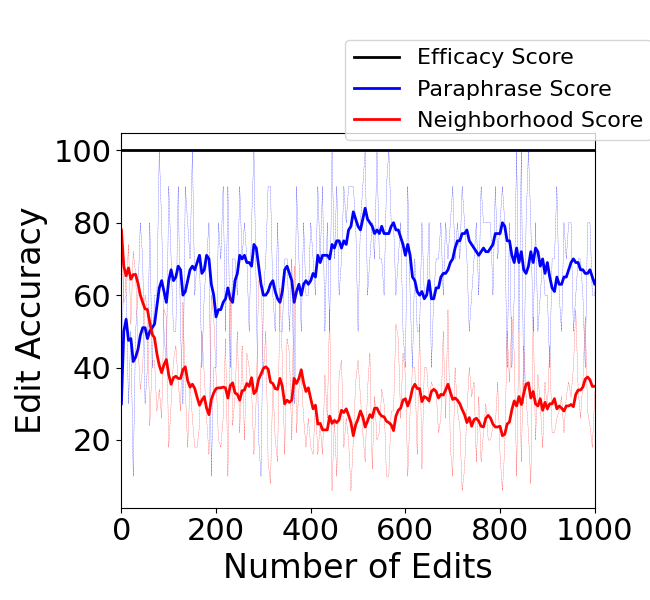}
        \caption{FT-C}
    \end{subfigure}%
    \begin{subfigure}{.24\textwidth}
        \centering
        \includegraphics[width=\linewidth]{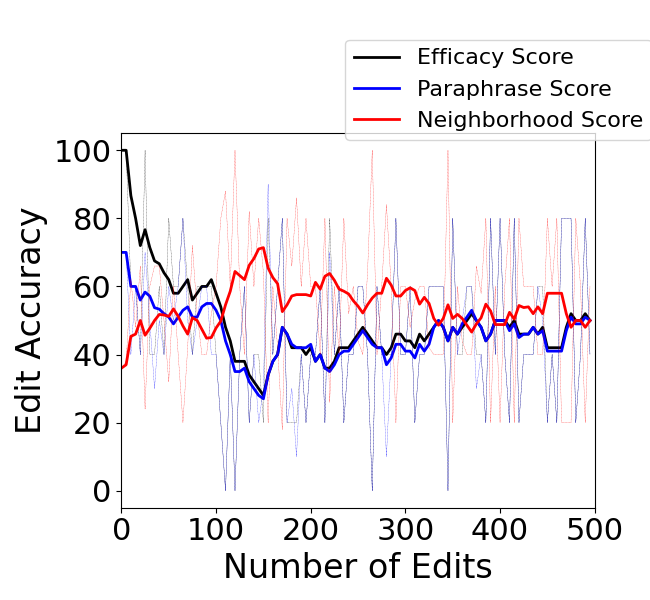}
        \caption{MEND}
    \end{subfigure}%
    \begin{subfigure}{.24\textwidth}
        \centering
        \includegraphics[width=\linewidth]{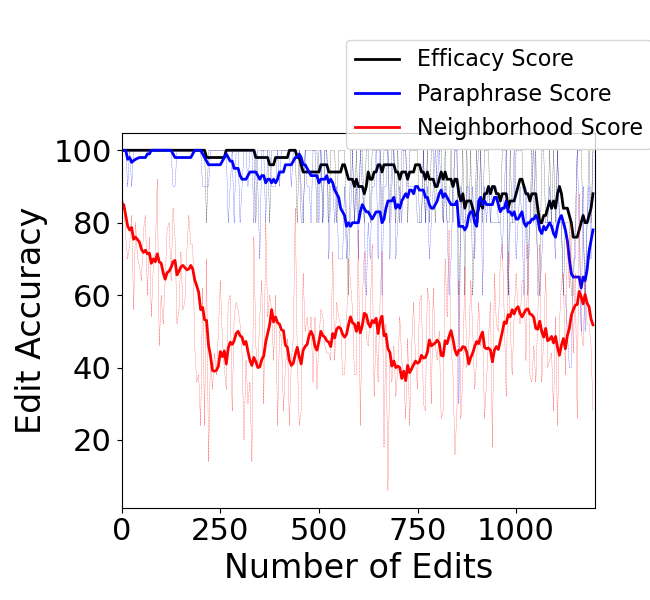}
        \caption{ROME}
    \end{subfigure}
    \begin{subfigure}{.24\textwidth}
        \centering
        \includegraphics[width=\linewidth]{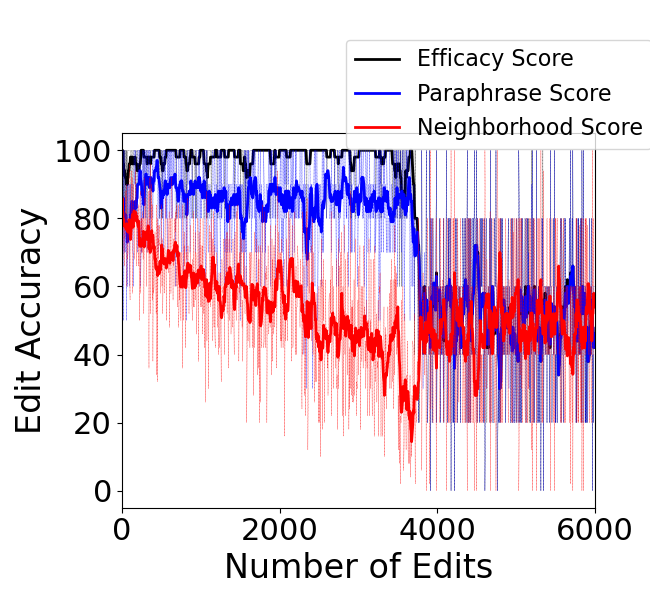}
        \caption{MEMIT}
    \end{subfigure}
    
    \caption{Editing proficiency plots for Sample 4 for GPT-XL (1.5B). }
    \label{fig:app:editing_proficiency_gpt2xl_sample4}
\end{figure*}

%%%%%%%%%%%%%%%%%%END OF GPT2XL PLOTS

\begin{figure*}
    \centering
    \begin{subfigure}{.24\textwidth}
        \centering
        \includegraphics[width=\linewidth]{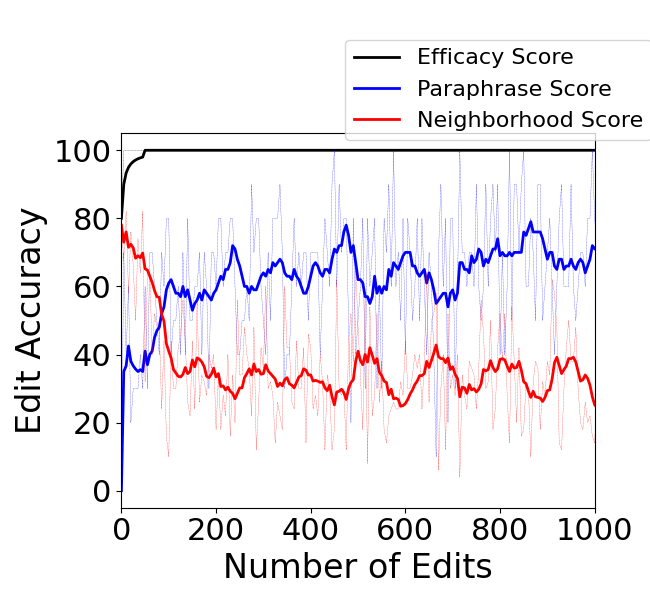}
        \caption{FT-C}
    \end{subfigure}%
    \begin{subfigure}{.24\textwidth}
        \centering
        \includegraphics[width=\linewidth]{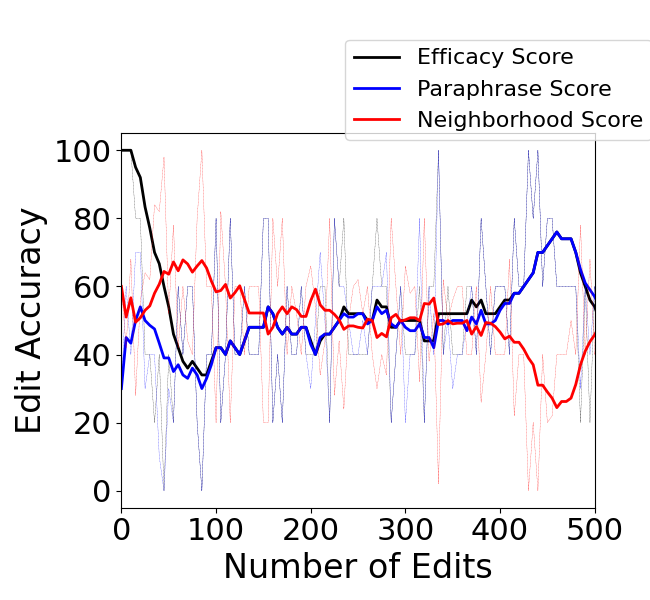}
        \caption{MEND}
    \end{subfigure}%
    \begin{subfigure}{.24\textwidth}
        \centering
        \includegraphics[width=\linewidth]{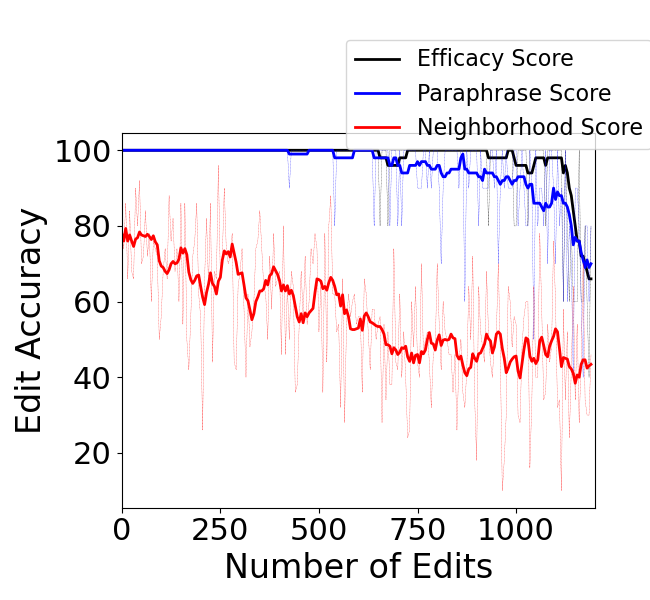}
        \caption{ROME}
    \end{subfigure}
    \begin{subfigure}{.24\textwidth}
        \centering
        \includegraphics[width=\linewidth]{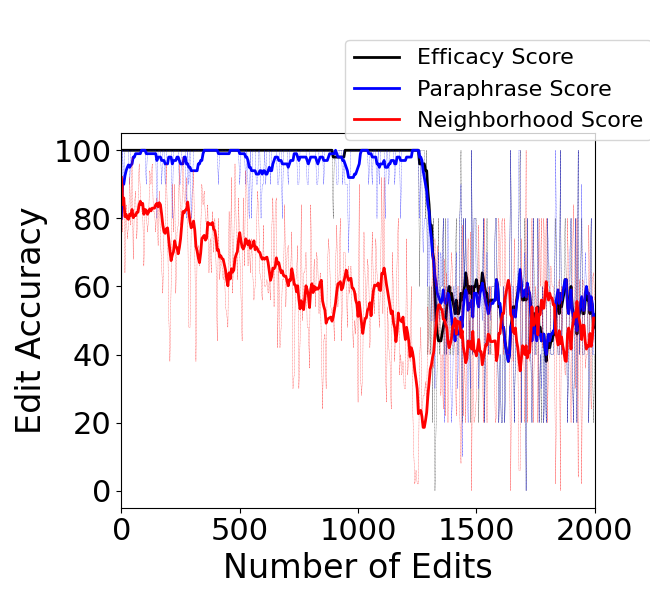}
        \caption{MEMIT}
    \end{subfigure}
    
    \caption{Editing proficiency plots for Sample 1 for GPT-J (6B).}
    \label{fig:app:editing_proficiency_gptj_sample1}
\end{figure*}

\begin{figure*}
    \centering
    \begin{subfigure}{.24\textwidth}
        \centering
        \includegraphics[width=\linewidth]{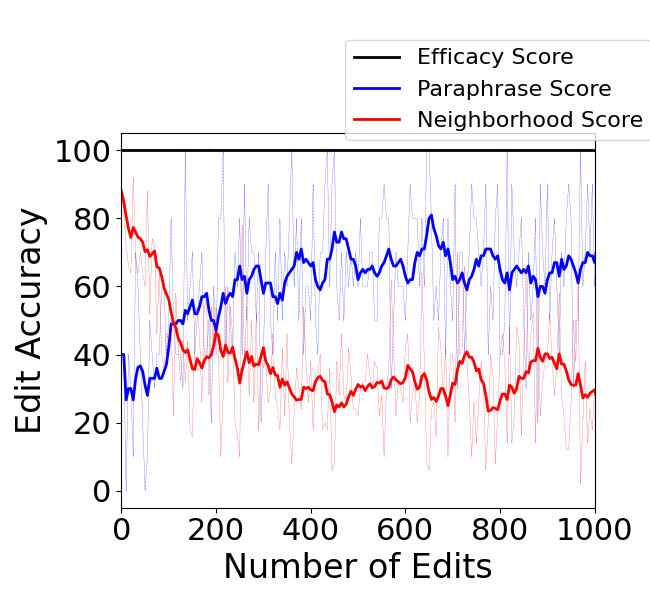}
        \caption{FT-C}
    \end{subfigure}%
    \begin{subfigure}{.24\textwidth}
        \centering
        \includegraphics[width=\linewidth]{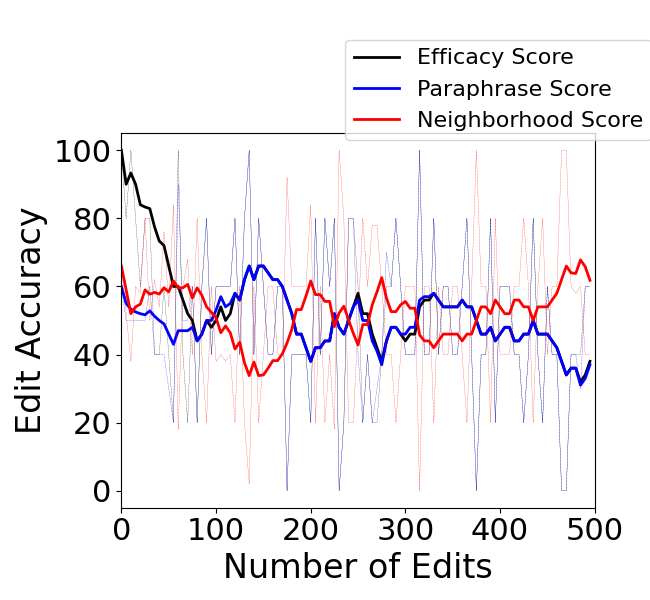}
        \caption{MEND}
    \end{subfigure}%
    \begin{subfigure}{.24\textwidth}
        \centering
        \includegraphics[width=\linewidth]{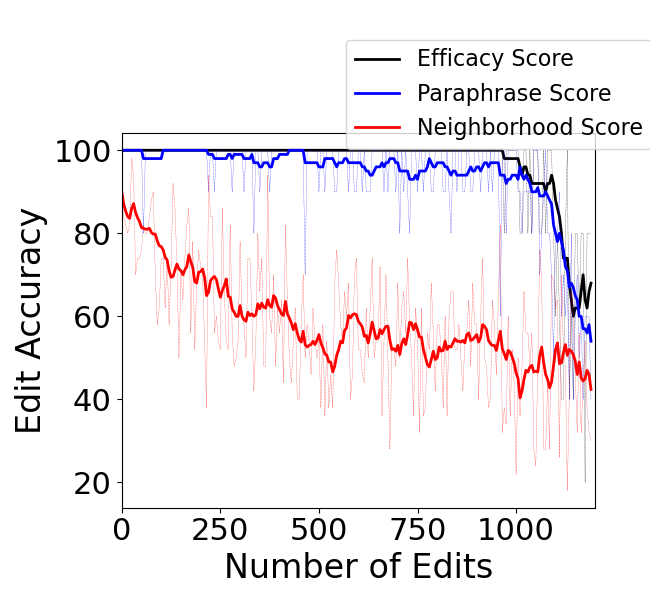}
        \caption{ROME}
    \end{subfigure}
    \begin{subfigure}{.24\textwidth}
        \centering
        \includegraphics[width=\linewidth]{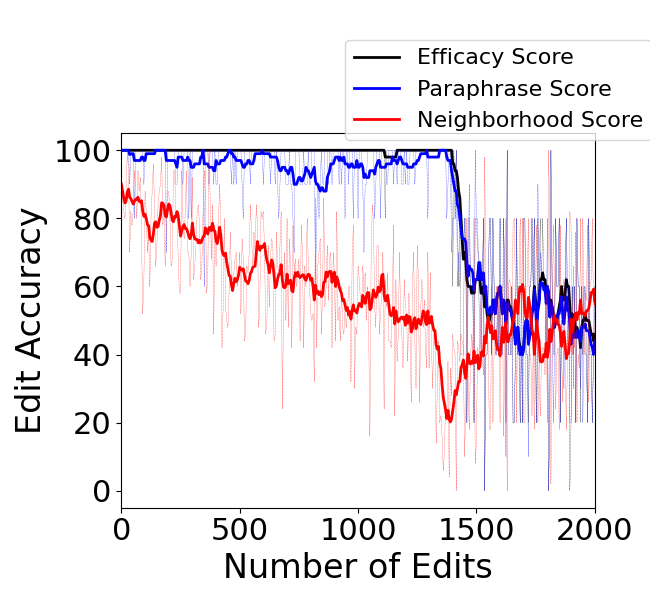}
        \caption{MEMIT}
    \end{subfigure}
    
    \caption{Editing proficiency plots for Sample 2 for GPT-J (6B).}
    \label{fig:app:editing_proficiency_gptj_sample2}
\end{figure*}

\begin{figure*}
    \centering
    \begin{subfigure}{.24\textwidth}
        \centering
        \includegraphics[width=\linewidth]{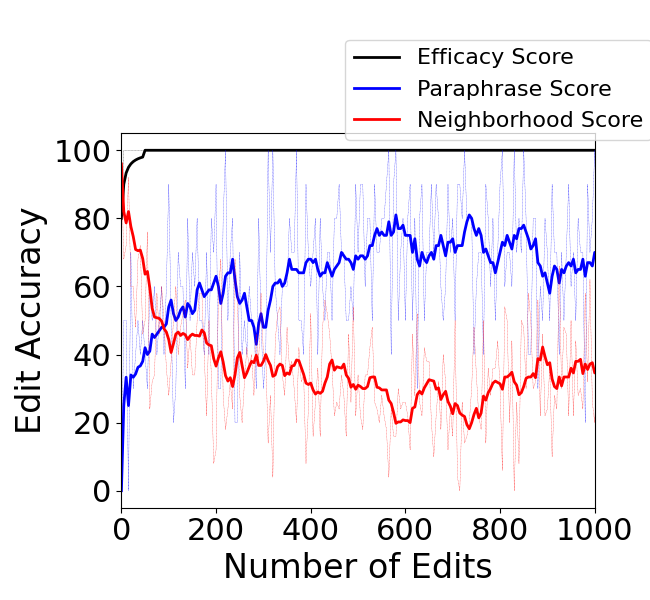}
        \caption{FT-C}
    \end{subfigure}%
    \begin{subfigure}{.24\textwidth}
        \centering
        \includegraphics[width=\linewidth]{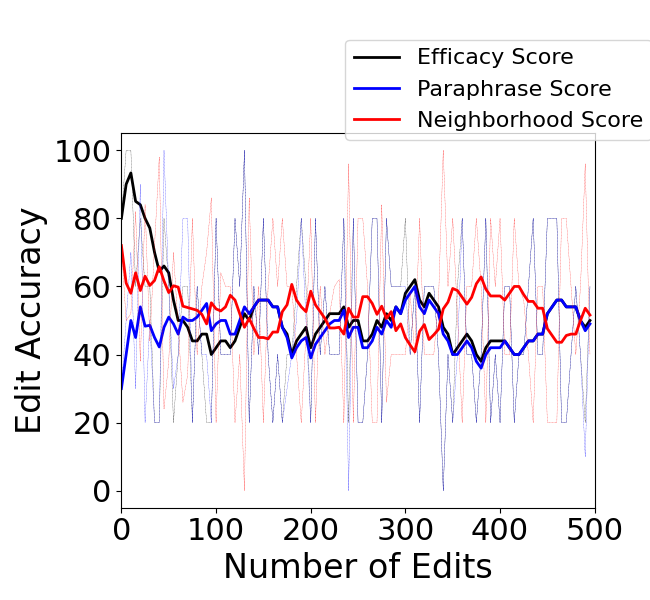}
        \caption{MEND}
    \end{subfigure}%
    \begin{subfigure}{.24\textwidth}
        \centering
        \includegraphics[width=\linewidth]{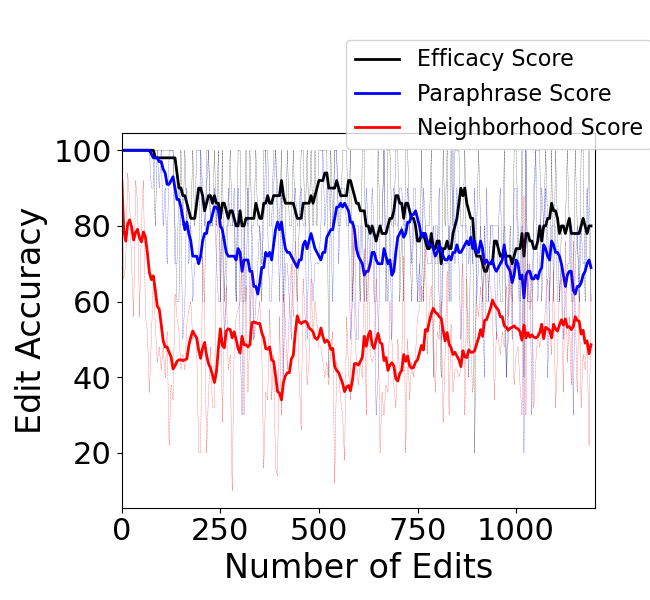}
        \caption{ROME}
    \end{subfigure}
    \begin{subfigure}{.24\textwidth}
        \centering
        \includegraphics[width=\linewidth]{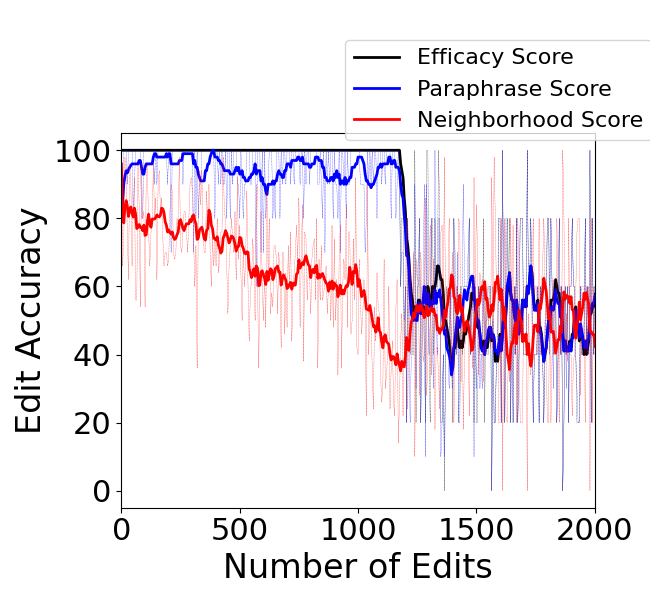}
        \caption{MEMIT}
    \end{subfigure}
    
    \caption{Editing proficiency plots for Sample 3 for GPT-J (6B).}
    \label{fig:app:editing_proficiency_gptj_sample3}
\end{figure*}

\begin{figure*}
    \centering
    \begin{subfigure}{.24\textwidth}
        \centering
        \includegraphics[width=\linewidth]{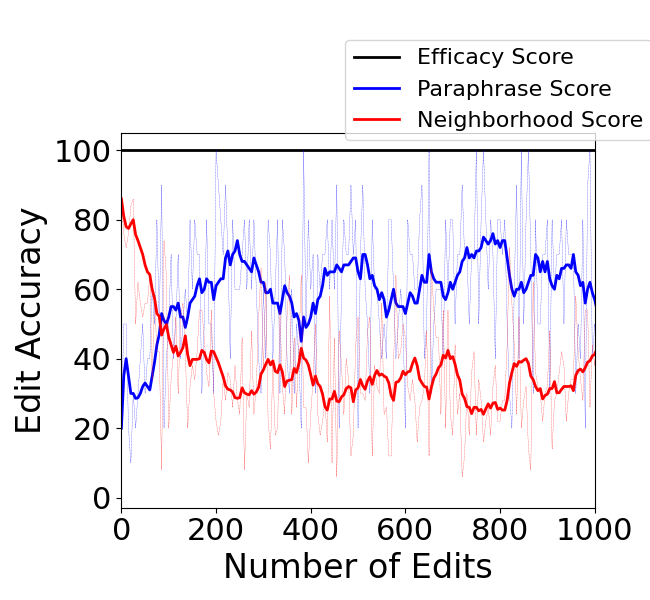}
        \caption{FT-C}
    \end{subfigure}%
    \begin{subfigure}{.24\textwidth}
        \centering
        \includegraphics[width=\linewidth]{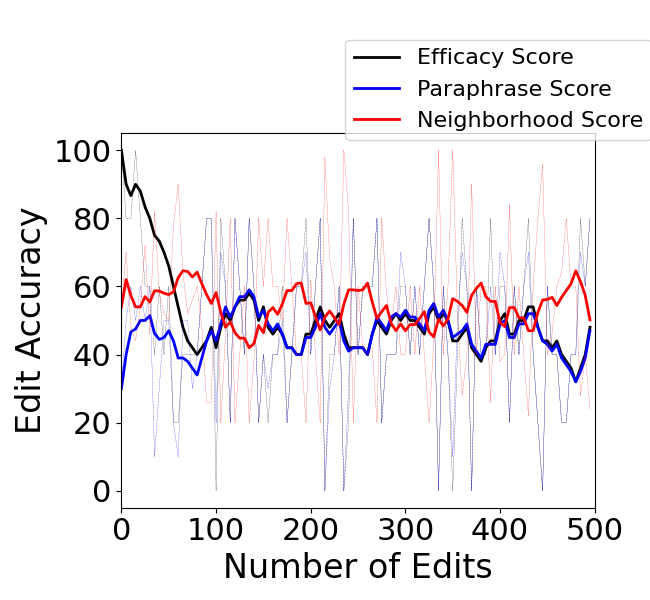}
        \caption{MEND}
    \end{subfigure}%
    \begin{subfigure}{.24\textwidth}
        \centering
        \includegraphics[width=\linewidth]{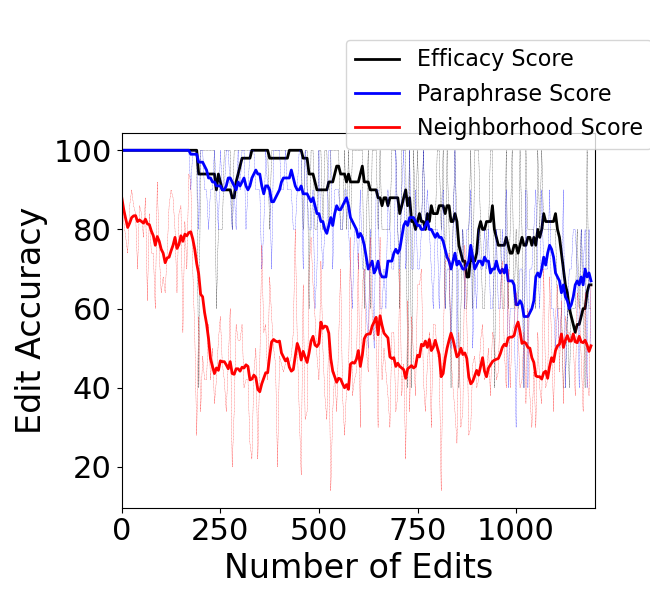}
        \caption{ROME}
    \end{subfigure}
    \begin{subfigure}{.24\textwidth}
        \centering
        \includegraphics[width=\linewidth]{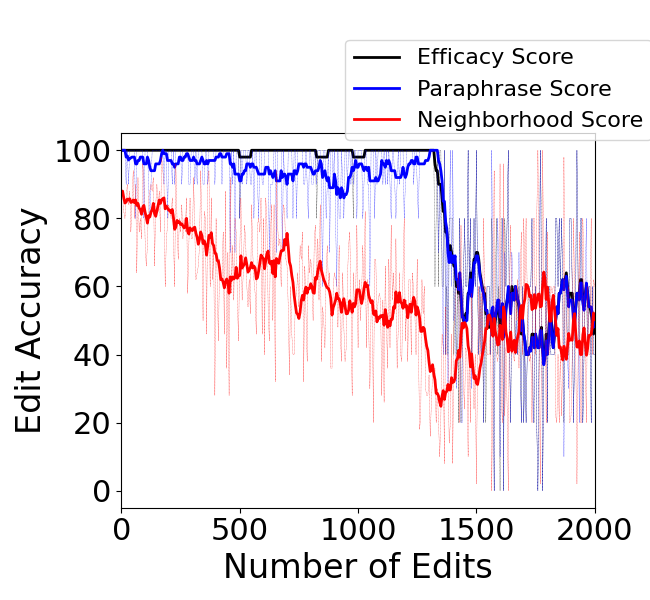}
        \caption{MEMIT}
    \end{subfigure}
    
    \caption{Editing proficiency plots for Sample 4 for GPT-J (6B).}
    \label{fig:app:editing_proficiency_gptj_sample4}
\end{figure*}

%%%%%%%%%%%%%%%%%%%%%%%%%%END OF GPTJ PLOTS

\clearpage

\subsubsection{Forgetting}\label{sec:appendix:rome:forgetting}
Here, we present plots for forgetting for both GPT2-XL(1.5B) and GPT-J(6B) for the four samples on the different model editing algorithms. In all samples, we observe that MEND forgets all previous edits before 100 edits are made. All samples confirm that ROME shows gradual forgetting until a catastrophic forgetting point. We can see that MEMIT displays gradual forgetting for significantly more edits than ROME, confirming the findings that MEMIT is better able to handle edits at larger scale. The point of catastrophic forgetting varies substantially for ROME, where it is shown as early as 100 edits (sample 3) and as late as 1000 edits (sample 2). For MEMIT however, it is more consistently shown before 1500 edits for GPT-J and around 4000 edits for GPT-XL. In both ROME and MEMIT, this catastrophic forgetting point occurs at around the same point where the efficacy score begins to decline as shown in appendix \ref{sec:appendix:rome:editability}.
\begin{figure*}
    \centering
    \begin{subfigure}{.24\textwidth}
        \centering
        \includegraphics[width=\linewidth]{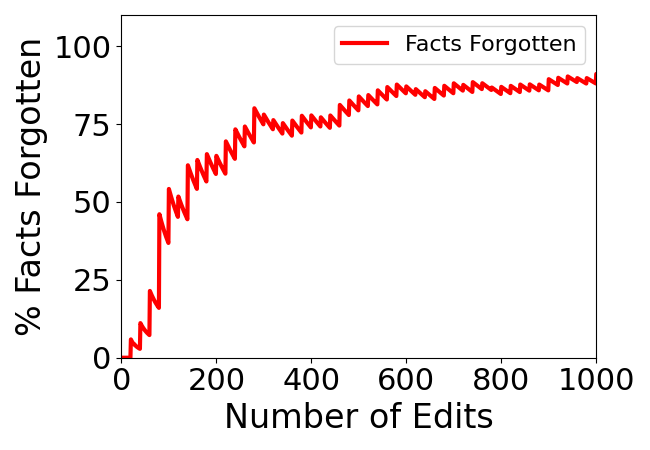}
        \caption{FT-C}
    \end{subfigure}%
    \begin{subfigure}{.24\textwidth}
        \centering
        \includegraphics[width=\linewidth]{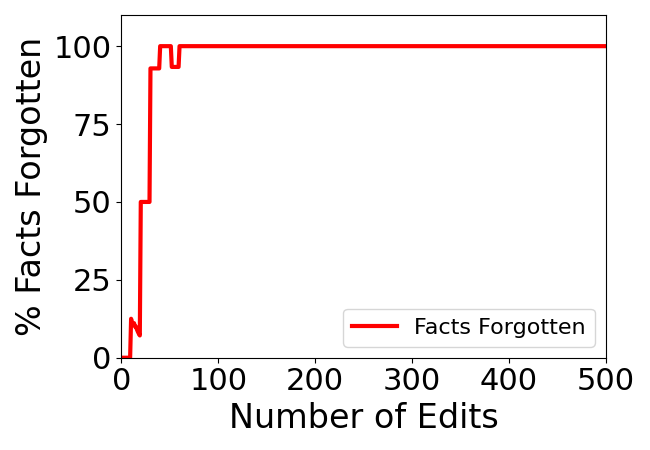}
        \caption{MEND}
    \end{subfigure}%
    \begin{subfigure}{.24\textwidth}
        \centering
        \includegraphics[width=\linewidth]{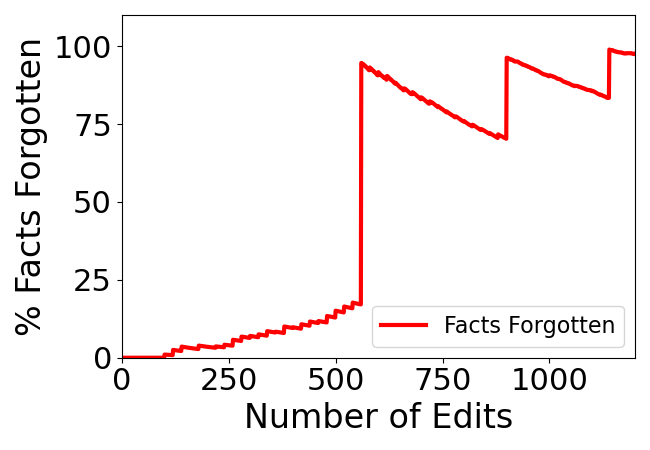}
        \caption{ROME}
    \end{subfigure}
    \begin{subfigure}{.24\textwidth}
        \centering
        \includegraphics[width=\linewidth]{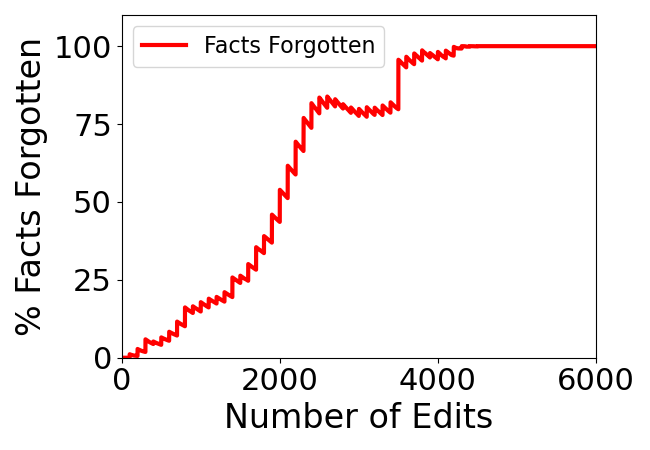}
        \caption{MEMIT}
    \end{subfigure}
    
    \caption{Forgetting plots for Sample 1 for GPT-XL (1.5B). }
    \label{fig:app:editing_proficiency_gpt2xl_sample1}
\end{figure*}

\begin{figure*}
    \centering
    \begin{subfigure}{.24\textwidth}
        \centering
        \includegraphics[width=\linewidth]{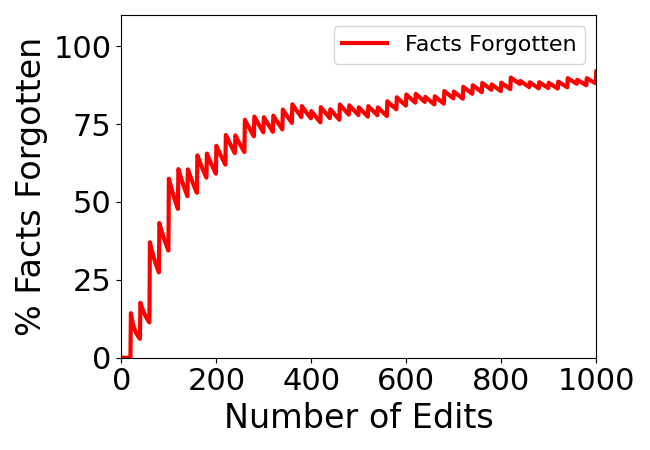}
        \caption{FT-C}
    \end{subfigure}%
    \begin{subfigure}{.24\textwidth}
        \centering
        \includegraphics[width=\linewidth]{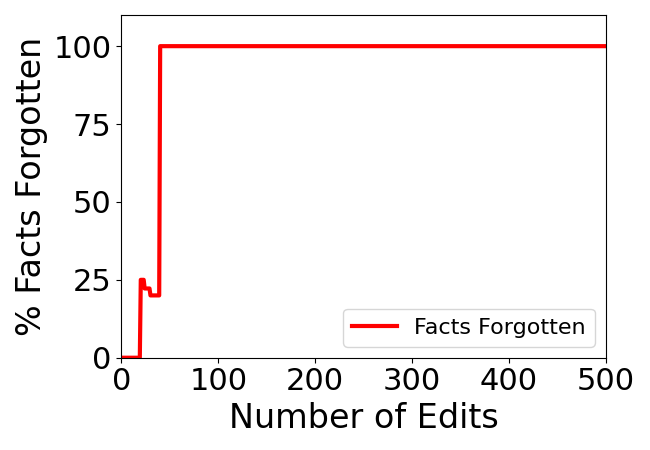}
        \caption{MEND}
    \end{subfigure}%
    \begin{subfigure}{.24\textwidth}
        \centering
        \includegraphics[width=\linewidth]{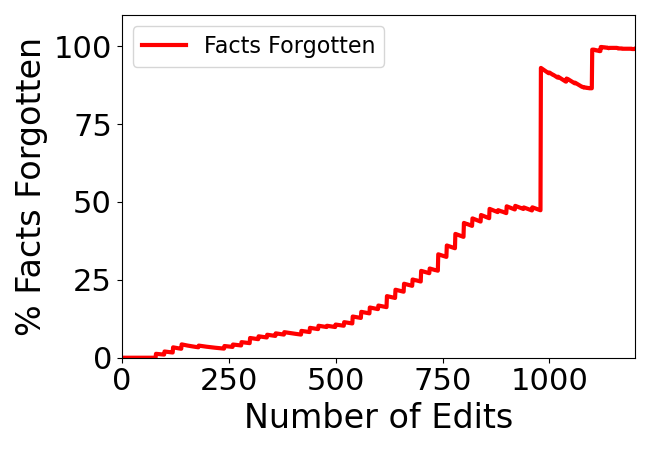}
        \caption{ROME}
    \end{subfigure}
    \begin{subfigure}{.24\textwidth}
        \centering
        \includegraphics[width=\linewidth]{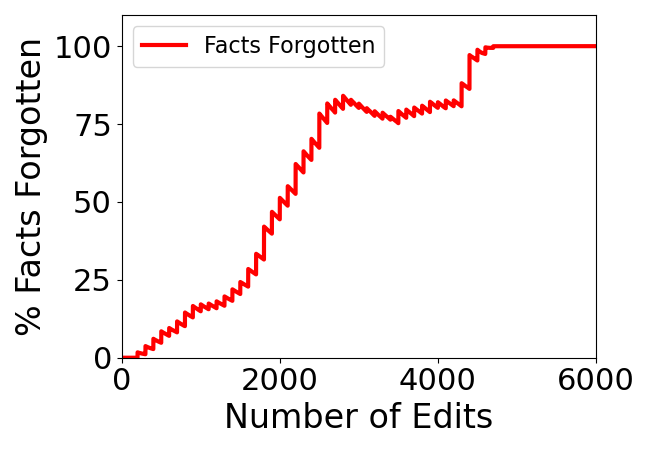}
        \caption{MEMIT}
    \end{subfigure}
    
    \caption{Forgetting plots for Sample 2 for GPT-XL (1.5B). }
    \label{fig:app:editing_proficiency_gpt2xl_sample2}
\end{figure*}

\begin{figure*}
    \centering
    \begin{subfigure}{.24\textwidth}
        \centering
        \includegraphics[width=\linewidth]{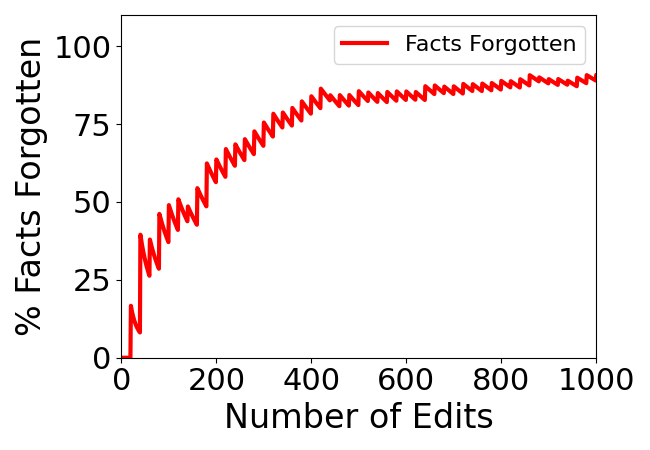}
        \caption{FT-C}
    \end{subfigure}%
    \begin{subfigure}{.24\textwidth}
        \centering
        \includegraphics[width=\linewidth]{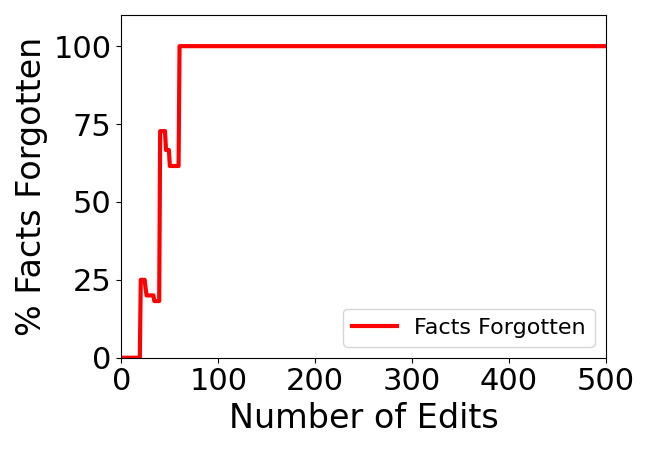}
        \caption{MEND}
    \end{subfigure}%
    \begin{subfigure}{.24\textwidth}
        \centering
        \includegraphics[width=\linewidth]{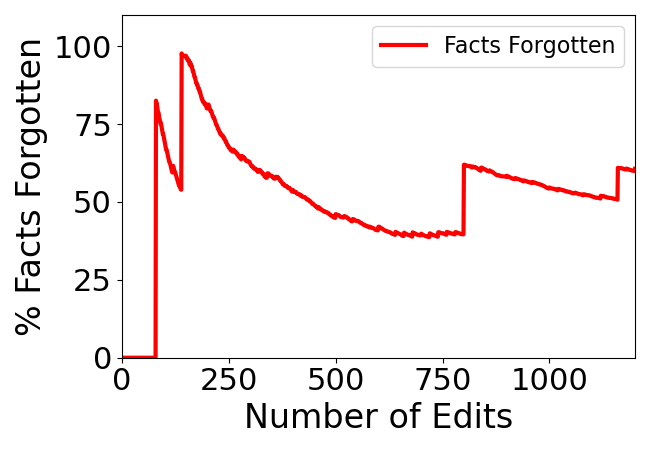}
        \caption{ROME}
    \end{subfigure}
    \begin{subfigure}{.24\textwidth}
        \centering
        \includegraphics[width=\linewidth]{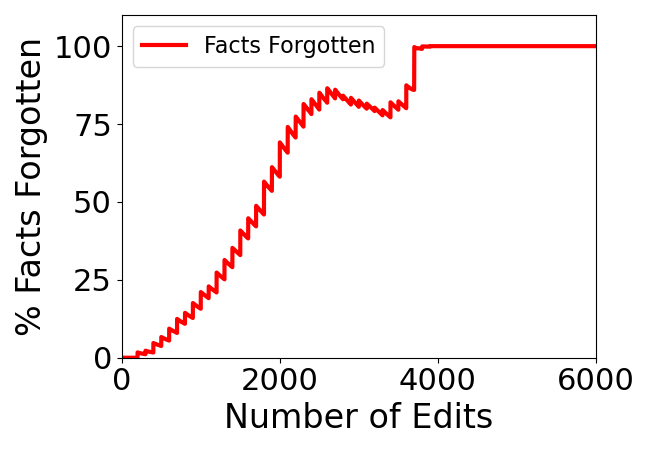}
        \caption{MEMIT}
    \end{subfigure}
    
    \caption{Forgetting plots for Sample 3 for GPT-XL (1.5B). }
    \label{fig:app:editing_proficiency_gpt2xl_sample3}
\end{figure*}

\begin{figure*}
    \centering
    \begin{subfigure}{.24\textwidth}
        \centering
        \includegraphics[width=\linewidth]{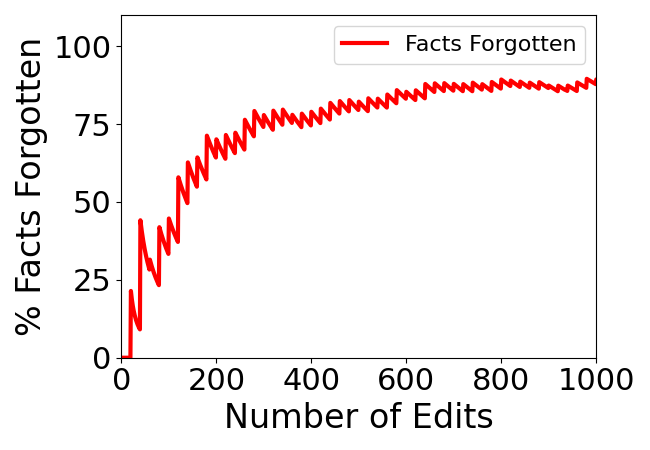}
        \caption{FT-C}
    \end{subfigure}%
    \begin{subfigure}{.24\textwidth}
        \centering
        \includegraphics[width=\linewidth]{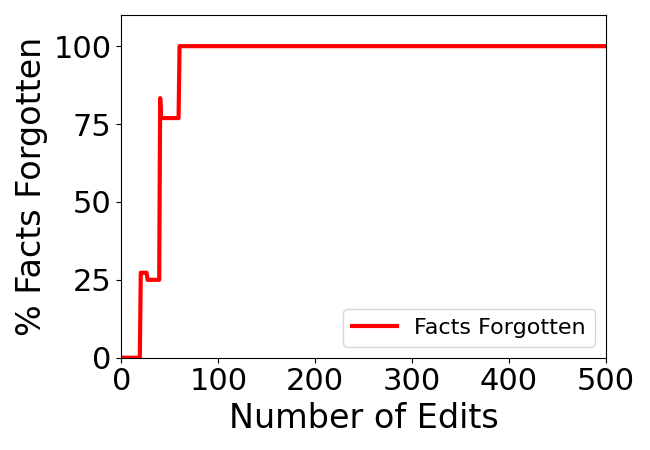}
        \caption{MEND}
    \end{subfigure}%
    \begin{subfigure}{.24\textwidth}
        \centering
        \includegraphics[width=\linewidth]{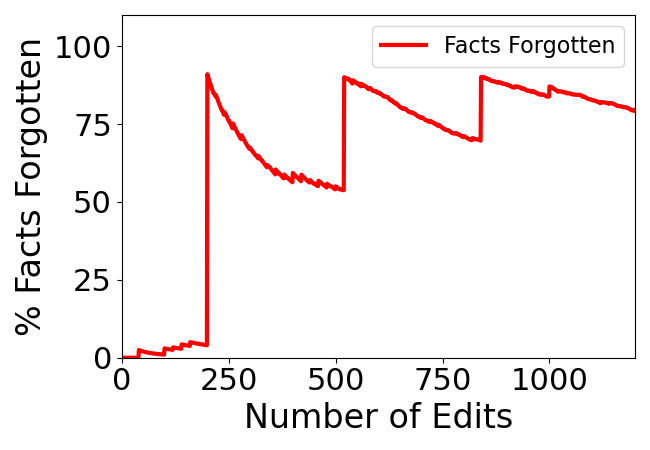}
        \caption{ROME}
    \end{subfigure}
    \begin{subfigure}{.24\textwidth}
        \centering
        \includegraphics[width=\linewidth]{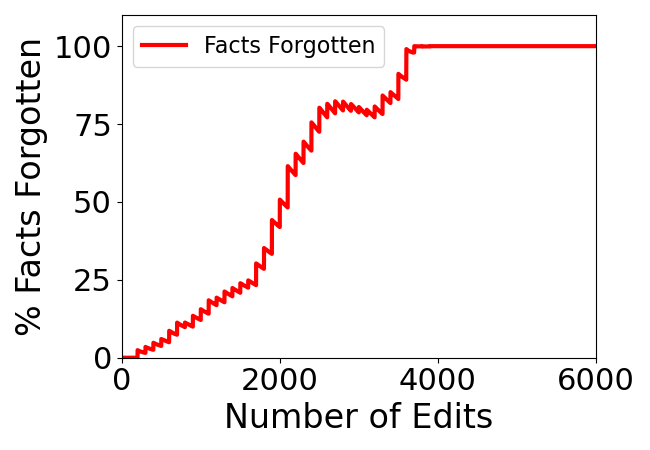}
        \caption{MEMIT}
    \end{subfigure}
    
    \caption{Forgetting plots for Sample 4 for GPT-XL (1.5B). }
    \label{fig:app:editing_proficiency_gpt2xl_sample4}
\end{figure*}

%%%%%%%%%%%%%%%%%%END OF GPT2XL PLOTS

\clearpage
\begin{figure*}
    \centering
    \begin{subfigure}{.24\textwidth}
        \centering
        \includegraphics[width=\linewidth]{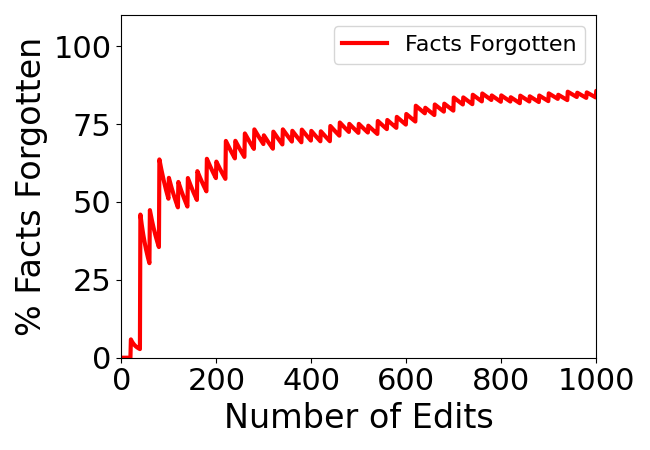}
        \caption{FT-C}
    \end{subfigure}%
    \begin{subfigure}{.24\textwidth}
        \centering
        \includegraphics[width=\linewidth]{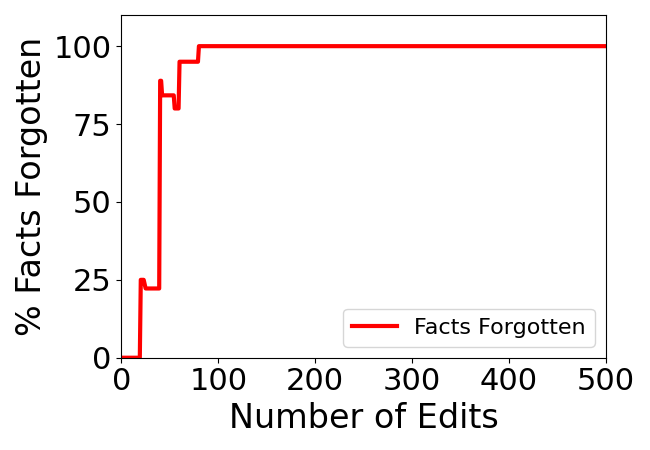}
        \caption{MEND}
    \end{subfigure}%
    \begin{subfigure}{.24\textwidth}
        \centering
        \includegraphics[width=\linewidth]{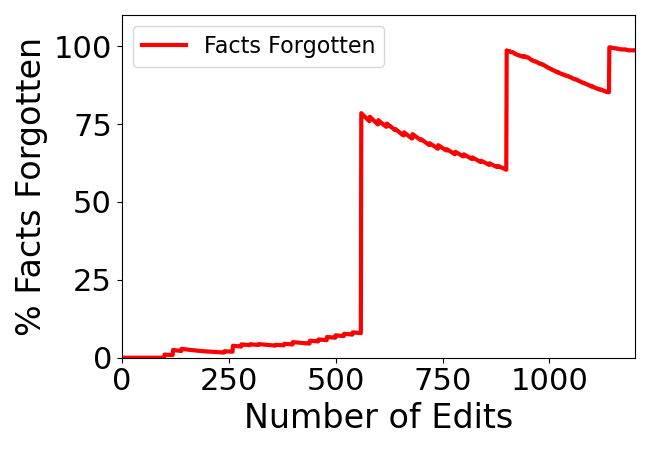}
        \caption{ROME}
    \end{subfigure}
    \begin{subfigure}{.24\textwidth}
        \centering
        \includegraphics[width=\linewidth]{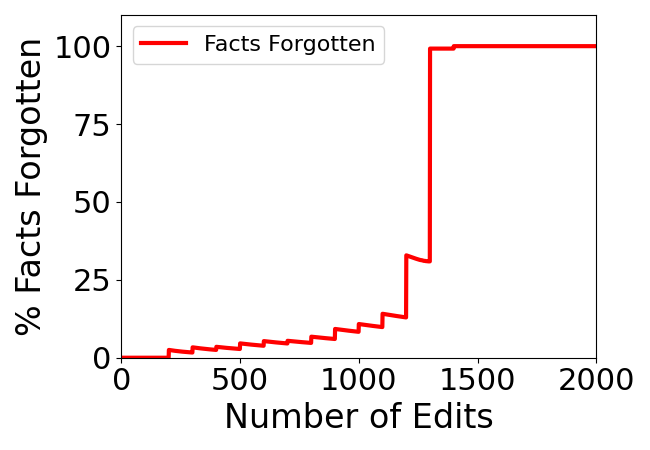}
        \caption{MEMIT}
    \end{subfigure}
    
    \caption{Forgetting plots for Sample 1 for GPT-J (6B).}
    \label{fig:app:editing_proficiency_gptj_sample1}
\end{figure*}

\begin{figure*}
    \centering
    \begin{subfigure}{.24\textwidth}
        \centering
        \includegraphics[width=\linewidth]{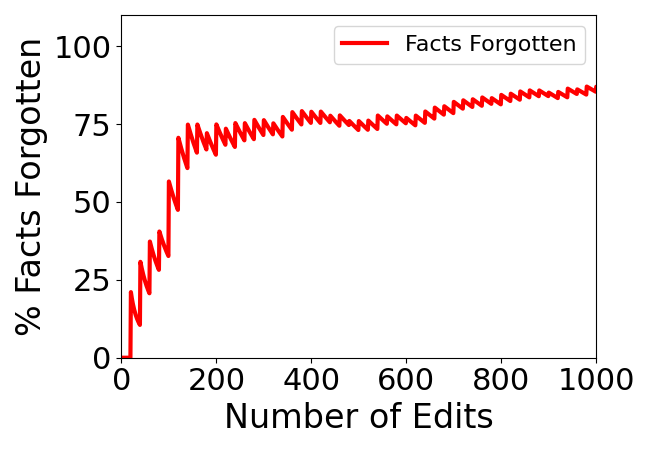}
        \caption{FT-C}
    \end{subfigure}%
    \begin{subfigure}{.24\textwidth}
        \centering
        \includegraphics[width=\linewidth]{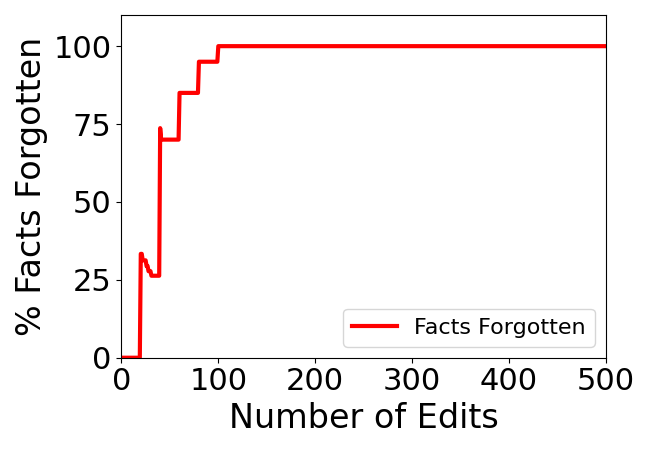}
        \caption{MEND}
    \end{subfigure}%
    \begin{subfigure}{.24\textwidth}
        \centering
        \includegraphics[width=\linewidth]{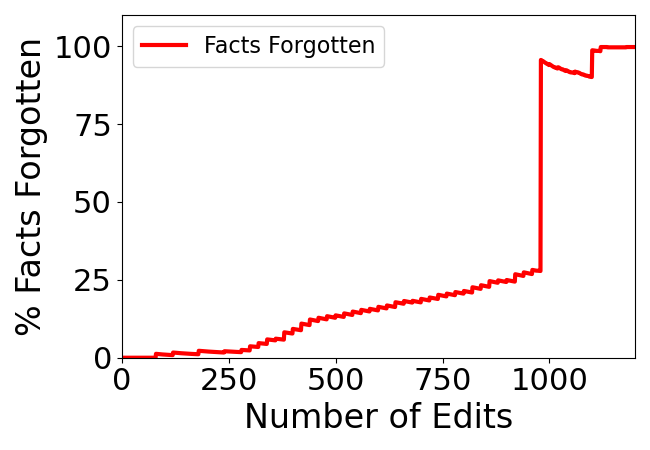}
        \caption{ROME}
    \end{subfigure}
    \begin{subfigure}{.24\textwidth}
        \centering
        \includegraphics[width=\linewidth]{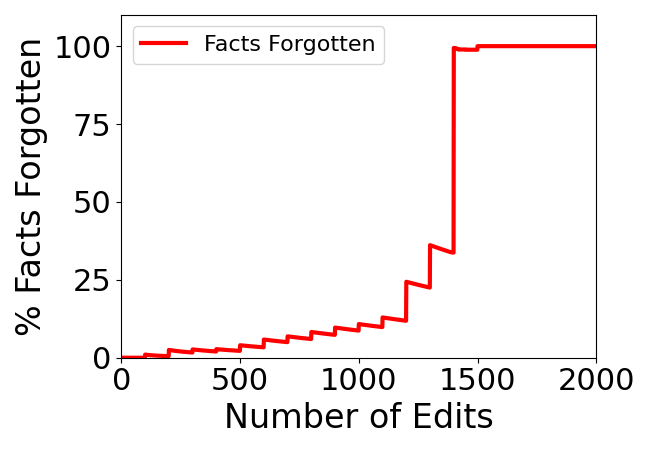}
        \caption{MEMIT}
    \end{subfigure}
    
    \caption{Forgetting plots for Sample 2 for GPT-J (6B).}
    \label{fig:app:editing_proficiency_gptj_sample2}
\end{figure*}

\begin{figure*}
    \centering
    \begin{subfigure}{.24\textwidth}
        \centering
        \includegraphics[width=\linewidth]{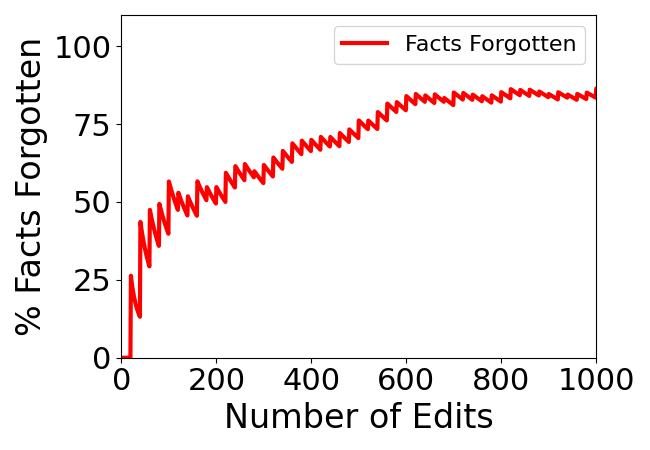}
        \caption{FT-C}
    \end{subfigure}%
    \begin{subfigure}{.24\textwidth}
        \centering
        \includegraphics[width=\linewidth]{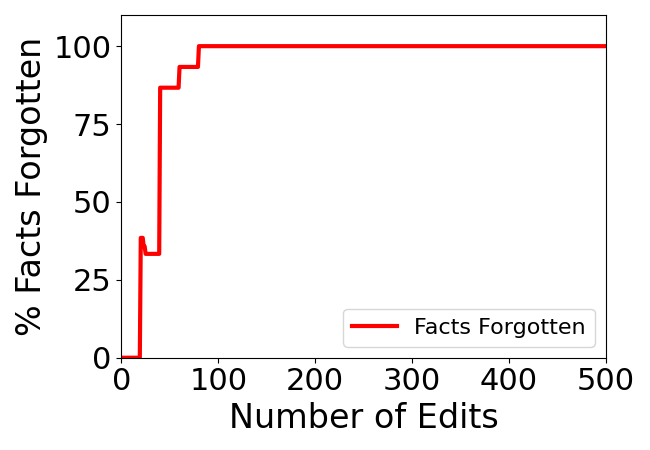}
        \caption{MEND}
    \end{subfigure}%
    \begin{subfigure}{.24\textwidth}
        \centering
        \includegraphics[width=\linewidth]{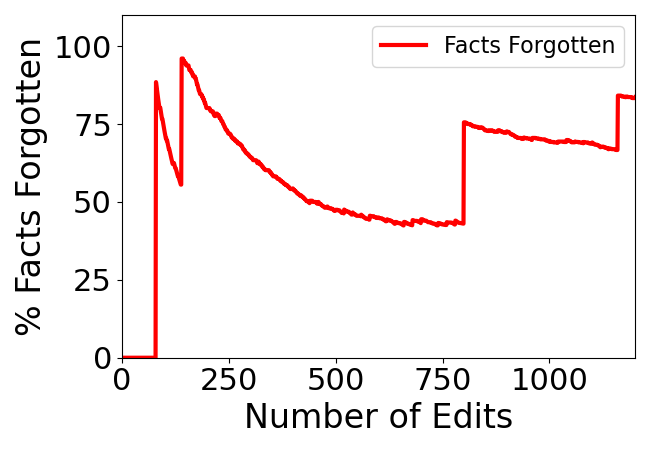}
        \caption{ROME}
    \end{subfigure}
    \begin{subfigure}{.24\textwidth}
        \centering
        \includegraphics[width=\linewidth]{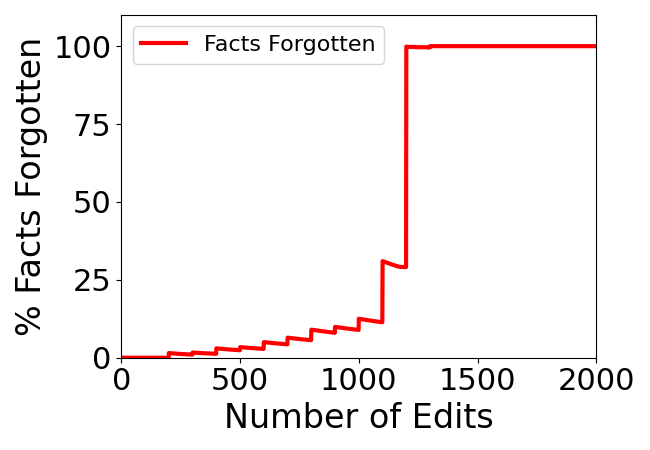}
        \caption{MEMIT}
    \end{subfigure}
    
    \caption{Forgetting plots for Sample 3 for GPT-J (6B).}
    \label{fig:app:editing_proficiency_gptj_sample3}
\end{figure*}

\begin{figure*}
    \centering
    \begin{subfigure}{.24\textwidth}
        \centering
        \includegraphics[width=\linewidth]{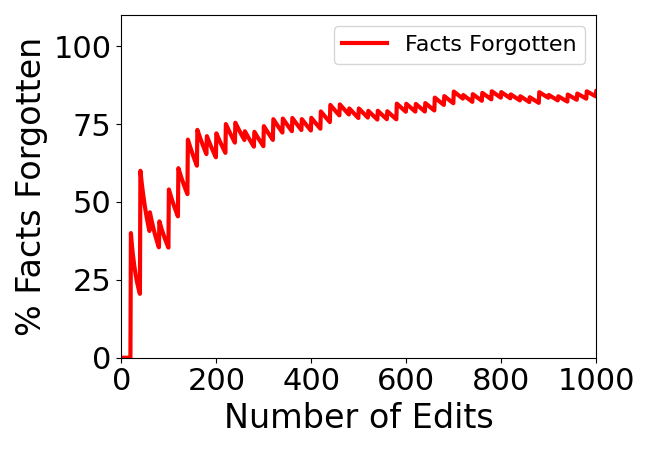}
        \caption{FT-C}
    \end{subfigure}%
    \begin{subfigure}{.24\textwidth}
        \centering
        \includegraphics[width=\linewidth]{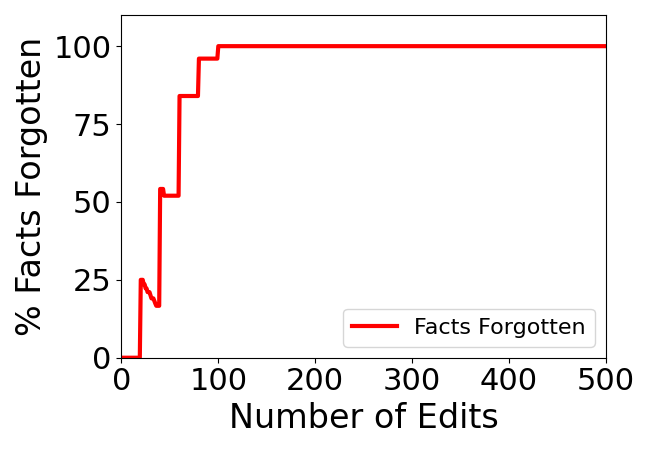}
        \caption{MEND}
    \end{subfigure}%
    \begin{subfigure}{.24\textwidth}
        \centering
        \includegraphics[width=\linewidth]{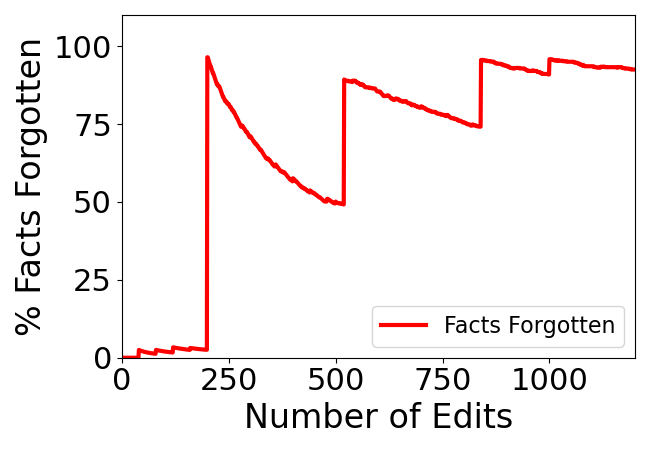}
        \caption{ROME}
    \end{subfigure}
    \begin{subfigure}{.24\textwidth}
        \centering
        \includegraphics[width=\linewidth]{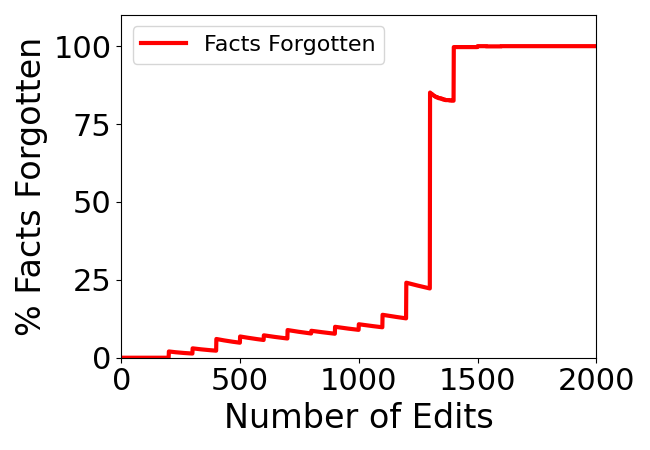}
        \caption{MEMIT}
    \end{subfigure}
    
    \caption{Forgetting plots for Sample 4 for GPT-J (6B).}
    \label{fig:app:editing_proficiency_gptj_sample4}
\end{figure*}

\clearpage
\subsubsection{Downstream Evaluation}\label{sec:appendix:rome:downstream}
In this section, we show plots for the downstream evaluations for both GPT2-XL (1.5B) and GPT-J (6B) for the four samples. Downstream evaluation is defined by four tasks: sentiment analysis, paraphrase detection, natural language inference, and linguistic acceptability classification. Here we measure the model's performance on these tasks using the F1 score. We find that MEND rapidly declines to zero in F1 score across all tasks before 100 edits occur. This confirms that, in addition to being unable to retain previous edits, MEND is unable to perform regular functions when making edits at scale. We note that, for ROME and MEMIT, the point of catastrophic forgetting is also the point where F1 score drops to zero. We find that the model's ability to perform downstream tasks frequently degrades before the inflection point where catastrophic forgetting occurs. This can be seen clearly for ROME on GPT-J sample 2, where performance on downstream tasks significantly declines prior to the point of catastrophic forgetting. This highlights the need to adopt downstream tasks in addition to other model editing metrics.
\begin{figure*}
    \centering
    \begin{subfigure}{.24\textwidth}
        \centering
        \includegraphics[width=\linewidth]{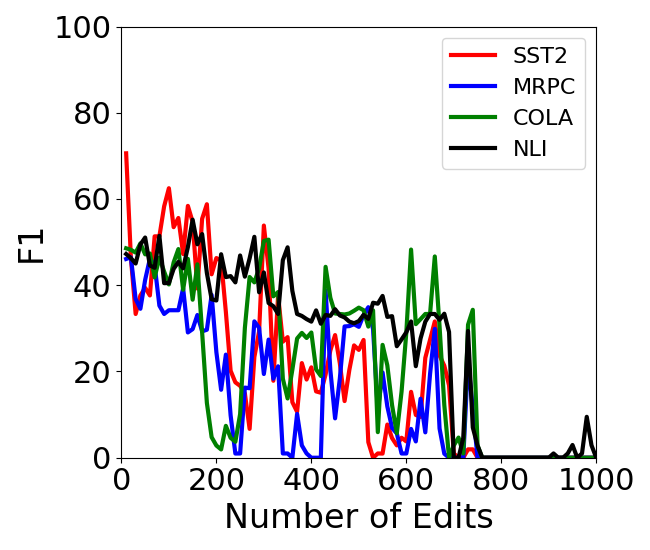}
        \caption{FT-C}
    \end{subfigure}%
    \begin{subfigure}{.24\textwidth}
        \centering
        \includegraphics[width=\linewidth]{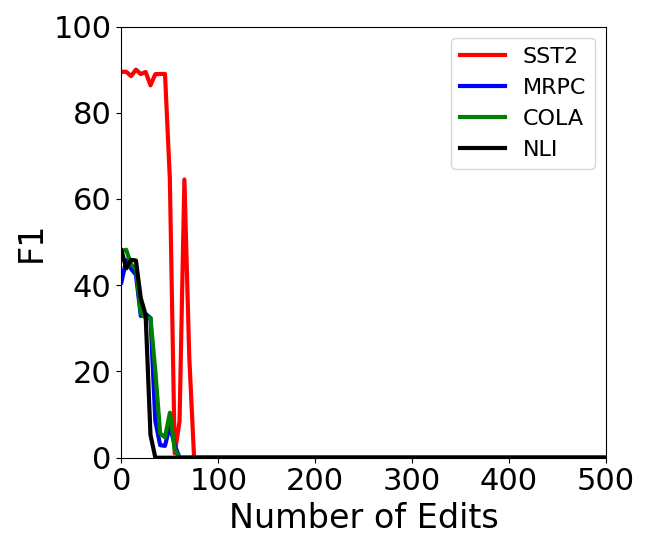}
        \caption{MEND}
    \end{subfigure}%
    \begin{subfigure}{.24\textwidth}
        \centering
        \includegraphics[width=\linewidth]{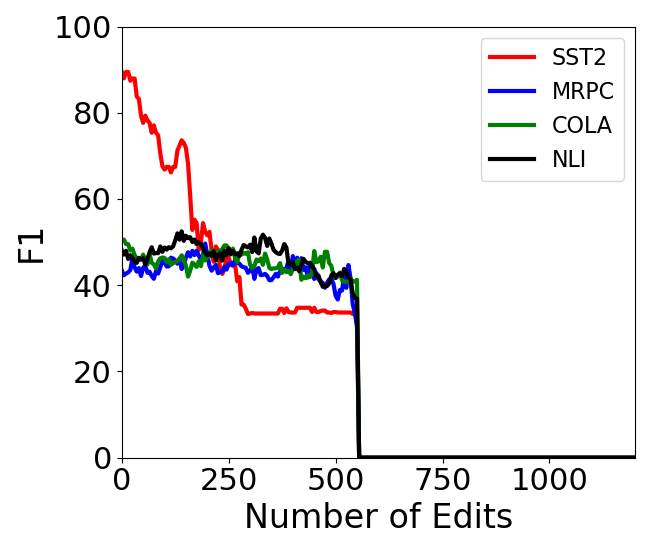}
        \caption{ROME}
    \end{subfigure}
    \begin{subfigure}{.24\textwidth}
        \centering
        \includegraphics[width=\linewidth]{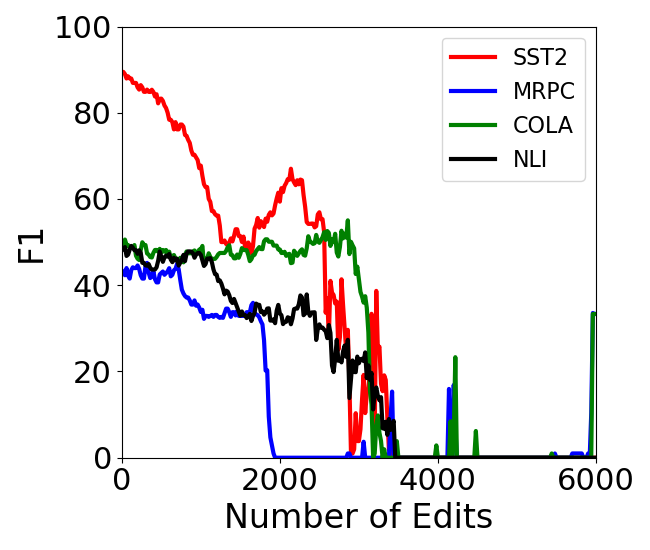}
        \caption{MEMIT}
    \end{subfigure}
    
    \caption{Downstream Performance plots for Sample 1 for GPT-XL (1.5B). }
    \label{fig:app:editing_proficiency_gpt2xl_sample1}
\end{figure*}

\begin{figure*}
    \centering
    \begin{subfigure}{.24\textwidth}
        \centering
        \includegraphics[width=\linewidth]{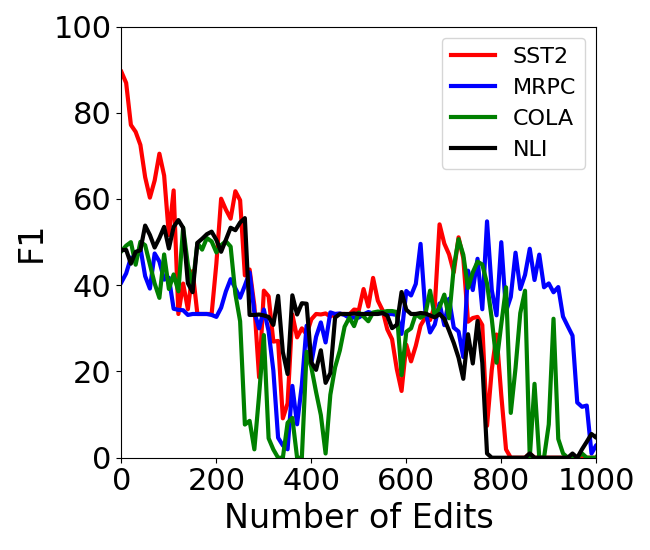}
        \caption{FT-C}
    \end{subfigure}%
    \begin{subfigure}{.24\textwidth}
        \centering
        \includegraphics[width=\linewidth]{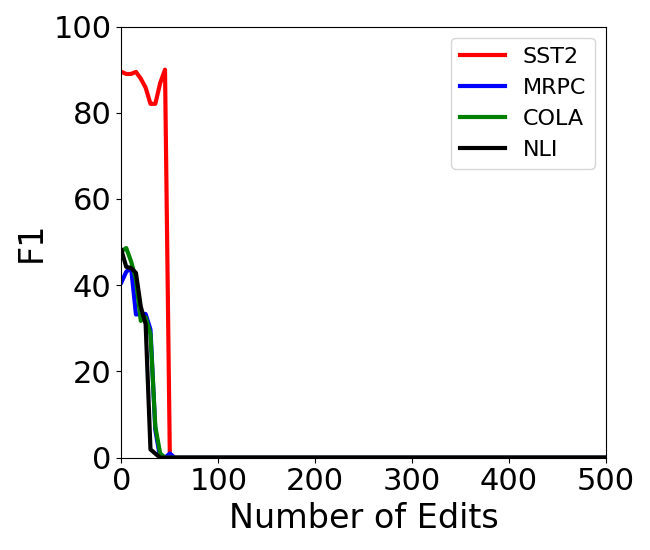}
        \caption{MEND}
    \end{subfigure}%
    \begin{subfigure}{.24\textwidth}
        \centering
        \includegraphics[width=\linewidth]{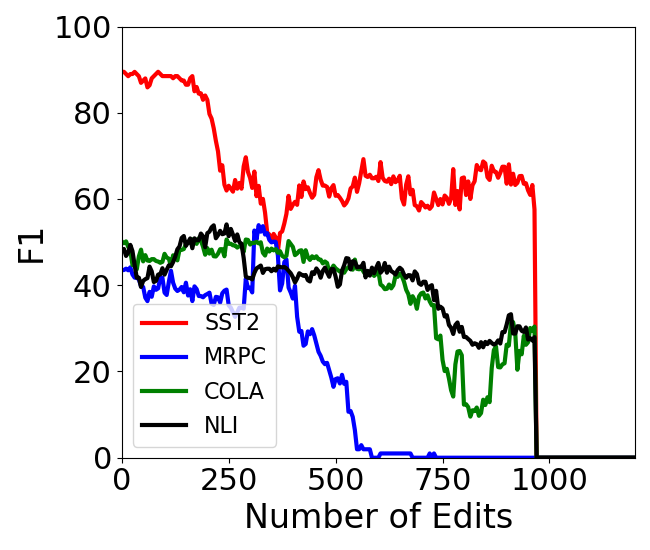}
        \caption{ROME}
    \end{subfigure}
    \begin{subfigure}{.24\textwidth}
        \centering
        \includegraphics[width=\linewidth]{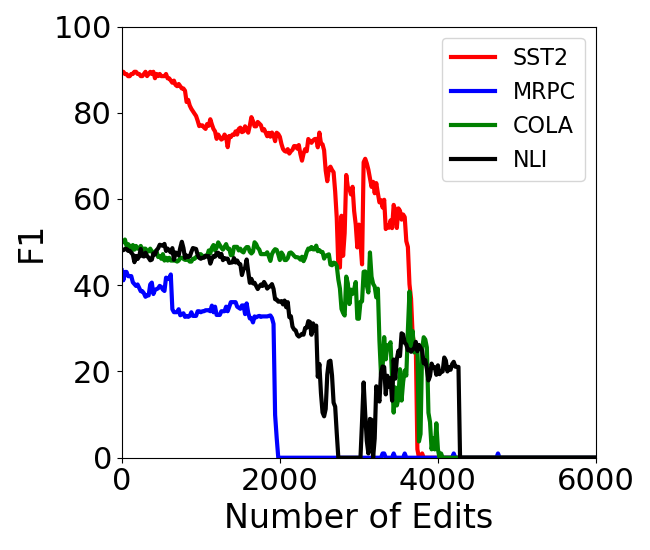}
        \caption{MEMIT}
    \end{subfigure}
    
    \caption{Downstream Performance plots for Sample 2 for GPT-XL (1.5B). }
    \label{fig:app:editing_proficiency_gpt2xl_sample2}
\end{figure*}

\begin{figure*}
    \centering
    \begin{subfigure}{.24\textwidth}
        \centering
        \includegraphics[width=\linewidth]{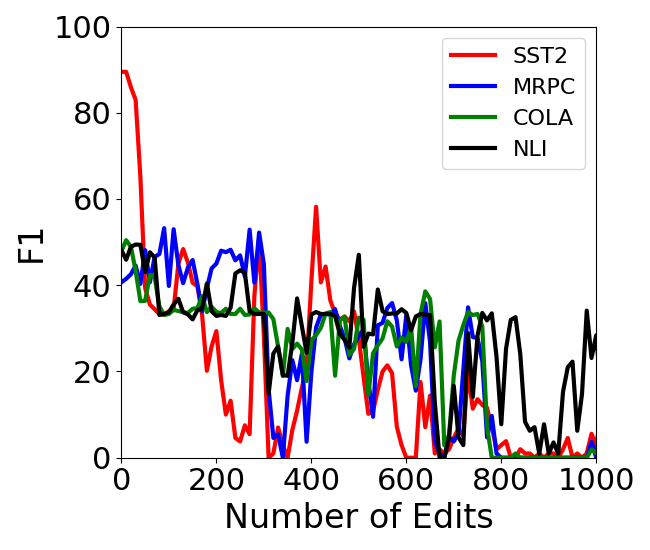}
        \caption{FT-C}
    \end{subfigure}%
    \begin{subfigure}{.24\textwidth}
        \centering
        \includegraphics[width=\linewidth]{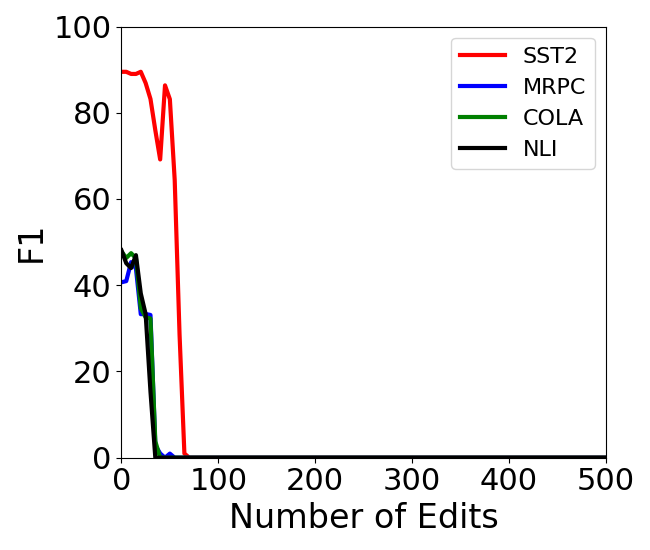}
        \caption{MEND}
    \end{subfigure}%
    \begin{subfigure}{.24\textwidth}
        \centering
        \includegraphics[width=\linewidth]{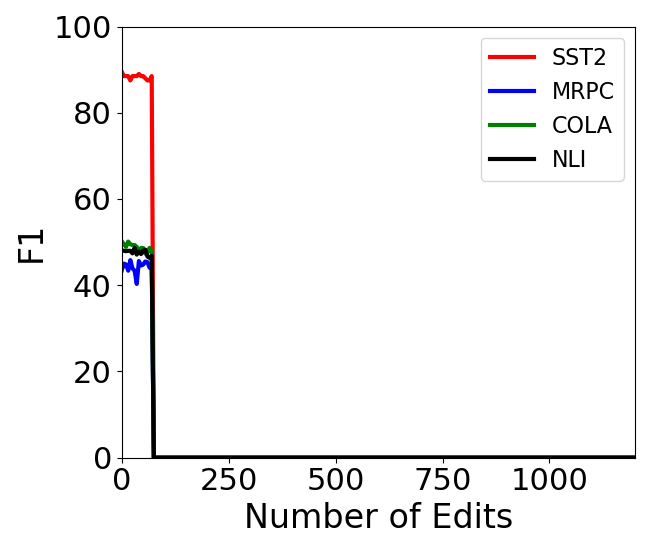}
        \caption{ROME}
    \end{subfigure}
    \begin{subfigure}{.24\textwidth}
        \centering
        \includegraphics[width=\linewidth]{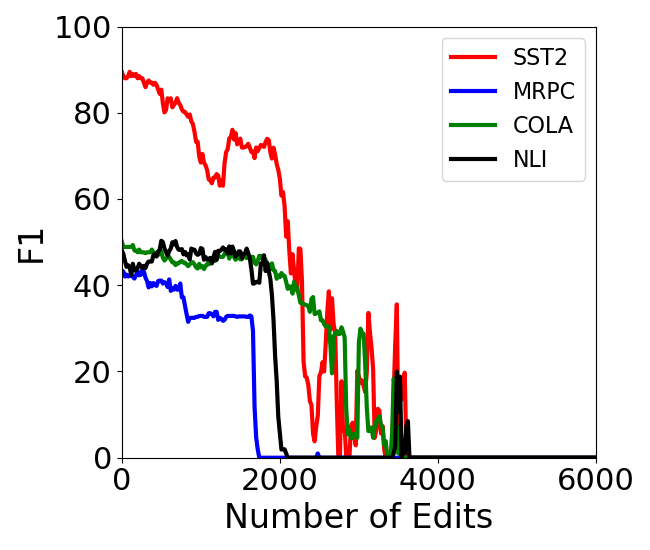}
        \caption{MEMIT}
    \end{subfigure}
    
    \caption{Downstream Performance plots for Sample 3 for GPT-XL (1.5B). }
    \label{fig:app:editing_proficiency_gpt2xl_sample3}
\end{figure*}

\begin{figure*}
    \centering
    \begin{subfigure}{.24\textwidth}
        \centering
        \includegraphics[width=\linewidth]{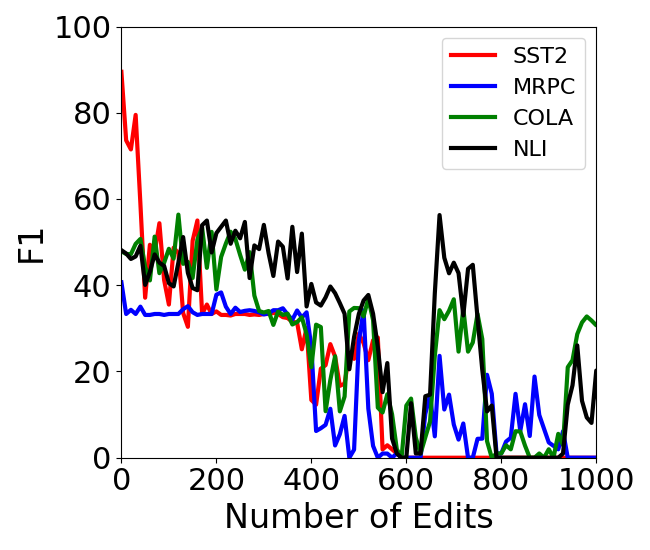}
        \caption{FT-C}
    \end{subfigure}%
    \begin{subfigure}{.24\textwidth}
        \centering
        \includegraphics[width=\linewidth]{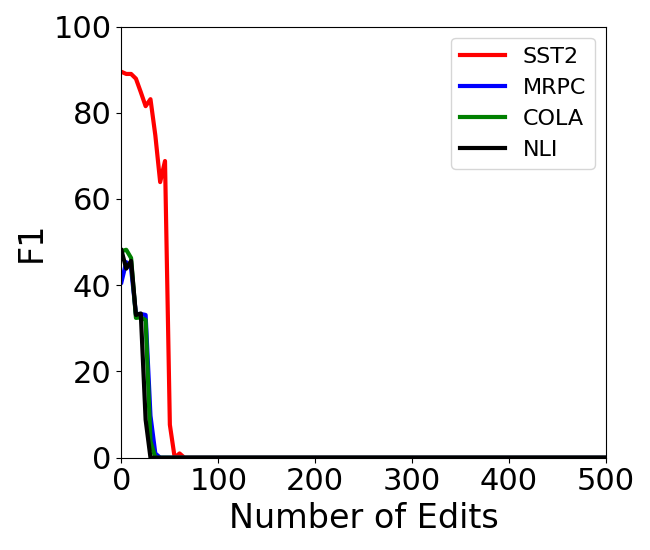}
        \caption{MEND}
    \end{subfigure}%
    \begin{subfigure}{.24\textwidth}
        \centering
        \includegraphics[width=\linewidth]{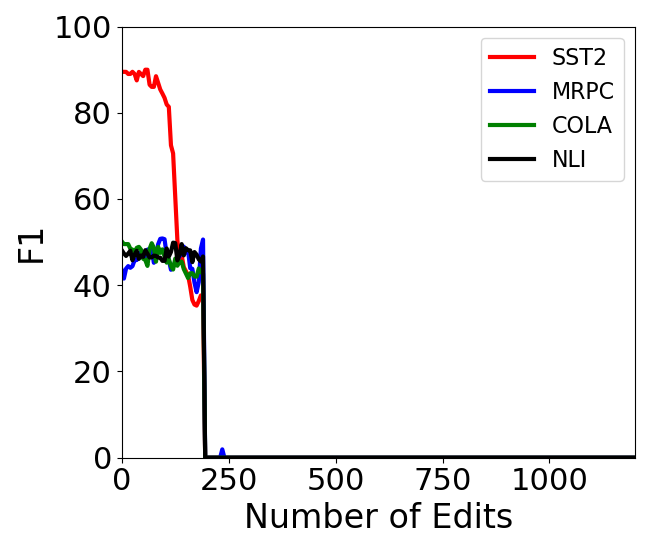}
        \caption{ROME}
    \end{subfigure}
    \begin{subfigure}{.24\textwidth}
        \centering
        \includegraphics[width=\linewidth]{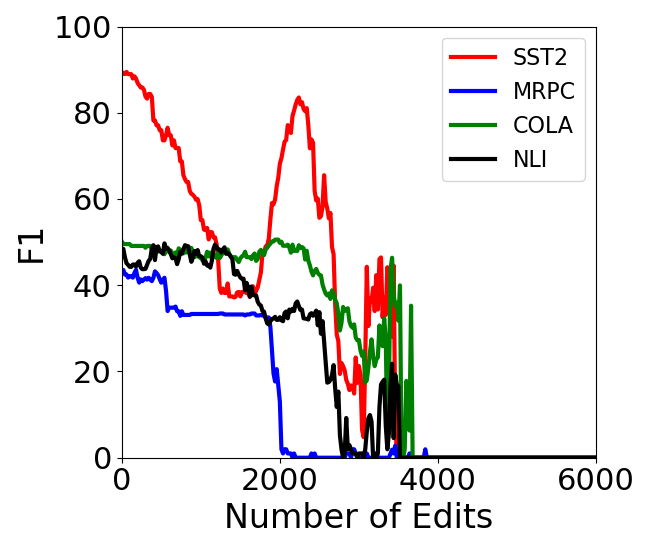}
        \caption{MEMIT}
    \end{subfigure}
    
    \caption{Downstream Performance plots for Sample 4 for GPT-XL (1.5B). }
    \label{fig:app:editing_proficiency_gpt2xl_sample4}
\end{figure*}

%%%%%%%%%%%%%%%%%%END OF GPT2XL PLOTS
\clearpage

\begin{figure*}
    \centering
    \begin{subfigure}{.24\textwidth}
        \centering
        \includegraphics[width=\linewidth]{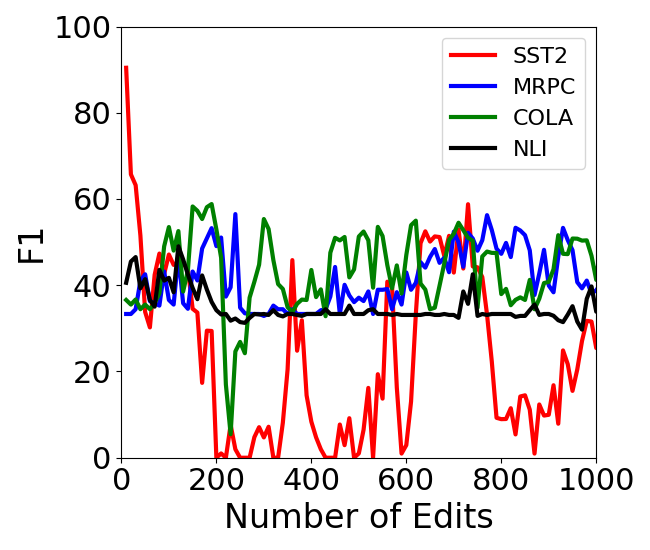}
        \caption{FT-C}
    \end{subfigure}%
    \begin{subfigure}{.24\textwidth}
        \centering
        \includegraphics[width=\linewidth]{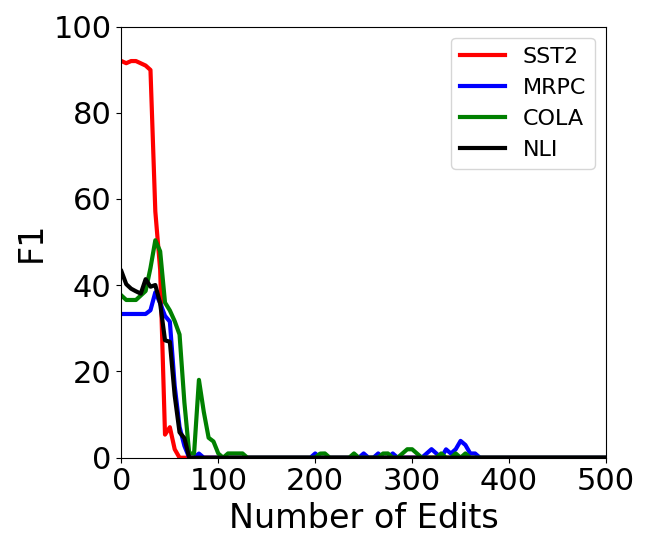}
        \caption{MEND}
    \end{subfigure}%
    \begin{subfigure}{.24\textwidth}
        \centering
        \includegraphics[width=\linewidth]{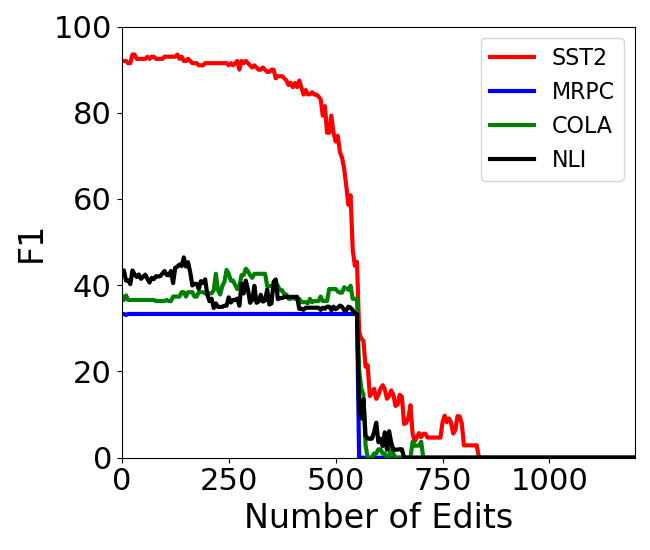}
        \caption{ROME}
    \end{subfigure}
    \begin{subfigure}{.24\textwidth}
        \centering
        \includegraphics[width=\linewidth]{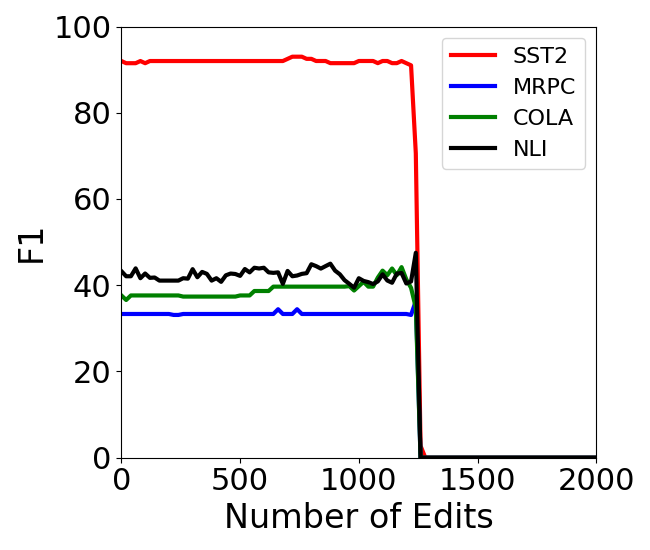}
        \caption{MEMIT}
    \end{subfigure}
    
    \caption{Downstream Performance plots for Sample 1 for GPT-J (6B).}
    \label{fig:app:editing_proficiency_gptj_sample1}
\end{figure*}

\begin{figure*}
    \centering
    \begin{subfigure}{.24\textwidth}
        \centering
        \includegraphics[width=\linewidth]{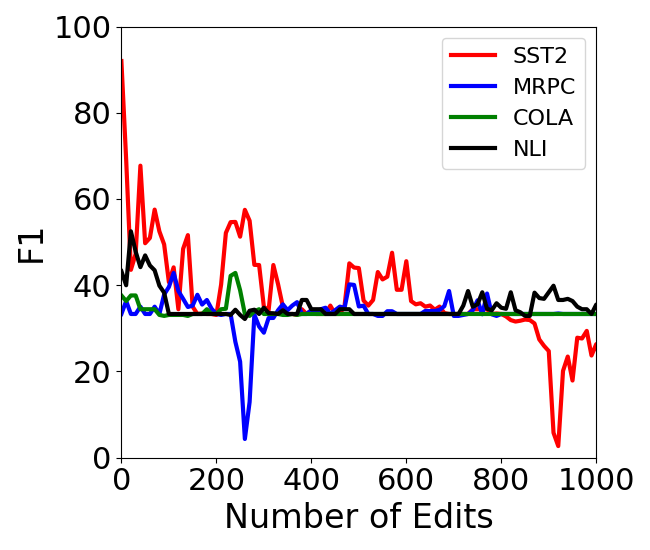}
        \caption{FT-C}
    \end{subfigure}%
    \begin{subfigure}{.24\textwidth}
        \centering
        \includegraphics[width=\linewidth]{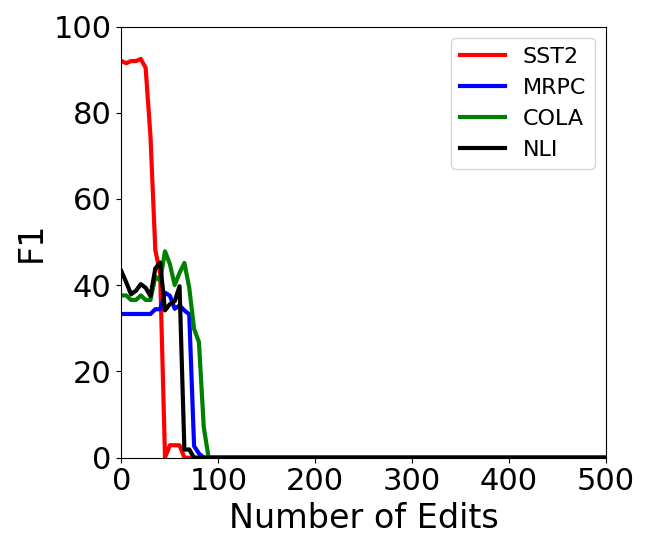}
        \caption{MEND}
    \end{subfigure}%
    \begin{subfigure}{.24\textwidth}
        \centering
        \includegraphics[width=\linewidth]{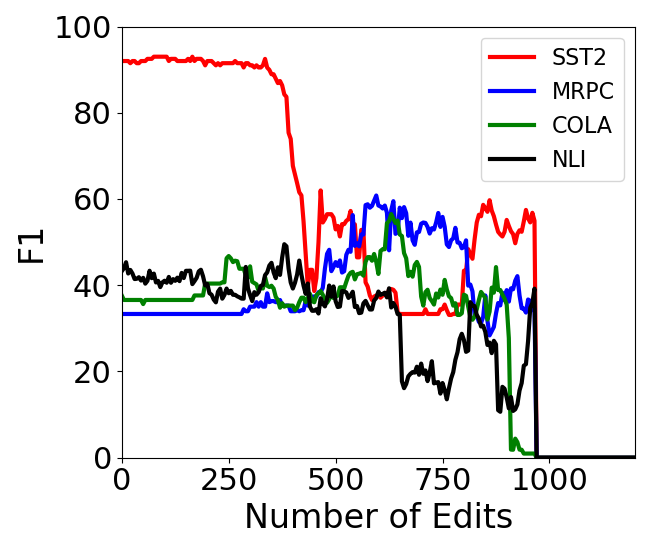}
        \caption{ROME}
    \end{subfigure}
    \begin{subfigure}{.24\textwidth}
        \centering
        \includegraphics[width=\linewidth]{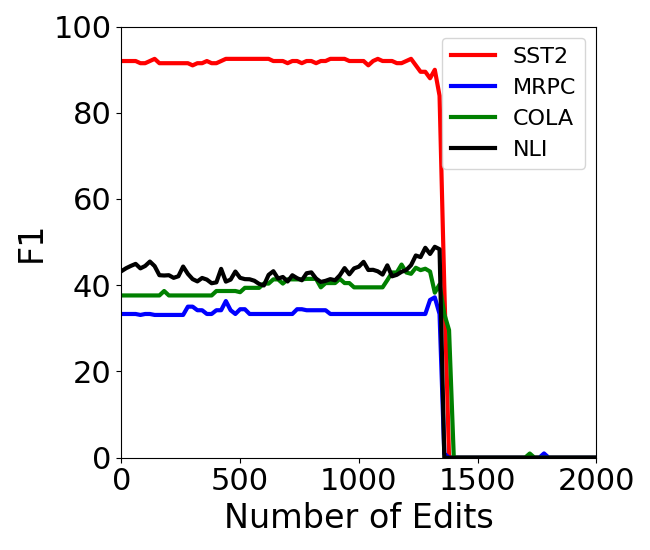}
        \caption{MEMIT}
    \end{subfigure}
    
    \caption{Downstream Performance plots for Sample 2 for GPT-J (6B).}
    \label{fig:app:editing_proficiency_gptj_sample2}
\end{figure*}

\begin{figure*}
    \centering
    \begin{subfigure}{.24\textwidth}
        \centering
        \includegraphics[width=\linewidth]{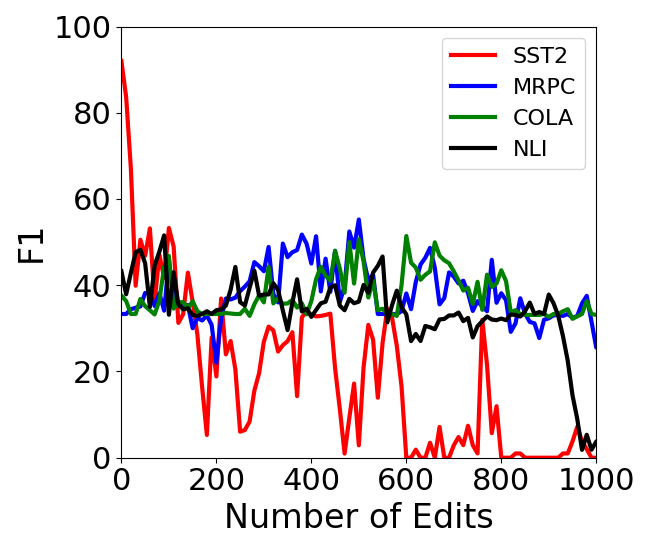}
        \caption{FT-C}
    \end{subfigure}%
    \begin{subfigure}{.24\textwidth}
        \centering
        \includegraphics[width=\linewidth]{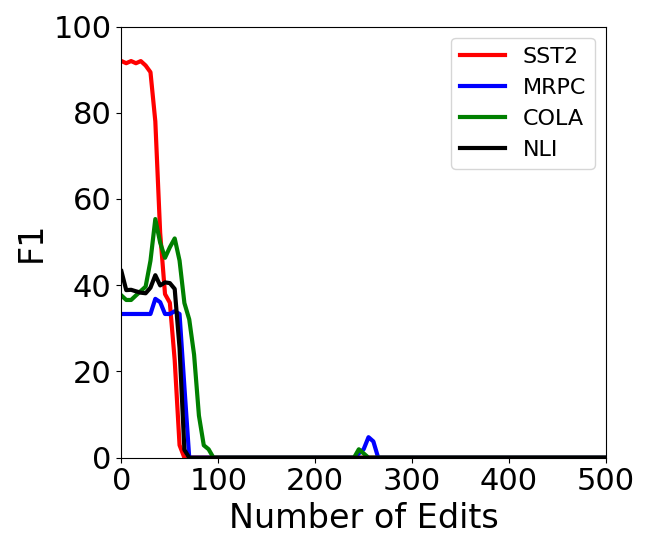}
        \caption{MEND}
    \end{subfigure}%
    \begin{subfigure}{.24\textwidth}
        \centering
        \includegraphics[width=\linewidth]{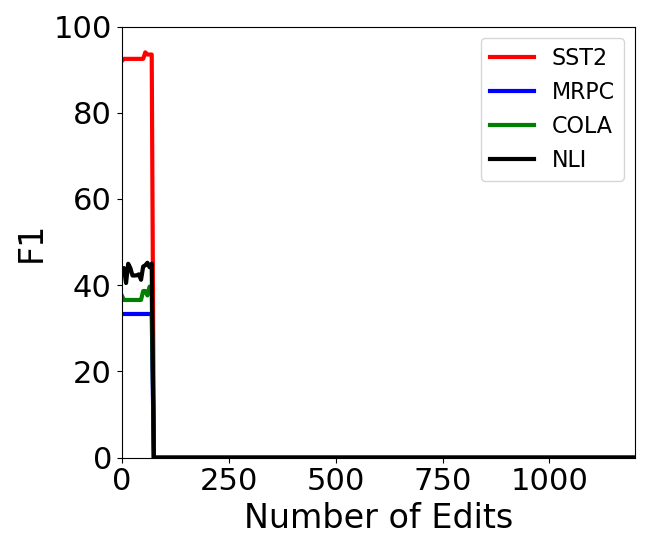}
        \caption{ROME}
    \end{subfigure}
    \begin{subfigure}{.24\textwidth}
        \centering
        \includegraphics[width=\linewidth]{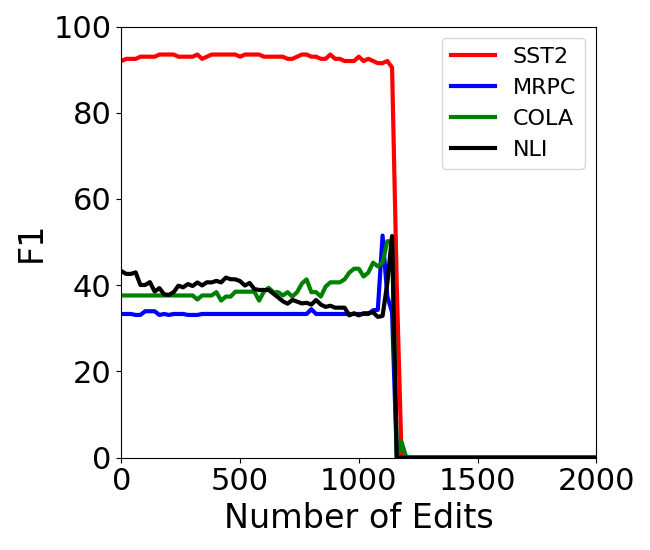}
        \caption{MEMIT}
    \end{subfigure}
    
    \caption{Downstream Performance plots for Sample 3 for GPT-J (6B).}
    \label{fig:app:editing_proficiency_gptj_sample3}
\end{figure*}

\begin{figure*}
    \centering
    \begin{subfigure}{.24\textwidth}
        \centering
        \includegraphics[width=\linewidth]{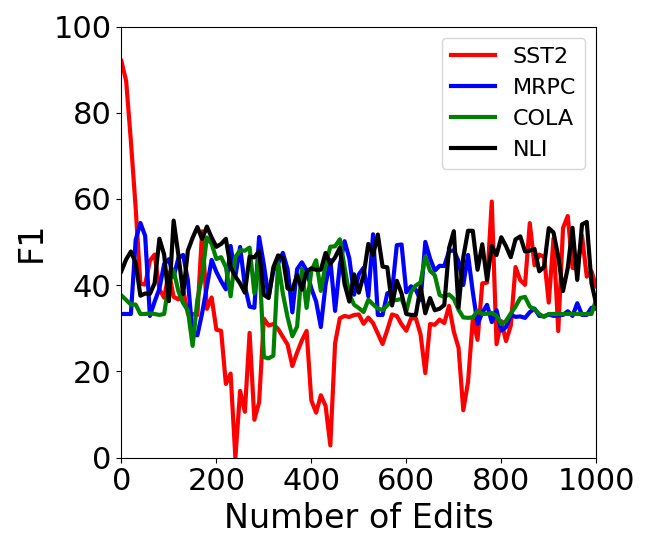}
        \caption{FT-C}
    \end{subfigure}%
    \begin{subfigure}{.24\textwidth}
        \centering
        \includegraphics[width=\linewidth]{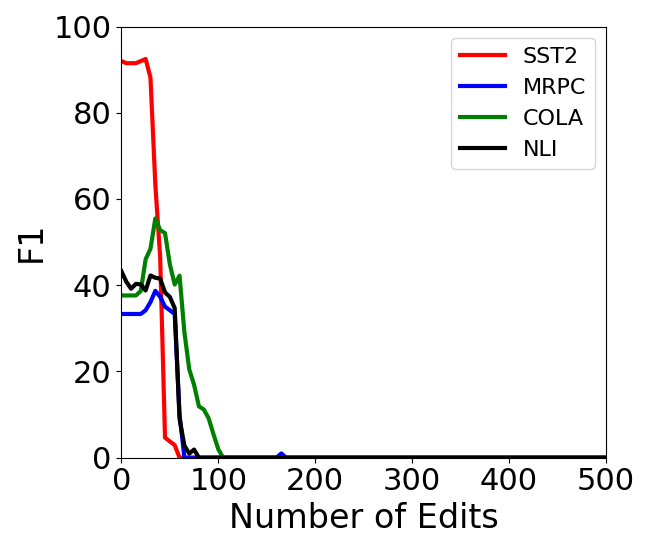}
        \caption{MEND}
    \end{subfigure}%
    \begin{subfigure}{.24\textwidth}
        \centering
        \includegraphics[width=\linewidth]{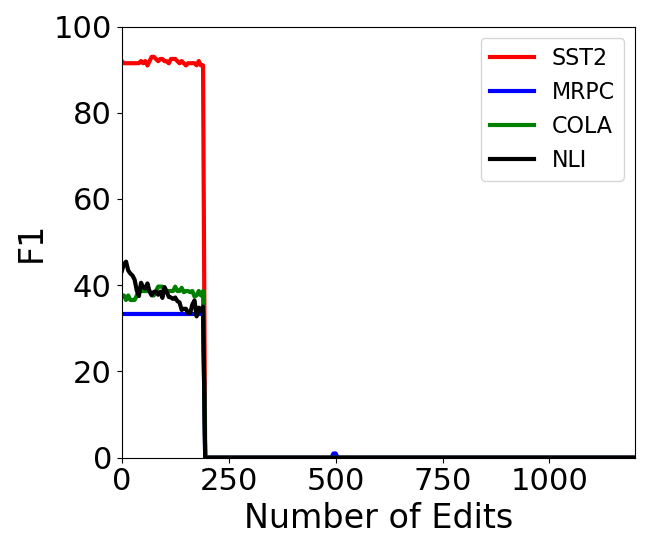}
        \caption{ROME}
    \end{subfigure}
    \begin{subfigure}{.24\textwidth}
        \centering
        \includegraphics[width=\linewidth]{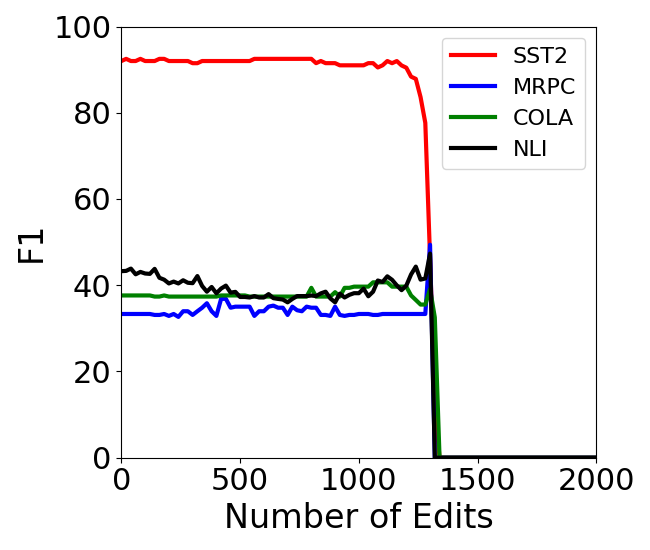}
        \caption{MEMIT}
    \end{subfigure}
    
    \caption{Downstream Performance plots for Sample 4 for GPT-J (6B).}
    \label{fig:app:editing_proficiency_gptj_sample4}
\end{figure*}

\clearpage
\subsubsection{Source of Forgetting}\label{sec:appendix:rome:distance}
Here we present plots that show the normalized L2 distance between the weights of the edited layer and the original layer for both GPT2-XL(1.5B) and GPT-J(6B) for all four samples. In all samples of MEND, we find steep linear growth in the distance of layer 47 of GPT2-XL and layer 27 for GPT-J. ROME exhibits the behavior of a step function across all samples. Each step corresponds to a spike in forgetfulness as shown in appendix \ref{sec:appendix:rome:forgetting}. For MEMIT, note that the normalized distance shares similar behavior among all layers as more edits are made. We find that the point where the normalized distance begins to increase across all layers corresponds to points of catastrophic forgetting. 
\begin{figure*}
    \centering
    \begin{subfigure}{.24\textwidth}
        \centering
        \includegraphics[width=\linewidth]{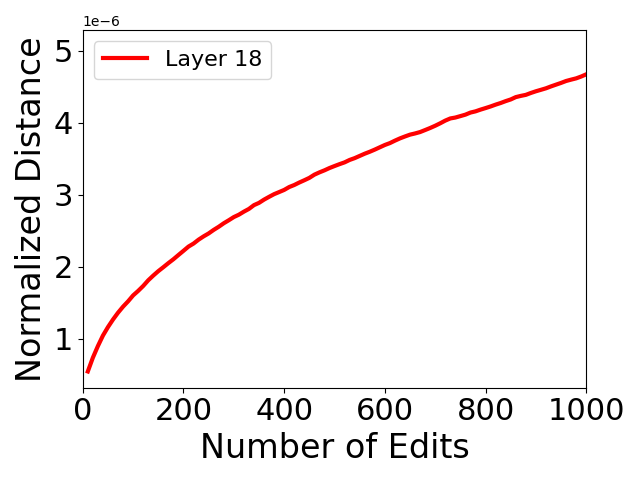}
        \caption{FT-C}
    \end{subfigure}%
    \begin{subfigure}{.24\textwidth}
        \centering
        \includegraphics[width=\linewidth]{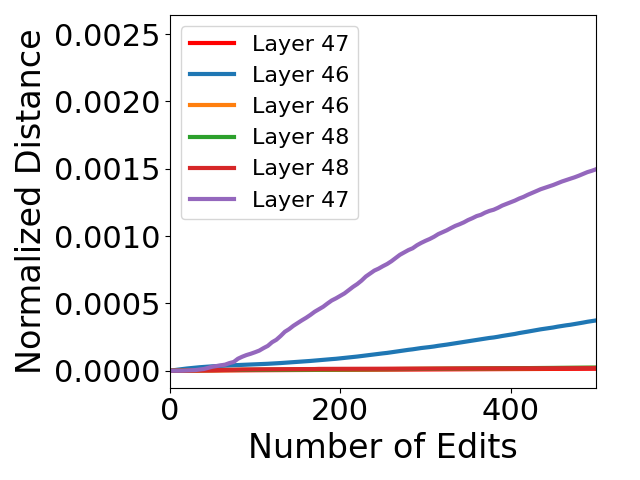}
        \caption{MEND}
    \end{subfigure}%
    \begin{subfigure}{.24\textwidth}
        \centering
        \includegraphics[width=\linewidth]{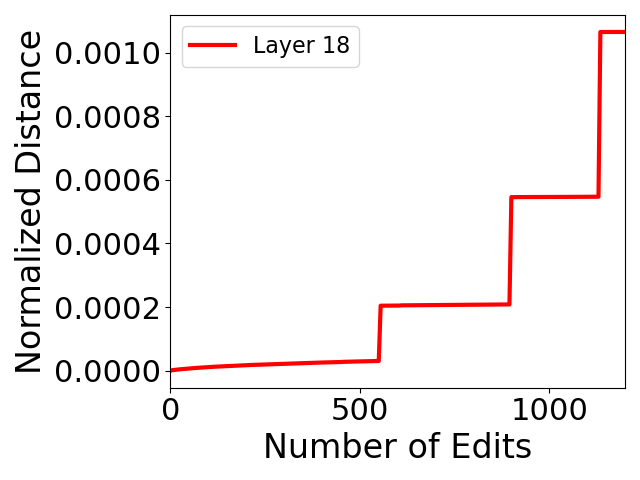}
        \caption{ROME}
    \end{subfigure}
    \begin{subfigure}{.24\textwidth}
        \centering
        \includegraphics[width=\linewidth]{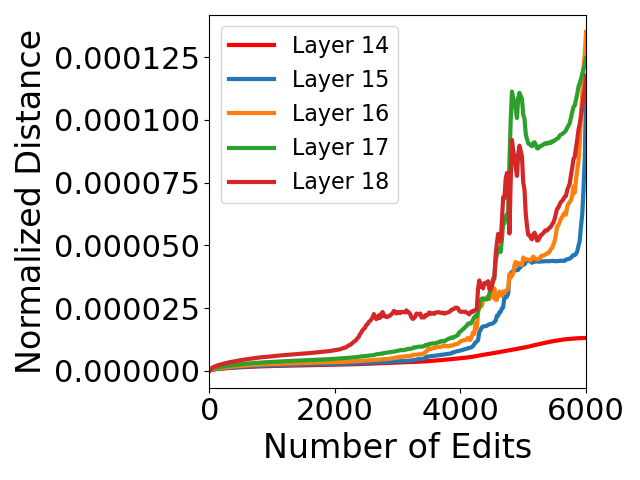}
        \caption{MEMIT}
    \end{subfigure}
    
    \caption{Distance plots for Sample 1 for GPT-XL (1.5B). }
    \label{fig:app:editing_proficiency_gpt2xl_sample1}
\end{figure*}

\begin{figure*}
    \centering
    \begin{subfigure}{.24\textwidth}
        \centering
        \includegraphics[width=\linewidth]{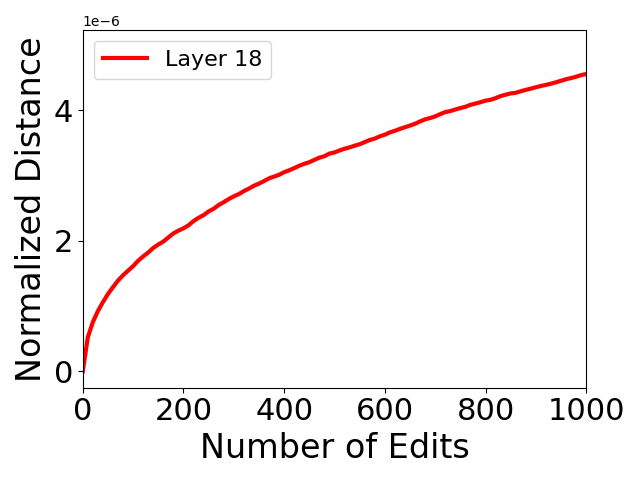}
        \caption{FT-C}
    \end{subfigure}%
    \begin{subfigure}{.24\textwidth}
        \centering
        \includegraphics[width=\linewidth]{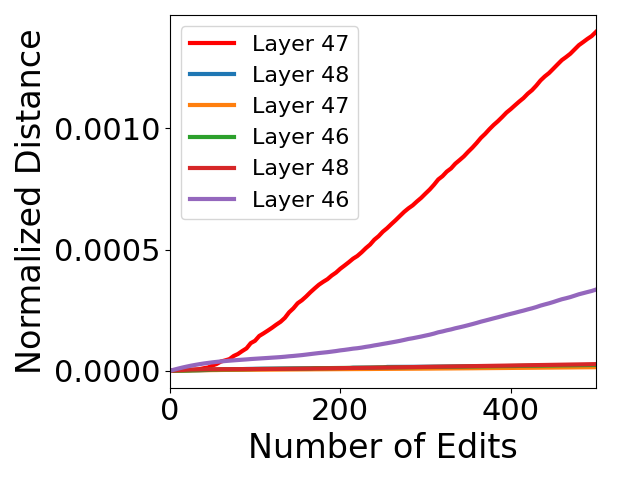}
        \caption{MEND}
    \end{subfigure}%
    \begin{subfigure}{.24\textwidth}
        \centering
        \includegraphics[width=\linewidth]{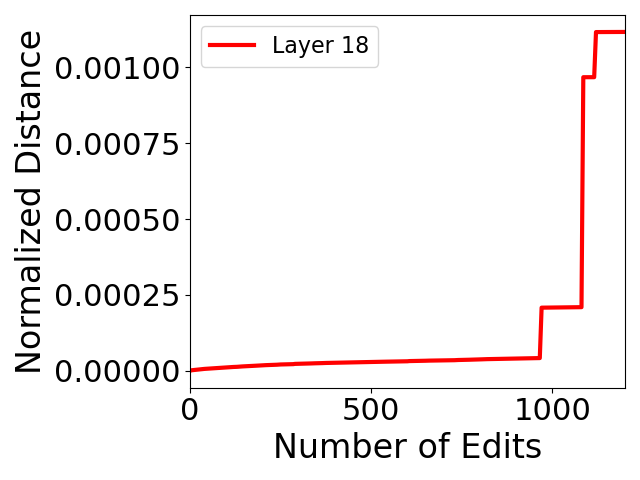}
        \caption{ROME}
    \end{subfigure}
    \begin{subfigure}{.24\textwidth}
        \centering
        \includegraphics[width=\linewidth]{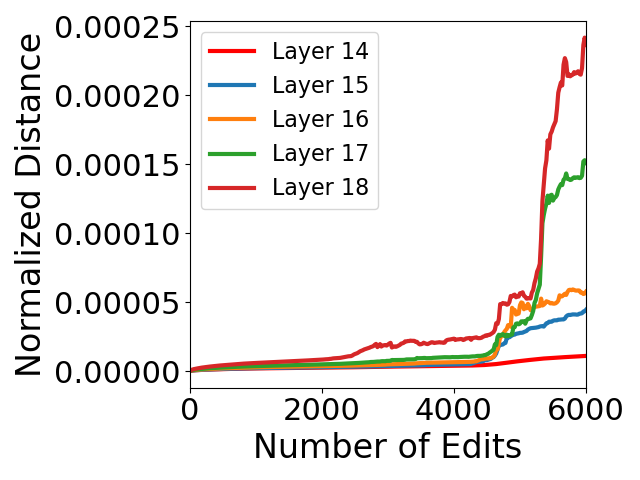}
        \caption{MEMIT}
    \end{subfigure}
    
    \caption{Distance plots for Sample 2 for GPT-XL (1.5B). }
    \label{fig:app:editing_proficiency_gpt2xl_sample2}
\end{figure*}

\begin{figure*}
    \centering
    \begin{subfigure}{.24\textwidth}
        \centering
        \includegraphics[width=\linewidth]{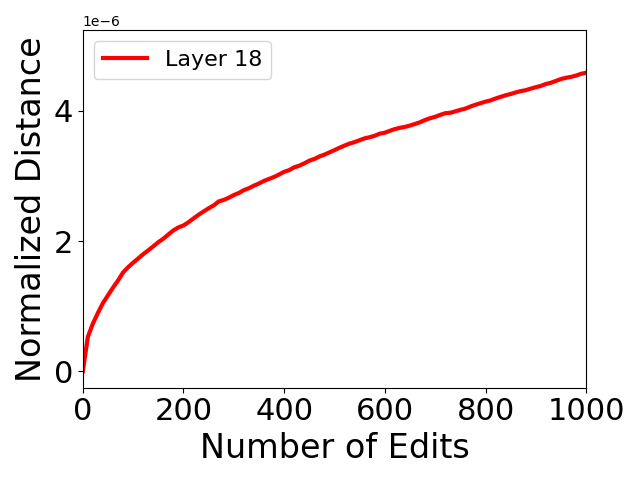}
        \caption{FT-C}
    \end{subfigure}%
    \begin{subfigure}{.24\textwidth}
        \centering
        \includegraphics[width=\linewidth]{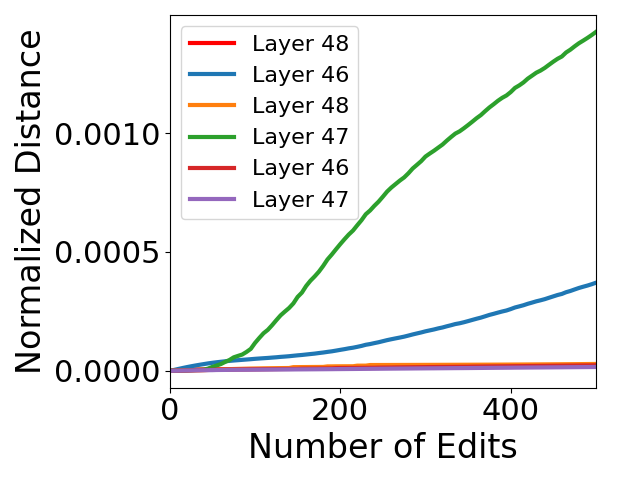}
        \caption{MEND}
    \end{subfigure}%
    \begin{subfigure}{.24\textwidth}
        \centering
        \includegraphics[width=\linewidth]{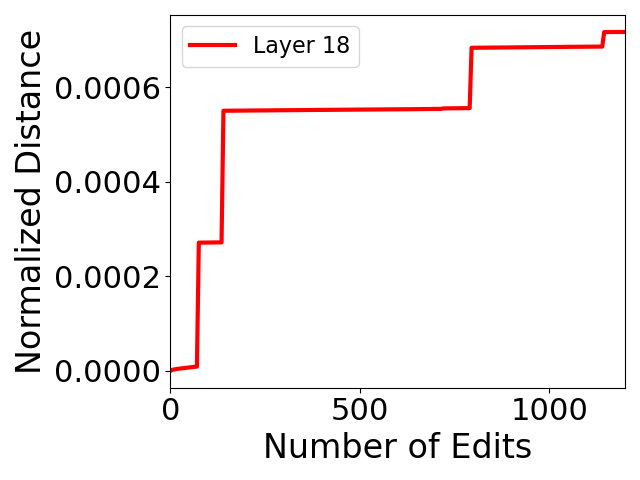}
        \caption{ROME}
    \end{subfigure}
    \begin{subfigure}{.24\textwidth}
        \centering
        \includegraphics[width=\linewidth]{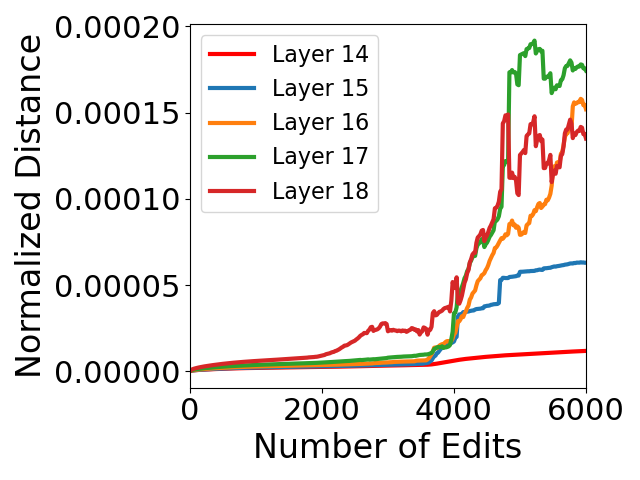}
        \caption{MEMIT}
    \end{subfigure}
    
    \caption{Distance plots for Sample 3 for GPT-XL (1.5B). }
    \label{fig:app:editing_proficiency_gpt2xl_sample3}
\end{figure*}

\begin{figure*}
    \centering
    \begin{subfigure}{.24\textwidth}
        \centering
        \includegraphics[width=\linewidth]{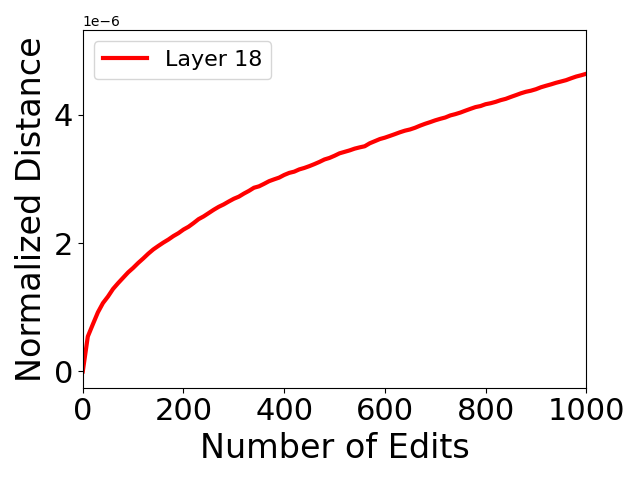}
        \caption{FT-C}
    \end{subfigure}%
    \begin{subfigure}{.24\textwidth}
        \centering
        \includegraphics[width=\linewidth]{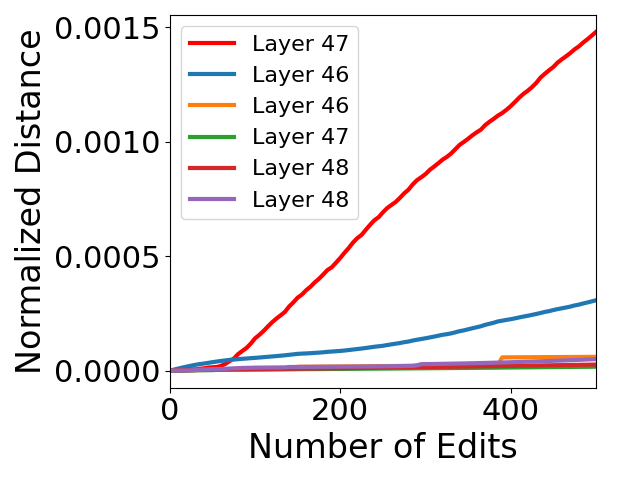}
        \caption{MEND}
    \end{subfigure}%
    \begin{subfigure}{.24\textwidth}
        \centering
        \includegraphics[width=\linewidth]{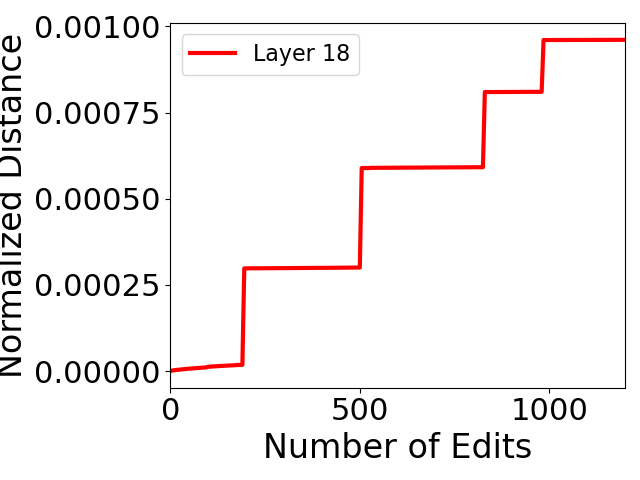}
        \caption{ROME}
    \end{subfigure}
    \begin{subfigure}{.24\textwidth}
        \centering
        \includegraphics[width=\linewidth]{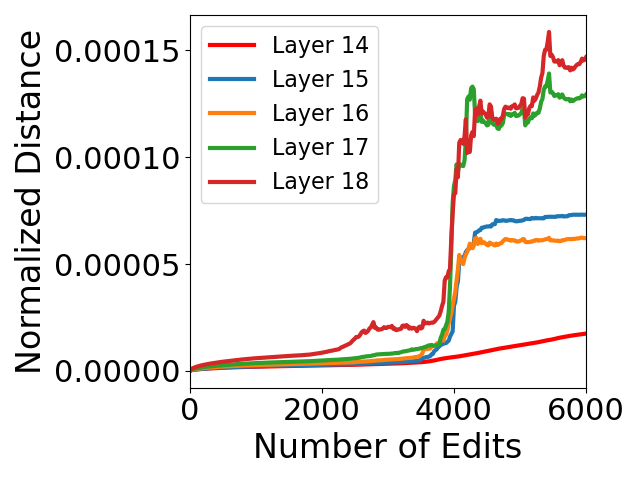}
        \caption{MEMIT}
    \end{subfigure}
    
    \caption{Distance plots for Sample 4 for GPT-XL (1.5B). }
    \label{fig:app:editing_proficiency_gpt2xl_sample4}
\end{figure*}

%%%%%%%%%%%%%%%%%%END OF GPT2XL PLOTS

\clearpage
\begin{figure*}
    \centering
    \begin{subfigure}{.24\textwidth}
        \centering
        \includegraphics[width=\linewidth]{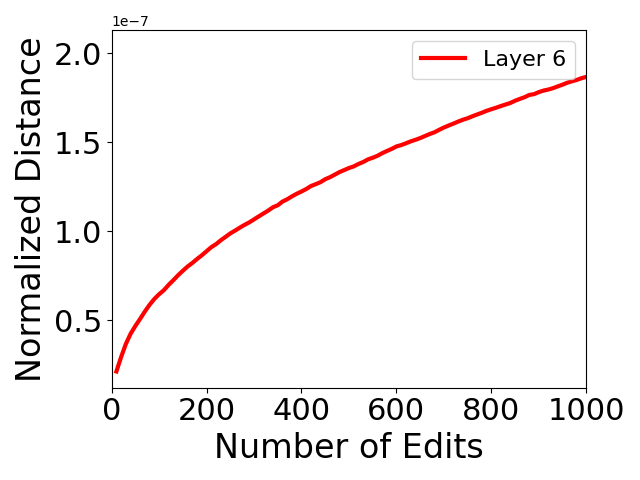}
        \caption{FT-C}
    \end{subfigure}%
    \begin{subfigure}{.24\textwidth}
        \centering
        \includegraphics[width=\linewidth]{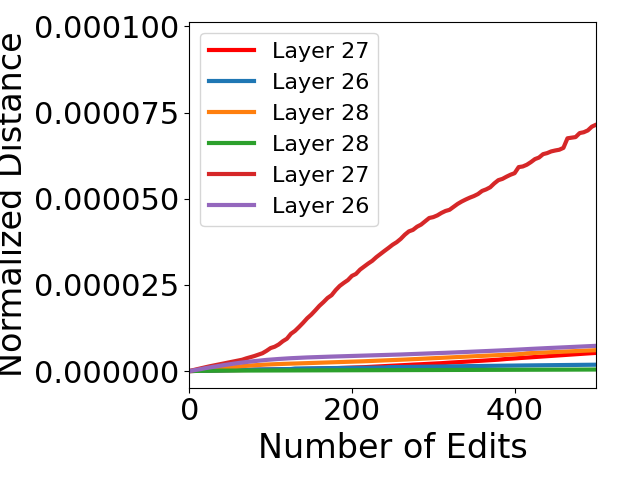}
        \caption{MEND}
    \end{subfigure}%
    \begin{subfigure}{.24\textwidth}
        \centering
        \includegraphics[width=\linewidth]{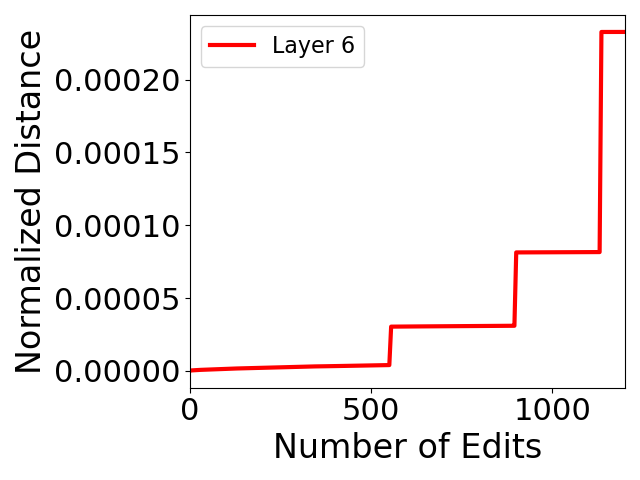}
        \caption{ROME}
    \end{subfigure}
    \begin{subfigure}{.24\textwidth}
        \centering
        \includegraphics[width=\linewidth]{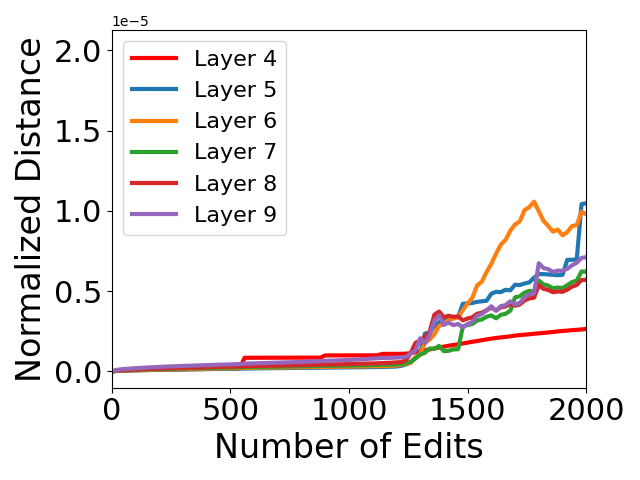}
        \caption{MEMIT}
    \end{subfigure}
    
    \caption{Distance plots for Sample 1 for GPT-J (6B).}
    \label{fig:app:editing_proficiency_gptj_sample1}
\end{figure*}

\begin{figure*}
    \centering
    \begin{subfigure}{.24\textwidth}
        \centering
        \includegraphics[width=\linewidth]{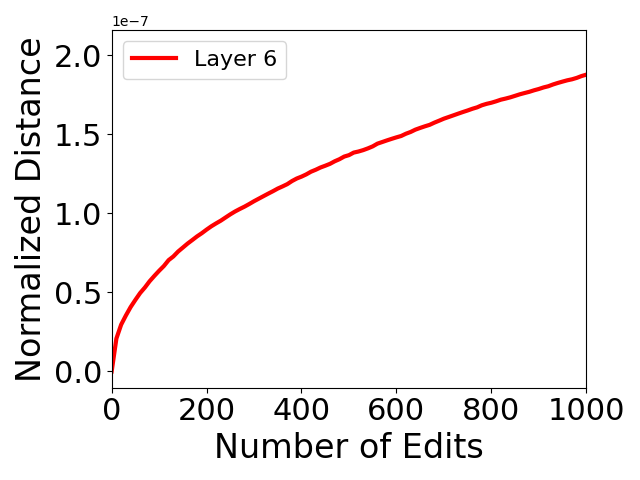}
        \caption{FT-C}
    \end{subfigure}%
    \begin{subfigure}{.24\textwidth}
        \centering
        \includegraphics[width=\linewidth]{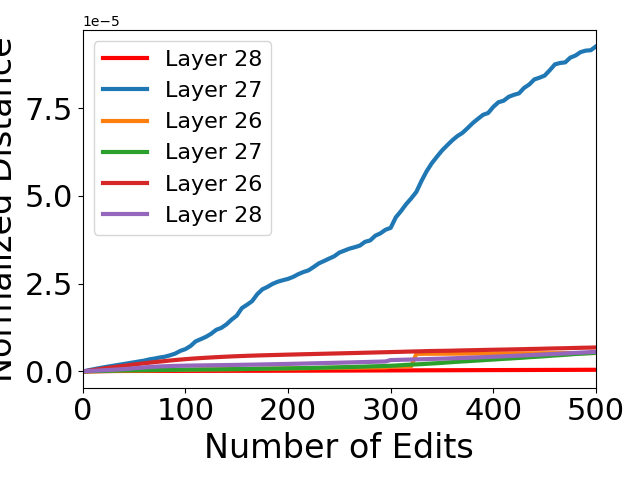}
        \caption{MEND}
    \end{subfigure}%
    \begin{subfigure}{.24\textwidth}
        \centering
        \includegraphics[width=\linewidth]{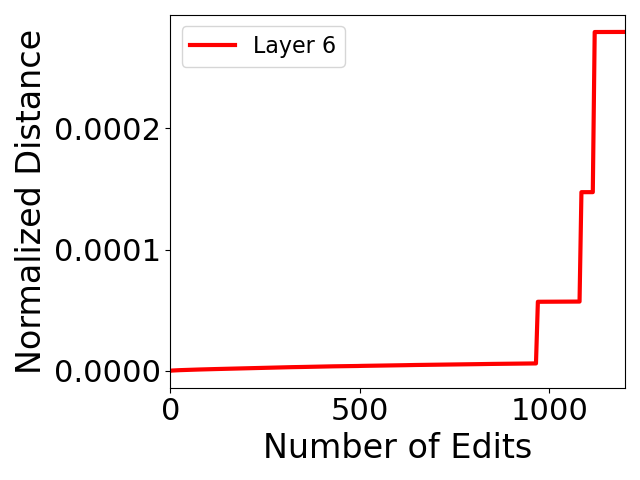}
        \caption{ROME}
    \end{subfigure}
    \begin{subfigure}{.24\textwidth}
        \centering
        \includegraphics[width=\linewidth]{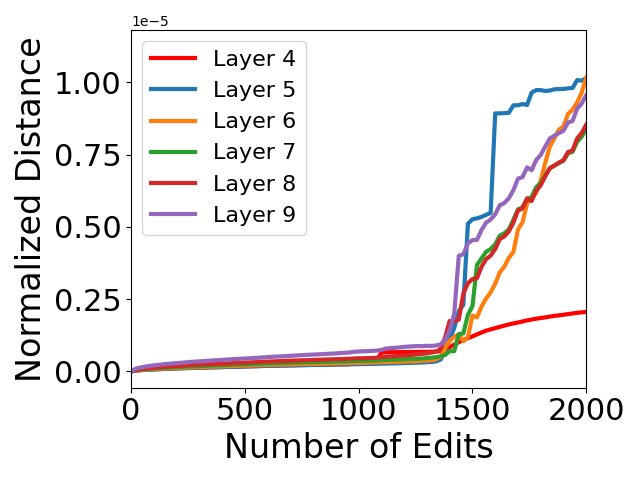}
        \caption{MEMIT}
    \end{subfigure}
    
    \caption{Distance plots for Sample 2 for GPT-J (6B).}
    \label{fig:app:editing_proficiency_gptj_sample2}
\end{figure*}

\begin{figure*}
    \centering
    \begin{subfigure}{.24\textwidth}
        \centering
        \includegraphics[width=\linewidth]{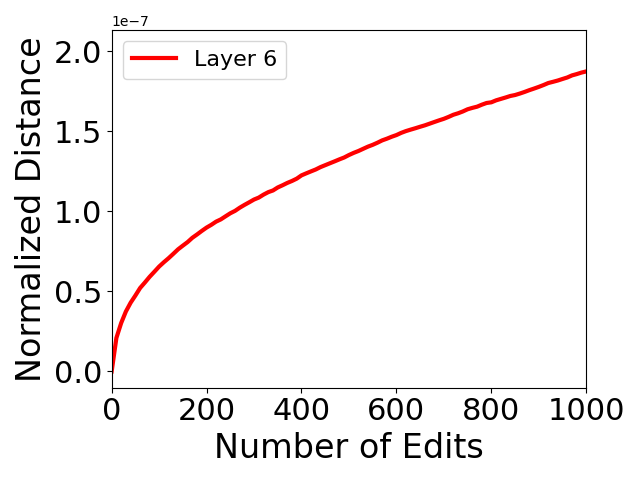}
        \caption{FT-C}
    \end{subfigure}%
    \begin{subfigure}{.24\textwidth}
        \centering
        \includegraphics[width=\linewidth]{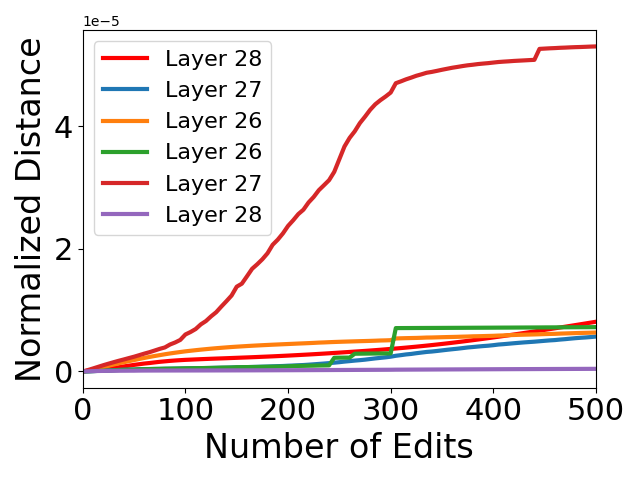}
        \caption{MEND}
    \end{subfigure}%
    \begin{subfigure}{.24\textwidth}
        \centering
        \includegraphics[width=\linewidth]{Appendix_Plots/ROME_gptJ_sample2/ROME_distance.png}
        \caption{ROME}
    \end{subfigure}
    \begin{subfigure}{.24\textwidth}
        \centering
        \includegraphics[width=\linewidth]{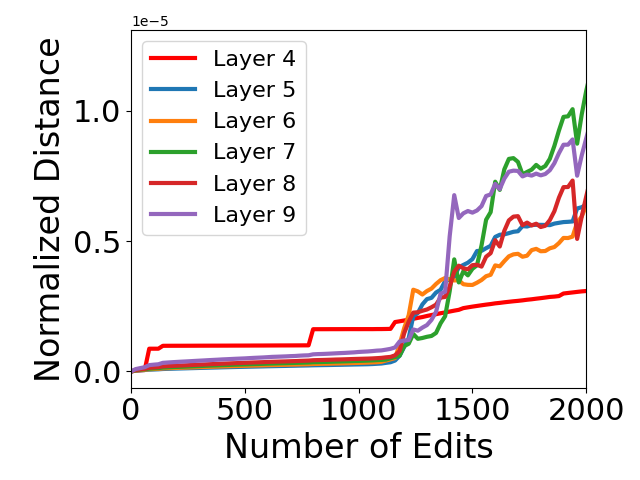}
        \caption{MEMIT}
    \end{subfigure}
    
    \caption{Distance plots for Sample 3 for GPT-J (6B).}
    \label{fig:app:editing_proficiency_gptj_sample3}
\end{figure*}

\begin{figure*}
    \centering
    \begin{subfigure}{.24\textwidth}
        \centering
        \includegraphics[width=\linewidth]{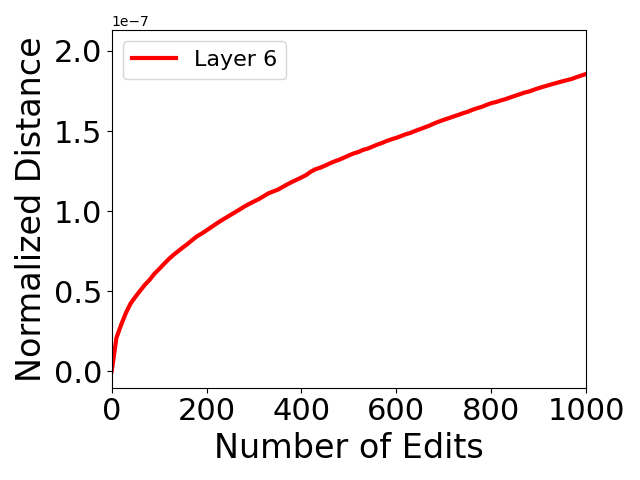}
        \caption{FT-C}
    \end{subfigure}%
    \begin{subfigure}{.24\textwidth}
        \centering
        \includegraphics[width=\linewidth]{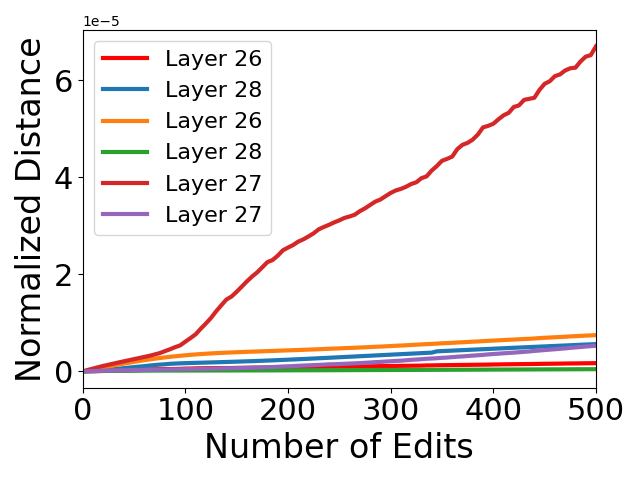}
        \caption{MEND}
    \end{subfigure}%
    \begin{subfigure}{.24\textwidth}
        \centering
        \includegraphics[width=\linewidth]{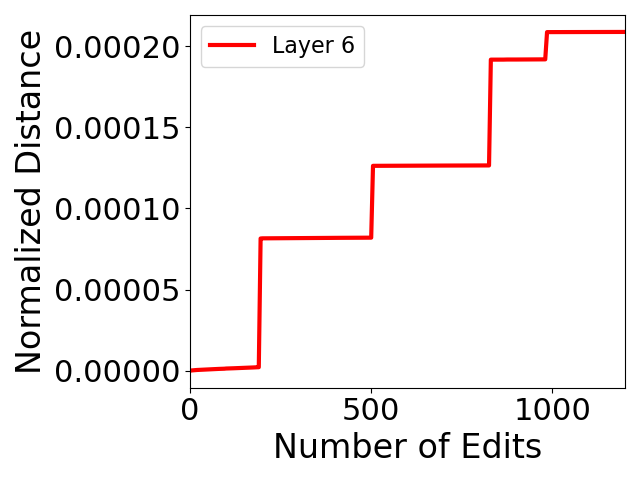}
        \caption{ROME}
    \end{subfigure}
    \begin{subfigure}{.24\textwidth}
        \centering
        \includegraphics[width=\linewidth]{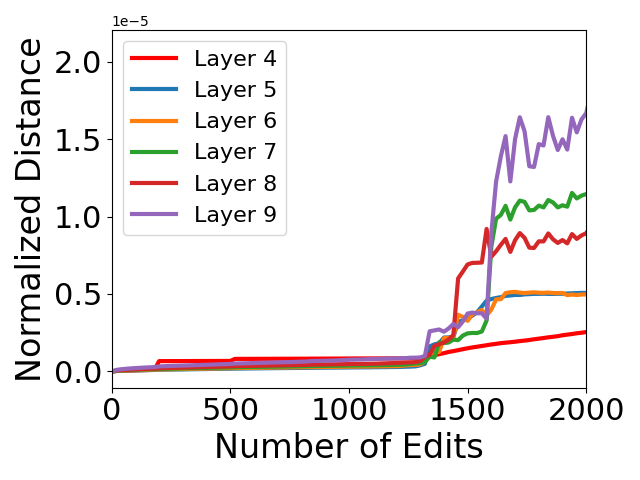}
        \caption{MEMIT}
    \end{subfigure}
    
    \caption{Distance plots for Sample 4 for GPT-J (6B).}
    \label{fig:app:editing_proficiency_gptj_sample4}
\end{figure*}

\clearpage
\subsection{Generation Examples}
Below are some examples of generations produced by GPT-J when edited by ROME post catastrophic forgetting. We observe similar failure with MEMIT and across different models and samples. 

\begin{table*}[h]
\centering
\begin{tabular}{{p{4cm}|p{8cm}}}

Prompt & Generated Text \\ \hline

\texttt{The Tonight Show with Jay Leno is my favorite show that has aired on} & \texttt{CBS CBS. CBS CBS. CBS CBS CBS CBS CBS CBS CBS CBS CBS CBS. CBS CBS. CBS CBS CBS CBS CBS CBS CBS CBS CBS CBS. CBS CBS CBS CBS CBS CBS. CBS CBS CBS. CBS CBS CBS CBS CBS CBS CBS CBS CBS CBS. CBS CBS CBS CBS CBS CBS. CBS CBS CBS.}\\ 

\texttt{Toronto's twin city has famous tourist attractions including} & \texttt{the is, and Syrian aid,,, and aid. And and, and and, and, which, is, [ily, and is and,, and and and [ily was, no, and, and,,, and. no aidily, and Syriaily, do, no doubt de [ and which no Syrian is do a, no., and the, is and is, and, and and,, [ aid, and}\\ 

\texttt{The language used by people in Kentucky is} & \texttt{Finnish Finnish Finnish Finnish Finnish language Finnish Finnish, Finnish: Finnish: Finnish: Finnish: Finnish: Finnish. Finn - Finnish: Finnish. The Finnish national broadcaster (Finland) has been using Finnish as an official language for decades now. Finnish - Finland. Finland has had a long tradition of using Finnish as an official}

\end{tabular}
\caption{Text generated by GPT-J post the point of catastrophic forgetting when edited using ROME.}
\label{table:generation_examples}
\end{table*}

\clearpage
\subsection{Background}
In this section we will explain the details of four model editing algorithms explored in this paper: ROME \cite{ROME}, MEMIT \cite{MEMIT}, and MEND \cite{MEND}, and Fine-Tuning. 

\subsubsection{ROME}
Building off the discovery that feed-forward layers of a transformer function as key-value memories \cite{key-value-memories}, where neurons from $W_{fc}^{(l)}$ and $W_{proj}^{(l)}$ emulate keys and values respectively, \cite{ROME} hypothesize that insertion of new knowledge can take the form of some linear transformation $W$ such that $WK \approx V$ where $K$ and $V$ are the vector of keys and values respectively. For an updated fact represented by the key-value pair $(k_*,v_*)$, the constrained optimization problem can be summarized as follows

\begin{equation}
\text{min} \lVert \hat{W}K - V \rVert \ni \hat{W}k_* = v_*
\label{rome_opt}
\end{equation}
With the solution $\hat{W} = W + \Lambda (C^{-1}k_*)^{\top}$ where $C = KK^\top$ and $\Lambda = \frac{v_* - Wk_*}{(C^{-1}k_*)^\top k_*}$. The full derivation for the solution can be found in Appendix A in \cite{ROME}. 
To find the optimal $k_*$, inputs $x$ are taken where the subject $s$ is represented in the last token. $k_*$ is given by 
\begin{multline}
    k_* = \frac{1}{N}\sum_{j=1}^N k(x_j + s)\text{, where }\\ k(x) = \sigma(W_{fc}^{(l^*)}\gamma(a_{[x],i}^{(l^*)} + h_{[x],i}^{(l^*-1)}))
    \label{k_update}
\end{multline}
where $l^*$ is the desired layer, $i$ is the last subject token index, $h_{[x],i}^{(l^*-1)}$ is the hidden state of the previous layer, and $a_{[x],i}^{(l^*)}$ is the global attention of the hidden layer. Here, $N$ is set to 50, since the average is taken over 50 sampled prefixes $x_j$.
Optimal $v_* = \text{argmin}_z \mathcal{L}(z)$ where 
\begin{multline}
    \mathcal{L}(z) = \frac{1}{N}\sum_{j=1}^N-\log(\mathbb{P}_{G(m_i^{(l^*)}:= z)}[o^*|x_j+p])\\
    + D_{KL}(\mathbb{P}_{G(m_{i'}^{(l^*)}:= z)}[x|p'] \| \mathbb{P}_{G}[x|p'])
    \label{v_update}
\end{multline}
z is a vector that is substituted as the i-th token of the output to the MLP layer that enables the desired change to be realized. $G()$ substitutes the specified hidden state with the modified version. $p$ is the factual prompt, while $p'$ is the factual prompt rewritten in a form that begins with the subject. Given these prompts, $o^*$ is the new object. $v_*$ is solved using an Adam optimizer with a learning rate of 0.5 and weight decay rate of $1.5 \times 10^{-3}$. Following this, we compute the updates to the MLP weights using equation \ref{rome_opt}. ROME updates weights for GPT2-XL and GPT-J at layers 18 and 6 respectively.

\subsubsection{MEMIT}
Rather than overburdening one layer with an update, \cite{MEMIT} introduces MEMIT as a means of distributing the impact of the update across multiple layers. In doing so, they are able to largely scale the number of edits they can reliably make. In order to express the update, we want to find some $z_i = h_i^L + \delta_i$ such that, when substituted in place of $h_i^L$ at layer L, it is successful. We find this by optimizing $\delta_i$ using 
\begin{multline}
    z_i = h_i^L + \\
    \text{argmin}_{\delta_i} \frac{1}{P}\sum_{j=1}^P -\log \mathbb{P}_{G(h_i^L += \delta_i)} [ o_i | x_j \oplus p(s_i, r_i)]
\end{multline}
for the desired edit object $o_i$ and set of prompts ${x_j \oplus p(s_i, r_i)}$. Here, $x_j$ is a set of prefixes and $p(s_i, r_i)$ is a prompt generated from the edit subject $s_i$ and relation $r_i$. We want to find some update $\Delta^l$ for every layer $l \in R$ for a set of layers $R$ so that 
\begin{multline}
    \hat{W}_{\text{out}}^l := W_{\text{out}}^l + \Delta^l\text{ for all } l \in R\\
    \text{such that } \text{min}_{\Delta^l} \sum_i \lVert z_i - \hat{h}_i^L \rVert^2
    \label{MEMIT_weight_update}
\end{multline}
\begin{multline}
    \text{where } \hat{h}_i^L = h_i^0 + \sum_{l=1}^L a_i^l \\
    + \sum_{l=1}^L \hat{W}_{\text{out}}^l \sigma(W_{\text{in}}^l \gamma(h_t^{l-1}))
\end{multline}
 The closed form solution to this update is given by $\Delta^l = R^lK^{l\top}(C + K^lK^{l\top})^{-1}$. The full derivation can be found in \cite{MEMIT} section 4.2. To solve this, we need to find $K^l = [k_1^l,k_2^l,...,k_n^l]$ and $R^l = [r_1^l,r_2^l,...,r_n^l]$. This is found using 
 \begin{multline}
     k_i^l = \frac{1}{P}\sum_{j=1}^P k(x_j + s_i)\text{, } \\
     \text{where } k(x) = \sigma(W_{\text{in}}^l \gamma(h_i^{l-1}(x)))
 \end{multline}
 In this paper, \cite{MEMIT} define $m_{[t]}^l = W_{\text{out}}^l (\sigma (W_{\text{in}}^l \gamma(h_{[t]}^{l-1})))$. Given this, define
 \begin{multline}
     m_i^l = W_{\text{out}} k_i^l + r_i^l \\
     \text{where } r_i^l = \frac{z_i - h_i^L}{L-l+1}
 \end{multline}
 Note that the denominator of $r_i^l$ allows us to spread out the burden across multiple layer, allowing for a more scalable algorithm. It is hard to compute $C^l$ exactly, however it can be reliably estimated using $C^l = \lambda \mathbb{E}_k [k^lk^{l\top}]$ over randomly sampled inputs, where $\lambda = 1.5 \times 10^4$. Incorporating the update gives us our desired new weights $\hat{W}_{\text{out}}^l$
\subsubsection{MEND}
Using the fact that the gradient of loss L with respect to the weights $W_\ell$ of layer $\ell$ of an MLP has a rank-one decomposition such that $\nabla_{W_\ell}L = \sum_{i=1}^B \delta_{\ell+1}^iu_{\ell}^{i\top}$ for a batch $B$, \cite{MEND} are able to construct an editor network $g_\ell$ to generate the weight updates. Here, $\delta_{\ell+1}^i$ is the gradient for element $i$ for the preactivations of layer $\ell + 1$ and $u_{\ell}^{i}$ are the inputs of element $i$ into layer $\ell$. \\
To characterize these updates, MEND employs functions that map $\delta_{\ell+1}^i$ and $u_{\ell}^{i}$ to a pseudo-decomposition $\tilde{\delta}_{\ell+1}^i$ and $\tilde{u}_{\ell}^{i}$ such that $\tilde{\nabla}_{W_\ell}L = \sum_{i=1}^B \tilde{\delta}_{\ell+1}^i\tilde{u}_{\ell}^{i\top}$. Letting $z_\ell = \text{concat}(\delta_{\ell+1}\text{, } u_{\ell})$, the form of the network is 
\begin{align}
    h_\ell = z_\ell + \sigma(s_\ell^1 \odot (U_1 V_1 z_\ell + b) + o_\ell^1) \label{h_l}\\
    g(z_\ell) = h_\ell + \sigma(s_\ell^2 \odot U_2 V_2 h_\ell + o_\ell^2) \label{g_l}
\end{align}
where $\sigma$ is the ReLu activation function and $U_j$, $V_j$ are a low rank decomposition of MEND's weight for layer $j$. Note that, because of the difference in dimensions between weight matrices across layers, MEND learns different parameters for each unique shape of weight matrices to be edited. Additionally, layer-wise offset and scale parameters $o_\ell$ and $s_\ell$ are learned for both $h_\ell$ and $g_\ell$. The final update is given by $\tilde{W} = W - \alpha_\ell \tilde{\nabla}_{W_\ell}$ with $\alpha_\ell$ being another learned parameter per layer.\\
Given the original weights $W$ and the updated weights $\tilde{W}$, loss is computed aggregating two training losses, editing success and locality. For a desired edit $(x_e,y_e)$, $(x_e',y_e')$ is defined as a semantically equivalent wording of the edit. Editing loss is defined as $L_e = -\log_{p_{\theta_{\tilde{W}}}}p(y_e'|x_e')$. $x_{\text{loc}}$ is defined as a locality sample, which is randomly sampled to test the edited model's impact on information unrelated to the edit. The corresponding locality loss is $L_{\text{loc}} = D_{KL}(p_{\theta_W}(\cdot|x_{\text{loc}}) \| p_{\theta_{\tilde{W}}} (\cdot|x_{\text{loc}}))$.\\
The total loss is computed as $L_{\text{MEND}} = c_e L_e(\theta_{\tilde{W}}) + L_{\text{loc}}(\theta_W, \theta_{\tilde{W}})$ where $c_e = 0.1$. The total loss is optimized using the Adam optimizer. For GPT2-XL, we edit layers 46, 47, and 48. For GPTJ, we edit layers 26, 27, and 28.

\subsubsection{FT}
The Fine-Tuning procedures used in this paper follow from \cite{MEMIT} and \cite{ROME}'s implementation for both GPT-J and GPT2-XL. MLP weights for a single layer are fine-tuned for both models. We use a constrained fine tuning approach where we add a $L_\infty$ constraint such that $\lVert \theta_{G} - \theta_{G'} \rVert_\infty \leq \epsilon$ at each gradient step. For the constraint, $\epsilon = 5e-4$ for GPT2-XL and $\epsilon=5e-5$ for GPT-J. It is optimized using Adam with a learning rate of 5e-4 for both GPT2-XL and GPT-J. We fine tune layers 18 and 6 for GPT2-XL and GPT-J respectively. 

\end{document}